\documentclass[10pt,journal,compsoc]{IEEEtran}
\usepackage{amsmath}
\usepackage{amsthm}
\usepackage{color}
\usepackage{amssymb}
\usepackage{algorithm}
\usepackage{algorithmic}
\usepackage{array,multirow,graphicx}
\usepackage[table, dvipsname]{xcolor}
\usepackage{epstopdf}
\usepackage{upgreek}
\usepackage{subfigure}

\newtheorem{Theorem}{Theorem}

\newtheorem{Property}{Property}

\ifCLASSOPTIONcompsoc
  \usepackage[nocompress]{cite}
\else
  \usepackage{cite}
\fi
\ifCLASSINFOpdf
\else
\fi
\ifCLASSOPTIONcompsoc
 \usepackage[caption=false,font=footnotesize,labelfont=sf,textfont=sf]{subfig}
\else
 \usepackage[caption=false,font=footnotesize]{subfig}
\fi
\begin{document}
%
\title{Composite Quantization}
%
%
%
%

\author{Jingdong~Wang and
Ting Zhang\IEEEcompsocitemizethanks{\IEEEcompsocthanksitem
J. Wang and T. Zhang are with Microsoft Research,
Beijing, P.R. China. \protect\\
E-mail: \{jingdw, tinzhan\}@microsoft.com
}
}

%
%

\markboth{Journal of \LaTeX\ Class Files,~Vol.~13, No.~9, June~2016}%
{Shell \MakeLowercase{\textit{et al.}}: A Survey on Learning to Hash}
%



\IEEEtitleabstractindextext{%
\begin{abstract}
This paper
studies the compact coding approach
to approximate nearest neighbor search.
We introduce a composite quantization framework.
It uses the composition of several ($M$) elements, each of which
is selected from a different dictionary,
to accurately approximate a $D$-dimensional vector,
thus  yielding accurate search,
and represents the data vector
by a short code
composed of the indices
of the selected elements
in the corresponding dictionaries.
Our key contribution lies in
introducing a near-orthogonality constraint,
which makes the search efficiency is guaranteed
as
the cost of the distance computation is
reduced to $O(M)$ from $O(D)$
through a distance table lookup scheme.
The resulting approach is called near-orthogonal composite quantization.
We theoretically justify the equivalence
between
near-orthogonal composite quantization
and minimizing an upper bound
of a function formed
by jointly considering the quantization error and the search cost
according to a generalized triangle inequality.
We empirically show the efficacy of the proposed approach
over several benchmark datasets.
In addition,
we demonstrate the superior performances
in other three applications:
combination with inverted multi-index,
quantizing the query for mobile search,
and inner-product similarity search.

\end{abstract}

}

\maketitle

\IEEEdisplaynontitleabstractindextext

%
\IEEEpeerreviewmaketitle

\IEEEraisesectionheading{\section{Introduction}\label{sec:introduction}}
\IEEEPARstart{N}{earest}
neighbor (NN) search has been a fundamental research topic
in machine learning, computer vision,
and information retrieval~\cite{ShakhnarovichDI06}.
The goal of NN search,
given a query $\mathbf{q}$,
is to find a vector $\operatorname{NN}(\mathbf{q})$
whose distance to the query is the smallest
from $N$ $D$-dimensional
reference (database) vectors.

The straightforward solution, linear scan, is to compute
the distances to all the database vectors
whose time cost is $O(ND)$
and is very time-consuming,
and thus impractical
for large scale high-dimensional cases.
Multi-dimensional indexing methods,
such as the $k$-d tree~\cite{FriedmanBF77},
have been developed
to speed up exact search.
For high-dimensional cases it
turns out that such approaches are not much more efficient (or
even less efficient) than linear scan.

Approximate nearest neighbor (ANN) search
has been attracting a lot of interests
because of competitive search accuracy
and tractable storage and search time cost.
The algorithms can be split into two main categories:
(1) accelerating the search by comparing
the query with a small part of reference vectors
through an index structure,
such as random $k$-d trees~\cite{Silpa-AnanH08},
FLANN~\cite{MujaL09}, and neighborhood graph~\cite{WangL12};
and (2)
accelerating the distance computation
between the query and the reference vectors
through the compact coding technique,
i.e.,
converting
the database vectors into short codes,
with typical solutions including hashing~\cite{GongLGP13,YuKGC14, LiuWJJC12, WangKC12},
quantization~\cite{JegouDS11, NorouziF13, ZhangDW14}.

In this paper,
we introduce a composite quantization framework
to convert vectors to compact codes.
The idea is to approximate a vector
using the composition (addition) of $M$ elements
each selected from one dictionary,
and to represent this vector by a short code
composed of the indices of the selected elements.
The way of adding the selected dictionary elements
is different from
concatenating the selected dictionary elements (subvectors)
adopted in
product quantization~\cite{JegouDS11}
and its extensions, Cartesian $k$-means~\cite{NorouziF13}
and optimized product quantization~\cite{GeHK013},
that divide the space
into partitions
and conduct $k$-means separately over each partition
to obtain the dictionaries.
The advantage is that
the vector approximation,
and accordingly the distance approximation of a query
to the database vector,
is more accurate,
yielding more accurate nearest neighbor search.

To efficiently evaluate the distance between a query
and the short code representing the database vector,
we first present a naive solution,
called orthogonal composite quantization,
by introducing orthogonality constraints,
i.e.,
the dictionaries are mutually orthogonal.
The benefit
is that
the approximated distance
can be calculated
from the distance of the query to each selected element,
taking only a few distance table lookups,
and that the time cost is reduced to $O(M)$ from $O(D)$.
Furthermore,
we propose a better solution,
called near-orthogonal composite quantization,
by relaxing orthogonality constraints
to near-orthogonality constraints,
i.e.,
the summation of the inner products
of all pairs of elements
that are used to approximate the vector
but from different dictionaries
is constant.
The distance computation is still efficient,
while
the distance approximation is more accurate,
and accordingly the search accuracy is higher.

The resulting near-orthogonal composite quantization (NOCQ) is justified
in both theory and experiments.
We present a generalized triangle inequality,
theoretically explaining
that near-orthogonal composite quantization
is equivalent to minimizing an upper bound
of a function
that is formed
by
jointly considering the quantization error and the search time cost.
We also show that
production quantization~\cite{JegouDS11} and Cartesian $k$-means~\cite{NorouziF13}
are constrained versions of
NOCQ:
NOCQ relaxes the orthogonality constraint between dictionaries
and does not require the explicit choice
of the dimension of each subspace corresponding to each dictionary.
We present empirical
results over several standard datasets
demonstrate that the proposed approach achieves
state-of-the-art performances
in approximate nearest neighbor search in terms of the Euclidean distance.
In addition,
we demonstrate the superior performances
in other three applications:
combination with inverted multi-index,
quantizing the query for mobile search,
and inner-product similarity search.

%

\section{Related work}
\label{sec:overview}
A comprehensive survey on learning to hash
is given in~\cite{WangZSSS16},
showing that
the quantization algorithms~\cite{JegouDS11, NorouziF13, WangSSJ14,WangZSSS16} achieve better search quality
than hashing algorithms
with Hamming distance, even with optimized or asymmetric distances~\cite{GordoPGL14, WangWSXSL14}.
Thus,
this paper only presents a brief review
of the quantization algorithms.

Hypercubic quantization,
such as iterative quantization~\cite{GongLGP13}, isotropic hashing~\cite{KongL12a},
harmonious hashing~\cite{XuBLCHC13}, angular quantization~\cite{GongKVL12},
can be regarded as a variant of scalar quantization
by optimally rotating the data space
and performing binary quantization along each dimension in the rotated space,
with the quantization centers fixed at $-1$ and $1$ (or equivalently $0$ and $1$).
Such a way of fixing quantization centers puts a limit on
the number of possible distances in the coding space, which
also limits the accuracy of distance approximation
even using optimized distances~\cite{GordoPGL14, WangWSXSL14}.
Therefore, the overall search performance is not comparable to product quantization and Cartesian $k$-means.

Product quantization~\cite{JegouDS11} divides the data space
into (e.g., $M$) disjoint subspaces.
Accordingly, each database vector is divided
into $M$ subvectors,
and the whole database
is also split into $M$ sub-databases. A number of clusters
are obtained by conducting $k$-means over each sub-database.  Then
a database vector is approximated
by concatenating the nearest cluster center
of each subspace,
yielding a representation with a short code
containing the indices of the nearest cluster centers.
The computation of the distance between two vectors is accelerated
by looking up a precomputed table.

Cartesian $k$-means~\cite{NorouziF13}
(or optimized product quantization~\cite{GeHK013})
improves the compositionabilty,
i.e., vector approximation accuracy,
by finding an optimal feature space rotation
and then performing product quantization over the rotated space.
Additive quantization~\cite{BabenkoK14, DuW14}
further improves the compositionabilty
by approximating a database vector using the summation of dictionary elements
selected from different dictionaries,
whose idea is similar to structured vector quantization~\cite{GrayN98}
(a.k.a., multi-stage vector quantization and residual quantization).
It has been applied to data compression~\cite{BabenkoK14}
and inner product similarity search~\cite{DuW14},
yet is not suitable for search with Euclidean distance
due to the lack of the acceleration of distance computation.

There are other attempts
to improve product quantization
in the other ways, such as distance-encoded product quantization~\cite{HeoLY14}
and locally optimized product quantization~\cite{KalantidisA14},
which can also be combined with our approach in the same way.
The hash table scheme is studied in~\cite{MatsuiYZ15}
to accelerate the linear scan over
the product quantization codes.
Inverted multi-index~\cite{BabenkoL12}
applies product quantization
to build an inverted index for searching a very large scale database,
with the ability
of efficiently retrieving the candidates
from a large number of inverted lists.
Bilayer product quantization~\cite{BabenkoL14}
improves the efficiency of distance computation
within the inverted multi-index framework.
We will also apply the proposed approach to inverted multi-index
to show its effectiveness.

This paper
represents a very substantial extension
of our previous conference paper~\cite{ZhangDW14}\footnote{The terminology in this paper is different from the conference paper.}
with the introduction
of alternative versions of
near-orthogonal composite quantization,
and an additional material added from our report~\cite{DuW14}.
The main technical novelties compared with~\cite{ZhangDW14}
lie in three-fold.
(1) We introduce two alternative objective functions:
one is from the strict orthogonality constraint,
and the other is from
the generalized triangle inequality.
(2) We present extensive analysis of composite quantization,
including why quantization is proper for approximate nearest neighbor search,
why multiple different dictionaries rather than
a single dictionary are used.
(3) We conduct more experiments,
such as
quantizing the query for mobile search,
and inner-product similarity search.

\section{Preliminaries}
\noindent\textbf{Approximate nearest neighbor search.}
Nearest neighbor (NN) search is a problem of,
given a query vector $\mathbf{q}$,
finding a vector $\operatorname{NN}(\mathbf{q})$
from a set of $N$ $D$-dimensional reference vectors
$\mathcal{X}=\{\mathbf{x}_1,\dots,\mathbf{x}_N\}$,
such that its distance to the query vector
is minimum,
i.e.,
$\operatorname{NN}(\mathbf{q}) = \arg\min_{\mathbf{x} \in \mathcal{X}}\operatorname{dist}(\mathbf{q}, \mathbf{x})$.
Here,
$\operatorname{dist}(\mathbf{q}, \mathbf{x})$ is a distance between $\mathbf{q}$ and $\mathbf{x}$,
and the Euclidean distance $\|\mathbf{q} - \mathbf{x}\|_2$ is a typical example.
Approximate nearest neighbor (ANN) search,
the focus of this paper,
aims efficiently find a good guess of the exact nearest neighbor
rather than the exact one.

\vspace{.1cm}
\noindent\textbf{Compact coding.}
The search cost of the naive linear scan algorithm
is $O(ND)$,
where $D$ comes from the distance evaluation $\operatorname{dist}(\mathbf{q}, \mathbf{x})$.
Compact coding in general
is a low-dimensional embedding approach
and
represents an input vector $\mathbf{x}$
by a short descriptor $\mathbf{y} = f(\mathbf{x}; \boldsymbol{\uptheta})$.
Here $\boldsymbol{\theta}$ is the parameters,
and $\mathbf{y}$ could be a binary representation vector
or a sequence of ($M$) binary codes,
e.g., $\mathbf{y} = [y_1~ y_2~ \cdots~ y_M]^\top$
where $y_m$ is a byte-valued variable.
The application of compact coding to approximate nearest neighbor search
uses the distance computed in the code space,
as an substitute of the distance in the input space,
to rank the reference vectors.
The benefit
is that the distance computation cost in the compact code space
is reduced to $O(D')$,
where $D' \ll D $,
and consequently the search efficiency is improved.

\vspace{.1cm}
\noindent\textbf{Quantization.}
Quantization is a process
of mapping a large set of input vectors
to a countable (smaller) set of representative vectors
(e.g., $\mathcal{C}=\{\mathbf{c}_1,\dots,\mathbf{c}_{K'}\}$),
called dictionary.
For instance, a reference vector $\mathbf{x}_n$
is mapped to a dictionary element,
$\bar{\mathbf{x}}_n = \mathbf{c}_{k^n}$,
such that
the reconstruction error is minimal,
$\mathbf{c}_{k^n} = \arg\min_{\mathbf{c} \in \mathcal{C}} \|\mathbf{x}_n - \mathbf{c}\|_2$,
and accordingly $\mathbf{x}_n$ can be represented by a short code $k^n$.
Quantization has been widely adopted
in image and signal compression.

As a compact coding approach,
quantization is successfully applied to
approximate nearest neighbor search
by using the asymmetric distance between the query $\mathbf{q}$ and the quantized vector $\bar{\mathbf{x}}_n$,
$\|\mathbf{q} - \bar{\mathbf{x}}_n\|_2$,
to approximate the original distance,
$\|\mathbf{q} - \mathbf{x}_n\|_2$.
The search accuracy is guaranteed
by the following theorem.
\begin{Theorem}[Triangle inequality]
	\label{theorem:upperbound}
	The reconstruction error of the distances is upper-bounded:
	$ |\|\mathbf{q} - \bar{\mathbf{x}}_n\|_2 - \|\mathbf{q} - \mathbf{x}_n\|_2|
	\leq \|\bar{\mathbf{x}}_n -  \mathbf{x}_n \|_2$.
\end{Theorem}

\begin{figure}[t]
	\centering
	\includegraphics[width=.9\linewidth, clip]{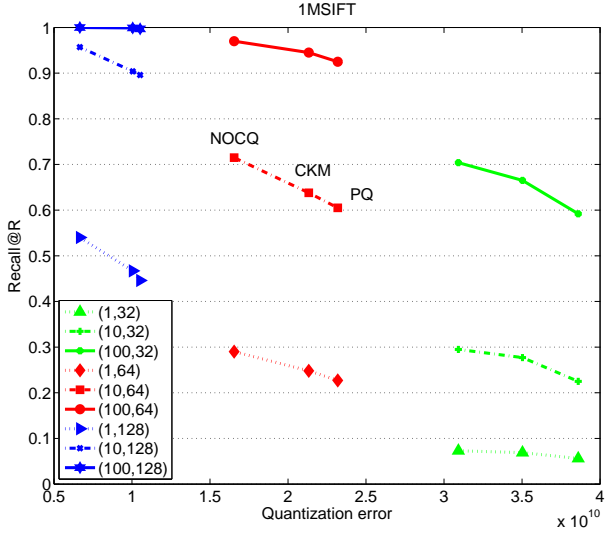}
\vspace{-.3cm}
	\caption{Illustrating the relation between recall and distortion error.
		Red, green, and blue lines correspond to recall at positions
		$1$, $10$, and $100$, respectively.
		Fore each color,
		the line segments from left to right correspond to
		$128$, $64$, and $32$ bits.
		The mark on each line represents
		NOCQ, CKM, and PQ from left to right respectively.
		For $(R,B)$ in the
		legend, $R$ indicates the rank position
		to evaluate the recall,
		and $B$ indicates the number of bits.}
	\label{fig:recalldistortion}
\end{figure}

This suggests that
the quantization error is an upper bound
of the distance reconstruction error,
and thus the search accuracy depends on
the quantization quality:
low quantization error usually
leads to high search accuracy.
An empirical comparison in terms of
the quantization error (i.e., $\sum_{n=1}^N \|\mathbf{x}_n-\bar{\mathbf{x}}_n\|_2^2$) vs. the search accuracy
among
product quantization (PQ),
cartesian k-means (CKM),
and our approach, near-orthogonal composite quantization (NOCQ),
is shown in Figure~\ref{fig:recalldistortion}.
We can see that the proposed approach
is able to achieve lower quantization error and hence has
better search performance.

\vspace{.1cm}
\noindent\textbf{Cartesian quantization.}
The state-of-the-art approach,
which we call Cartesian quantization,
constructs the dictionary $\mathcal{C}$
from several (e.g., $M$) small dictionaries
$\{\mathcal{C}_1, \mathcal{C}_2, \cdots, \mathcal{C}_M\}$,
called source dictionary,
where $\mathcal{C}_m = \{\mathbf{c}_{m1}, \mathbf{c}_{m2}, \cdots, \mathbf{c}_{mK_m}\}$
(for simplicity,
we assume $K_1 = K_2 = \cdots = K_M = K$),
through the Cartesian product operation,
$\mathcal{C} = \mathcal{C}_1 \times \mathcal{C}_2 \times \cdots \times \mathcal{C}_M = \{(\mathbf{c}_{1k_1}, \mathbf{c}_{2k_2}, \cdots, \mathbf{c}_{Mk_M}) ~|~ \mathbf{c}_{mk_m} \in \mathcal{C}_m \}$,
where $K^M$ $=\prod_{m=1}^M K_m$ $M$-tuples $(\mathbf{c}_{1k_1}, \mathbf{c}_{2k_2}, \cdots, \mathbf{c}_{Mk_M})$
form the composite dictionary.
The benefits include that several small dictionaries generate a larger composite dictionary,
with the consequence that
the quantization error can be smaller
and that
the encoding time is reduced while a longer code (larger dictionary)
is actually used.

In contrast to the application to compression
whose goal is to reconstruct the vector from the compact code,
the application to approximate nearest neighbor search
aims to accelerate the distance reconstruction.
Typically, the distance between a query $\mathbf{q}$
and a $M$-tuple $(\mathbf{c}_{1k_1}, \mathbf{c}_{2k_2}, \cdots, \mathbf{c}_{Mk_M})$
that approximates a vector $\mathbf{x}$,
is computed
from the $M$ distances $\{\operatorname{dist}(\mathbf{q}, \mathbf{c}_{mk_m})\}_{m=1}^M$
through looking up the distance table
recording the distances between the query and the elements in the source dictionary,
which reduces the time cost
from $O(D)$ to $O(M)$.

Our proposed approach belongs to this Cartesian quantization category,
which also contains the product quantization~\cite{JegouDS11}
and Cartesian $k$-means~\cite{NorouziF13} (or optimized product quantization~\cite{GeHK013}),
in which the dictionaries are mutual-orthogonal.
The novelty of our approach lies in
adopting the addition scheme to form the composite dictionary element from the $M$-tuples,
which removes the mutual orthogonality constraint between the dictionaries,
and introducing the near-orthogonality condition to theoretically
guarantee the efficiency of the distance reconstruction.

\section{Formulation}
\noindent\textbf{Composite quantization.}
Our approach forms the composite dictionary element
from the $M$-tuples $(\mathbf{c}_{1k_1}, \mathbf{c}_{2k_2}, \cdots, \mathbf{c}_{Mk_M})$
through an addition operation,
$\sum_{m=1}^M \mathbf{c}_{mk_m}$.
A vector $\mathbf{x}_n$
is approximated as
\begin{align}
\mathbf{x}_n \approx \bar{\mathbf{x}}_n
= \sum\nolimits_{m=1}^M \mathbf{c}_{mk^n_m},
\label{eqn:vectorapproximation}
\end{align}
where $k^n_m$ is the index of the element
selected from the source dictionary $\mathcal{C}_m$
for the $n$th reference vector $\mathbf{x}_n$.
We find the dictionary
by minimizing
\begin{align}
\sum\nolimits_{n=1}^N \|\mathbf{x}_n - \bar{\mathbf{x}}_n \|_2^2
=
\sum\nolimits_{n=1}^N \|\mathbf{x}_n - \sum\nolimits_{m=1}^M \mathbf{c}_{mk_m^n}\|_2^2.\label{eqn:quantizationerrors}
\end{align}

Given the approximation through composite quantization,
$\mathbf{x} \approx \bar{\mathbf{x}} = \sum\nolimits_{m=1}^M \mathbf{c}_{mk_m}$,
the distance of a query $\mathbf{q}$ to the approximation $\bar{\mathbf{x}}$
is
\begin{align}
\|\mathbf{q} - \bar{\mathbf{x}}\|_2
= \|\mathbf{q} - \sum\nolimits_{m=1}^M \mathbf{c}_{mk_m}\|_2.
\end{align}
It is time-consuming to
reconstruct the approximate vector $\bar{\mathbf{x}}$
(taking $O(MD)$)
and then compute the distance
between $\mathbf{q}$ and $\bar{\mathbf{x}}$ (taking $O(D)$).
In the following, we introduce two solutions to accelerate the distance computation.

\vspace{.1cm}
\noindent\textbf{Orthogonal composite quantization.}
We introduce an extra constraint,
mutual orthogonality of the $M$ subspaces,
each spanned by the corresponding dictionary:
\begin{align}
\mathbf{C}_i^{\top}\mathbf{C}_j = \mathbf{0}, ~\forall i \neq j \label{eqn:strictorthogonality}
\end{align}
where $\mathbf{C}_i$ is the matrix form
of the $i$th dictionary.
We show that composite quantization
with this constraint,
called orthogonal composite quantization (OCQ),
makes the distance computation efficient.

Let $\mathbf{P}_m$ be the subspace
spanned by the dictionary $\mathcal{C}_m$.
Assume that the $M$ subspaces cover the whole $D$-dimensional space.
The query $\mathbf{q}$ is then transformed as below,
\begin{align}
\mathbf{q} = \sum\nolimits_{m=1}^M \mathbf{P}_m \mathbf{P}_m^{\top}\mathbf{q}.
\end{align}
The distance between the query and the approximate database vector
is calculated in the following way,
\begin{align}
&~\|\mathbf{q} - \sum\nolimits_{m=1}^M \mathbf{c}_{mk_m}\|_2^2 \nonumber \\
= &~\| \sum\nolimits_{m=1}^M \mathbf{P}_m \mathbf{P}_m^{\top}\mathbf{q} -  \sum\nolimits_{m=1}^M \mathbf{c}_{mk_m}\|_2^2 \nonumber \\
= &~\| \sum\nolimits_{m=1}^M (\mathbf{P}_m \mathbf{P}_m^{\top}\mathbf{q} -  \mathbf{c}_{mk_m})\|_2^2 \nonumber \\
= &~\sum\nolimits_{m=1}^M  \| \mathbf{P}_m \mathbf{P}_m^{\top}\mathbf{q} -  \mathbf{c}_{mk_m}\|_2^2 \nonumber \\
= &~\sum\nolimits_{m=1}^M  \|\mathbf{P}_m^{\top}\mathbf{q} -  \mathbf{P}_m^\top \mathbf{c}_{mk_m}\|_2^2.
\end{align}
Consequently,
the cost of the distance computation
is reduced from $O(D)$
to $O(M)$
by looking up the precomputed table
storing the distances
between the query and the dictionary elements in each subspace.

\vspace{.1cm}
\noindent\textbf{Near-orthogonal composite quantization.}
We expand the distance computation
into three terms:
\begin{align}\small
&\|\mathbf{q} - \sum\nolimits_{m=1}^M \mathbf{c}_{mk_m}\|_2^2
= \sum\nolimits_{m=1}^M \|\mathbf{q}- \mathbf{c}_{mk_m} \|_2^2\\
& - (M-1)\|\mathbf{q}\|_2^2
+ \sum\nolimits_{i=1}^M \sum\nolimits_{j=1, j\neq i}^M \mathbf{c}_{ik_i}^T \mathbf{c}_{jk_j}.
\label{eqn:distancetransform}
\end{align}
We can see that,
given the query $\mathbf{q}$,
the second term in the right-hand side,
$- (M-1)\|\mathbf{q}\|_2^2$,
is constant for all the database vectors.
and hence is unnecessary to compute
for nearest neighbor search.

The first term $\sum\nolimits_{m=1}^M \|\mathbf{q}- \mathbf{c}_{mk_m} \|_2^2$ is the summation of
the distances of the query to the selected dictionary elements,
and can be efficiently computed
using a few ($M$) additions by looking up a distance table,
where the distance table,
$\mathbf{T} = [t_{mk}]_{M\times K}$, is precomputed
before comparing the query with each reference vector,
and stores the distances of the query to the $MK$ dictionary elements,
$$\{t_{mk} = \|\mathbf{q} - \mathbf{c}_{mk}\|_2^2;
m=1, \cdots, M, k=1, \cdots, K\}.$$

Similarly, we can build a table
storing the inner products between dictionary elements,
$$\{\mathbf{c}_{mk}^T \mathbf{c}_{m'k'}; m \neq m', k=1,\cdots,K, k'=1,\cdots,K\},$$
and compute the third term
using $O(M^2)$ distance table lookups.
This results in
the distance computation cost is changed
from $O(MD)$ to $O(M^2)$,
which is still large.
For instance, in the case where $D=128$ and $M=16$,
$M^2 = 256$ is larger than $D = 128$,
which means that the cost is greater than
that using the original vector.

It can be seen that
with the orthogonality constraint~(\ref{eqn:strictorthogonality})
the third term is equal to $0$:
\begin{align}
\sum\nolimits_{i=1}^M \sum\nolimits_{j=1, j\neq i}^M \mathbf{c}_{ik_i}^T \mathbf{c}_{jk_j} = 0.
\end{align}
Thus, the computation cost is reduced to $O(M)$.
We notice that, if the third term is a constant,
\begin{align}
\sum\nolimits_{i=1}^M \sum\nolimits_{j=1, j\neq i}^M \mathbf{c}_{ik_i}^T \mathbf{c}_{jk_j} = \epsilon, \label{eqn:nearorthogonality}
\end{align}
called near-orthogonality,
the third term can be discarded
for the distance computation.
Consequently, we only need to compute the first term
for nearest neighbor search,
and the computation cost is also reduced to $O(M)$.
The resulting approach is called near-orthogonal composite quantization (NOCQ).
The goal is to minimize the quantization errors (problem~(\ref{eqn:quantizationerrors}))
subject to the near-orthogonality constraint~(\ref{eqn:nearorthogonality}).

Near-orthogonality can be viewed as
a relaxation of strict orthogonality
because Equation~(\ref{eqn:nearorthogonality}) (near-orthogonality)
is a necessary but not sufficient condition
of Equation~(\ref{eqn:strictorthogonality}) (strict orthogonality)
while
Equation~(\ref{eqn:strictorthogonality})
is a sufficient but not necessary condition
of Equation~(\ref{eqn:nearorthogonality}).
Thus, the vector approximation with the relaxed condition,
near-orthogonality,
is more accurate.
Empirical comparisons in Table~\ref{tab:dictionaryorthogonal}
show that
the vector approximation and accordingly the distance approximation
of NOCQ
is more accurate than OCQ.


\vspace{.1cm}
\noindent\textbf{Joint accuracy and efficiency optimization.}
We first introduce several notations.
(1) The square root of the first term
in the right-hand side of Equation~(\ref{eqn:distancetransform})
is denoted by $\tilde{d}(\mathbf{q},\bar{\mathbf{x}}) $:
$$\tilde{d}(\mathbf{q},\bar{\mathbf{x}}) = (\sum\nolimits_{m=1}^M\|\mathbf{q} - \mathbf{c}_{mk_m}\|_2^2)^{1/2}.$$
(2) The square root of
the summation of the square of the true Euclidean distance
and a query-dependent term $(M-1)\|\mathbf{q}\|_2^2$
is written as
$$\hat{d}(\mathbf{q},\mathbf{x}) = (\|\mathbf{q} - \mathbf{x}\|_2^2 + (M-1)\|\mathbf{q}\|_2^2)^{1/2}.$$
(3) Accordingly we define the approximate version,
$$\hat{d}(\mathbf{q},\mathbf{x}) \approx \hat{d}(\mathbf{q},\bar{\mathbf{x}}) = (\|\mathbf{q} - \bar{\mathbf{x}}\|_2^2 + (M-1)\|\mathbf{q}\|_2^2)^{1/2}.$$
(4) The third term
in the right-hand side of Equation~(\ref{eqn:distancetransform})
is denoted as $\delta$:
$\delta = \sum_{i\neq j}\mathbf{c}_{ik_i}^T\mathbf{c}_{jk_j}$.

The near-orthogonal composite quantization approach
uses $\tilde{d}(\mathbf{q},\bar{\mathbf{x}})$
as the distance for nearest neighbor search,
which is essentially an approximation of $\hat{d}(\mathbf{q},\bar{\mathbf{x}})$
(with $\delta$ dropped)
as we have, by definition,
$$\hat{d}(\mathbf{q},\bar{\mathbf{x}}) = (\tilde{d}^2(\mathbf{q},\bar{\mathbf{x}}) + \delta)^{1/2},$$
and thus an approximation of $\hat{d}(\mathbf{q},\mathbf{x})$.
Notice that $\hat{d}(\mathbf{q},\mathbf{x})$ only depends on
the true distance between $\mathbf{q}$ and $\mathbf{x}$
and a query-dependent term
that is a constant
for the search with a specific query.
Ideally, if $\tilde{d}(\mathbf{q},\bar{\mathbf{x}}) = \hat{d}(\mathbf{q},\mathbf{x})$,
the search accuracy would be $100\%$.

In general, the absolute difference
$|\tilde{d}(\mathbf{q},\bar{\mathbf{x}}) - \hat{d}(\mathbf{q},\mathbf{x})|$
is expected to be small to guarantee high search accuracy.
We have the following theorem:
\begin{Theorem}[Generalized triangle inequality]
	\label{theorem:upperbound}
	The reconstruction error of the distances is upper-bounded:
	$ |\tilde{d}(\mathbf{q},\bar{\mathbf{x}})-\hat{d}(\mathbf{q},\mathbf{x})|
	\leq \|\mathbf{x}-\bar{\mathbf{x}}\|_2+|\delta|^{1/2} $.
\end{Theorem}

This theorem suggests a solution
to minimize the distance reconstruction error
by minimizing the upper-bound:
$$\min \|\mathbf{x} - \bar{\mathbf{x}}\|_2 + {|\delta|}^{1/2},$$
which is then transformed to a minimization problem
with a looser upper bound:
$$\min \|\mathbf{x} - \bar{\mathbf{x}}\|^2_2 + {|\delta|}.$$
Accumulating the upper bounds over all the database vectors,
we get
\begin{align}
\min \sum\nolimits_{n=1}^N (\|\mathbf{x}_n - \bar{\mathbf{x}}_n\|^2_2 + |\delta|). \label{eqn:minimizeupperbound}
\end{align}
The near-orthogonal composite quantization formulation
divided the accumulated upper bound into two parts:
\begin{align}
\min \sum\nolimits_{n=1}^N \|\mathbf{x}_n - \bar{\mathbf{x}}_n\|^2_2 ~~~~
\operatorname{s. t.} ~&~\delta = \epsilon, \label{eqn:NOCQformulation}
\end{align}
which is essentially
an approximation of (\ref{eqn:minimizeupperbound}).

The upper bound in (\ref{eqn:minimizeupperbound})
consists of two terms,
the quantization error term,
indicating the degree of the vector approximation,
and
the near-orthogonality term,
determining the computation efficiency.
In this sense, our approach provides a solution of
jointly optimizing the search accuracy and the search efficiency.

\begin{figure*}[t]
	\centering
	(a)~\includegraphics[width=.45\linewidth, clip]{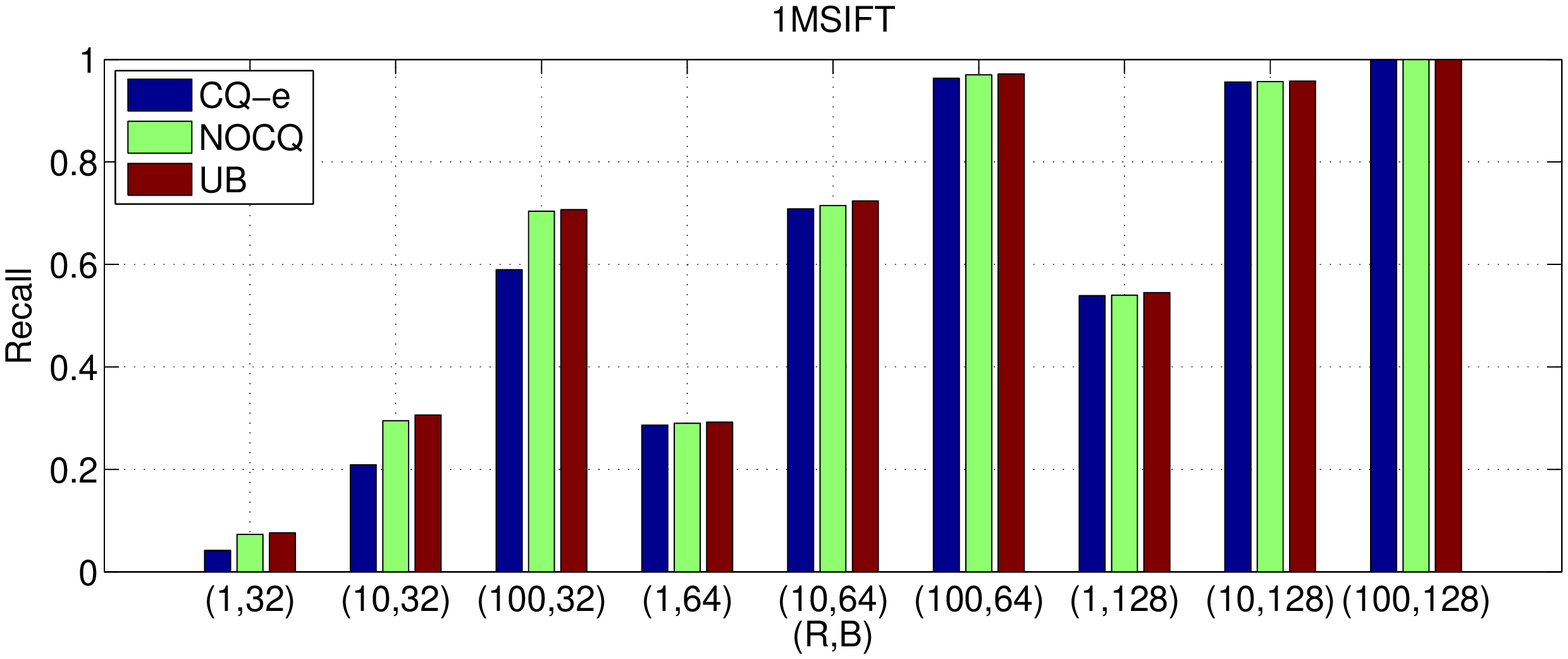} ~~~
	(b)~\includegraphics[width=.45\linewidth, clip]{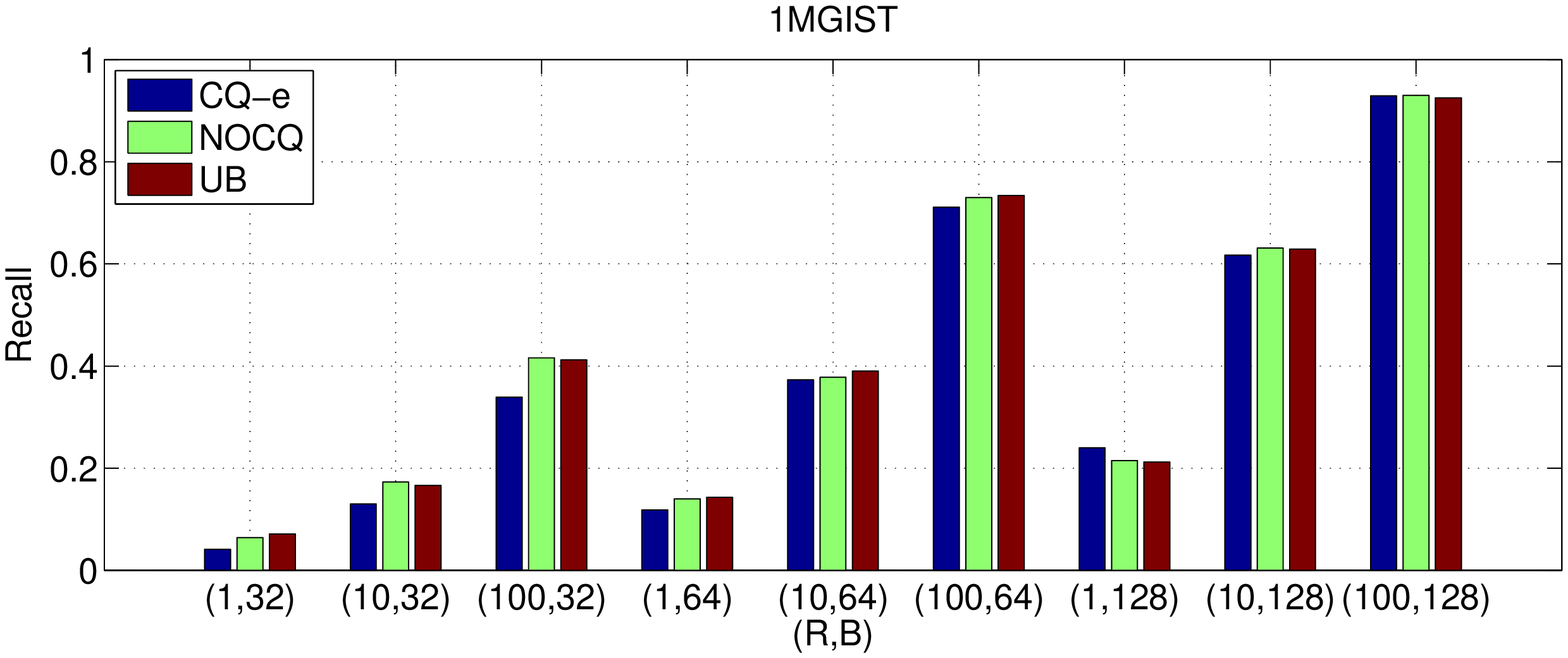}
\vspace{-.3cm}
	\caption{The results of directly minimizing the upper bound, denoted as UB,
		and composite quantization with encoding the third term, denoted as CQ-e,
		on $1M$SIFT and $1M$GIST.
		The horizontal axis labelled as
		$(R,B)$ indicates the recall@$R$ performance
		when encoded with $B$ bits. }
	\label{fig:upperbound}

\end{figure*}

An alternative formulation to optimize the search efficiency
is to quantize the third term $\delta = \sum_{i\neq j}\mathbf{c}_{ik_i}^T\mathbf{c}_{jk_j}$
in Equation~(\ref{eqn:distancetransform}) into a single byte
and decrease the dictionary size for guaranteeing
the whole code size not changed.
We empirically compare the results of (a) composite quantization with
quantizing $\delta$,
(b) direct minimization of the upper bound
(solving the problem (\ref{eqn:minimizeupperbound}))
and (c) near-orthogonal composite quantization.
The comparison on the $1M$SIFT and $1M$GIST datasets
is shown in Figure~\ref{fig:upperbound},
which indicates that (a) performs slightly lower when the code sizes are smaller,
and (b) and (c) performs similarly.

\section{Connections and Discussions}
\noindent\textbf{$M$ source dictionaries versus a single dictionary.}
Composite quantization uses $M$ source dictionaries $\{\mathcal{C}_1, \mathcal{C}_2, \cdots, \mathcal{C}_M\}$
to generate an $M$-tuple $(\mathbf{c}_{1k_1}, \mathbf{c}_{2k_2}, \cdots, \mathbf{c}_{Mk_M})$
for vector approximation.
Each element in the $M$-tuple is selected from a different source dictionary.
In the following,
we discuss two possible $M$-tuple choices in terms of different source dictionaries construction:
(1) Merge $M$ source dictionaries as one dictionary,
and select $M$ elements from the merged dictionary
to form an $M$-tuple;
and (2) reduce the size of the merged dictionary
from $MK$ to $K$
(i.e., perform an \emph{$M$-selection}
\footnote{In mathematics, an $M$-selection of
	a set $\mathcal{S}$
	is a subset of $S$ not necessarily distinct elements of $\mathcal{S}$.} of a dictionary $\mathcal{C}$ of size $K$).

\noindent\emph{A single dictionary of size $MK$:}
The main issue is that
it increases the code length.
When using $M$ source dictionaries
with each dictionary containing $K$ elements,
the compact code is $(k_1, k_2, \cdots, k_M)$
and the code length is $M\log K$.
In contrast,
if using the merged dictionary,
which contains $MK$ elements,
the code length is changed to $M\log(MK)$,
larger than $M\log K$.

\noindent\emph{A single dictionary of size $K$:}
The code length is reduced to $M \log K$,
the same to that in composite quantization,
but the vector approximation is not as accurate
as that in composite quantization.
We denote our selection scheme as~\emph{group $M$-selection}
since it selects $M$ elements from a group of $M$ dictionaries.
It is easy to show
that the $M$-selection from a source dictionary
is equivalent to the group $M$-selection
when the $M$ source dictionaries are the same.

In the following,
we present a property,
to compare the optimal objective function values (quantization errors)
of $M$-selection
and group $M$-selection,
which are denoted by
$f^*_{ms}$
and
$f^*_{gms}$,
respectively.

\begin{Property}
	\label{property:relationsofobjectivevalues}
	Given the same database $\mathcal{X}$
	and the same values of $K$ and $M$,
	we have
	$f^*_{gms} \leqslant f^*_{ms}$.
\end{Property}

We compute
the cardinalities of the composite dictionaries
to show the difference in another way.
Generally,
the objective value would be smaller
if the cardinality of the composite dictionary is larger.
The cardinalities are summarized as follows.

\begin{Property}
	\label{property:cardinalities}
	The maximum cardinalities of group $M$-selection
	and
	$M$-selection
	are $K^M$
	and
	$\binom{K+M-1}{M}
	=\frac{(K+M-1)!}{M!(K-1)!}$,
	respectively.
	We have
	$K^M \geqslant \binom{K+M-1}{M}$,
	the maximum cardinality of group $M$-selection
	is greater than that of $M$-selection.
\end{Property}

The above analysis shows that
composite quantization
with group $M$-selection
can achieve more accurate vector approximation,
which can be easily extended to
its (near-)orthogonal version.

\vspace{.1cm}
\noindent\textbf{$K$-means and sparse coding.}
Composite quantization as well as the near-orthogonal version,
when only one dictionary is used
(i.e., $M=1$),
are degraded to the $k$-means approach.
Compared with $k$-means,
composite quantization is able to produce
a larger number of quantized centers ($K^M$)
using a few dictionary elements ($MK$),
resulting in that
the composite quantizer can be indexed
in memory for large scale quantized centers.

Composite quantization is also related
to coding with block sparsity~\cite{YuanL06},
in which the coefficients are divided into several blocks
and the sparsity constraints are imposed in each block separately.
Composite quantization
can be regarded as a sparse coding approach,
where the coefficients that can only be
valued by $0$ or $1$
are divided into $M$ groups,
for each group the non-sparsity degree is $1$,
and an extra constraint,
near-orthogonality,
is considered.

\begin{figure}[t]
	\centering
	\includegraphics[width=.9\linewidth, clip]{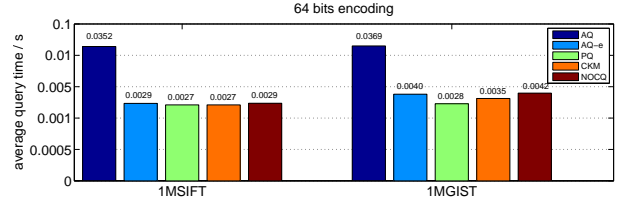}
\vspace{-.3cm}
	\caption{Average query time on $1M$SIFT and $1M$GIST. }
	\label{fig:timecost}
\end{figure}

\vspace{.1cm}
\noindent\textbf{Additive quantization and residual vector quantization.}
The idea summing several elements selected from different dictionaries
is very close to residual vector quantization or multi-stage vector quantization~\cite{GrayN98}.
The key difference is that most residual vector quantization performs sequential optimization
while composite quantization performs joint optimization.
Composite quantization is the same to the parallelly-developed  additive quantization~\cite{BabenkoK14}
with slightly different optimization technique,
which both
achieve more accurate vector approximation
and thus more accurate search results
than near-orthogonal composite quantization.
The comparison with additive quantization and its extension,
optimized tree quantization~\cite{BabenkoK15}
are shown in Table~\ref{tab:comparsionwithAQandTQ}.
The search time cost, however,
for the search under the widely-used Euclidean distance,
is much higher than that of near-orthogonal composite quantization.
Optimized tree quantization performs two times slower than Cartesian $k$-means, as observed in~\cite{BabenkoK15}.

From Figure~\ref{fig:timecost}, which shows the average query time cost including the lookup table construction cost and the linear scan search cost, it can be seen that
near-orthogonal composite quantization takes slightly more time than Cartesian $k$-means.
It can also be seen that additive quantization (AQ) takes much more time than other methods because the linear scan search cost of AQ is quadratic with respect to $M$
while that of other methods
is linear in $M$.
With encoding the square of the $L_2$ norm of the reconstructed database vector into one byte, AQ-e can achieve competitive query time but with deteriorated search accuracy.

\begin{table}[t]
	\caption{The search accuracy comparison
		for
		our approach, NOCQ,
		and additive/composite quantization (AQ)~\cite{BabenkoK14},
		and optimized tree quantization (OTQ)~\cite{BabenkoK15} on $1M$SIFT.
		AQ-e means the scheme using $1$ byte to encode the norm
		of the reconstructed vector.
		The results about AQ, AQ-e and OTQ are from the corresponding papers~\cite{BabenkoK14,BabenkoK15}.}
	\label{tab:comparsionwithAQandTQ}
\vspace{-.3cm}
	\centering
	\begin{tabular}{|c|c|ccc|}
		\hline
		\#Bits & Methods & Recall@1 & Recall@10 & Recall@100 \\
		\hline
		\hline
		\multirow{3}{*}{32} & AQ  &   0.10  &  0.37 & 0.76 \\
		& OTQ  &  0.09    & 0.32 & 0.73 \\
		& NOCQ    &  0.07 & 0.29 & 0.70 \\
		\hline
		\multirow{3}{*}{64} & AQ  &   0.31  &  0.75 & 0.97 \\
		& OTQ  &  0.32    & 0.75 & 0.97 \\
		& AQ-e  &   0.25  &  0.69 & 0.96 \\
		& NOCQ    &  0.29 & 0.72 & 0.97 \\
		\hline
	\end{tabular}
\end{table}

\begin{table}[t]
	\caption{The search accuracy comparison
		for our approach, NOCQ,  orthogonal composite quantization (OCQ)
		and Cartesian $k$-means (CKM)
		on $1M$SIFT.}
	\label{tab:dictionaryorthogonal}
	\centering
\vspace{-.3cm}
	\begin{tabular}{|c|c|ccc|}
		\hline
		\#Bits & Methods & Recall@1 & Recall@10 & Recall@100 \\
		\hline
		\hline
		\multirow{3}{*}{32} & CKM  &   0.069  &  0.277 & 0.665 \\
		& OCQ  &  0.070    & 0.284 & 0.670 \\
		& NOCQ    &  0.073 & 0.295 & 0.704 \\
		\hline
		\multirow{3}{*}{64} & CKM  &   0.245  &  0.638 & 0.945 \\
		& OCQ  &  0.247    & 0.643 & 0.942 \\
		& NOCQ    &  0.290 & 0.715 & 0.970 \\
		\hline
		\multirow{3}{*}{128} & CKM  &   0.466  &  0.903 & 0.998 \\
		& OCQ  &  0.469    & 0.905 & 0.997 \\
		& NOCQ    &  0.540 & 0.957 & 1 \\
		\hline
	\end{tabular}
\end{table}

\vspace{.1cm}
\noindent\textbf{Product quantization and Cartesian $k$-means.}
Product quantization~\cite{JegouDS11}
decomposes the space into $M$ low dimensional subspaces
and quantizes each subspace separately.
A vector $\mathbf{x}$ is
decomposed into $M$ subvectors,
$\{\mathbf{x}^1, \cdots, \mathbf{x}^M\}$.
Let the quantization dictionaries
over the $M$ subspaces
be $\mathcal{C}_1, \mathcal{C}_2, \cdots, \mathcal{C}_M$
with $\mathcal{C}_m$ being
a set of centers $\{\mathbf{c}_{m1}, \cdots, \mathbf{c}_{mK}\}$.
A vector $\mathbf{x}$ is represented
by the concatenation of $M$ centers,
$[\mathbf{c}_{1{k_1^*}}^T, \mathbf{c}_{2{k_2^*}}^T,
\cdots , \mathbf{c}_{m{k_m^*}}^T,
\cdots,  \mathbf{c}_{M{k_M^*}}^T]^T$,
where $\mathbf{c}_{m{k_m^*}}$ is the one nearest
to $\mathbf{x}^m$
in the $m$th quantization dictionary.

Rewrite each center $\mathbf{c}_{mk}$
as a $D$-dimensional vector $\tilde{\mathbf{c}}_{mk}$
so that
$\tilde{\mathbf{c}}_{mk} = [\boldsymbol{0}^T, \cdots,(\mathbf{c}_{mk})^T, \cdots, \boldsymbol{0}^T]^T$,
i.e., all entries are zero
except that the subvector corresponding to the $m$th subspace
is equal to $\mathbf{c}_{mk}$.
The approximation of a vector $\mathbf{x}$
using the concatenation
$\mathbf{x} = [\mathbf{c}_{1{k_1^*}}^T,\mathbf{c}_{2{k_2^*c}}^T, \cdots, \mathbf{c}_{M{k_M^*}}^T]^T$
is then equivalent to
the composition
$\mathbf{x} = \sum_{m=1}^M \tilde{\mathbf{c}}_{m{k_m^*}}$.
Similarly,
it can also be shown that
there is a same equivalence
in Cartesian $k$-means~\cite{NorouziF13}.

The above analysis indicates that
both product quantization
and Cartesian $k$-means
can be regarded as a constrained version
of composition quantization,
with the orthogonality constraint:
$\mathbf{C}_i^T\mathbf{C}_j = \mathbf{0}$, $i \neq j$,
which guarantees
that the near-orthogonality constraint in our approach holds.
In addition,
unlike product quantization and Cartesian $k$-means
in which each dictionary (subspace)
is formed by $D/M$ dimensions,
near-orthogonal composite quantization when $\epsilon = 0$,
i.e., orthogonal composite quantization is able to automatically decide
how many dimensions belong to one dictionary.

One may have a question:
Is near-orthogonal composite quantization
equivalent to product quantization
over linearly transformed vectors and
with proper choices of
the number of dimensions in each subspace (dictionary)?
The answer is NO.
NOCQ does not transform data vectors,
but relaxes the space spanned by the dictionary
with a near-orthogonality constraint
instead of orthogonality constraint.

\begin{table}[t]
	\caption{The quantization error of composite quantization and near-orthogonal composite quantization on $1M$SIFT and $1M$GIST with various number of bits.}
	\label{tab:distortioneerorCQandNOCQ}
\vspace{-.3cm}
	\centering
	\begin{tabular}{|c|c|cc|}
		\hline
		\#Bits & Methods & $1M$SIFT  & $1M$GIST \\
		\hline
		\hline
		\multirow{2}{*}{32} & CQ  &   $2.94\times10^{10}$  &  $8.19\times 10^5$   \\
		& NOCQ &  $3.09\times10^{10}$    & $8.46\times 10^5$  \\
		\hline
		\multirow{2}{*}{64} & CQ  &   $1.54\times10^{10}$  &  $6.39\times 10^5$  \\
		& NOCQ  &  $1.65\times10^{10}$    & $6.63\times 10^5$   \\
		\hline
		\multirow{2}{*}{128} & CQ  &   $6.09\times10^9$  &  $4.69\times 10^5$   \\
		& NOCQ  &  $6.66\times10^9$    & $4.86\times 10^5$   \\
		\hline
	\end{tabular}
\end{table}

\vspace{.1cm}
\noindent\textbf{Near-orthogonality constraint.}
The near-orthogonality constraint (\ref{eqn:nearorthogonality})
is introduced to accelerate the search process.
Table~\ref{tab:distortioneerorCQandNOCQ}
shows that introducing this constraint
results in higher quantization errors.
One question comes:
can we have alternative search process acceleration scheme
if we learn the dictionaries using composite quantization
without such a constraint to improve the approximation quality?
We consider two possible schemes:
when computing the distances in the search stage
by (1) simply discarding the third term
$\sum_{i=1}^M \sum_{j=1, j\neq i}^M \mathbf{c}_{ik_i}^T \mathbf{c}_{jk_j}$ in Equation~(\ref{eqn:distancetransform})
or (2)
precomputing $\sum_{i=1}^M \sum_{j=1, j\neq i}^M \mathbf{c}_{ik_i}^T \mathbf{c}_{jk_j}$
and storing it as an extra code.

\begin{table}[t]
	\caption{Comparing the performances
		of alternative schemes without the near-orthogonality constraint:
		CQ with extra bytes to storing the third term (CQ),
		CQ with discarding the third term in the search stage (CQ-d),
		CQ with counting the extra bytes into the total encoding cost (CQ-c),
		and our approach (NOCQ).
		In the columon \#Bits, the first number
		indicates the number of bits for encoding the vector,
		and the second number indicates
		the number of bits for encoding the third term in Equation~(\ref{eqn:distancetransform}).}
	\label{tab:IgnoreTerm}
\vspace{-.3cm}
	\centering
\resizebox{1\linewidth}{!}{
	\begin{tabular}{|c|l|l|ccc|}
		\hline
		Dataset & \#Bits & Methods & Recall@1 & Recall@10 & Recall@100 \\
		\hline
		\hline
		\multirow{9}{*}{$1M$SIFT} &  32, 32 & CQ   &  0.110  &  0.374 & 0.775 \\
		& 32, 0 & CQ-d   &   0.028   &  0.129  & 0.417  \\
		& 32, 0 & NOCQ  &  0.073    &  0.295 & 0.704 \\
		\cline{2-6}
		&  64, 32 & CQ   &   0.337  &  0.769 & 0.984 \\
		& 64, 0 & CQ-d  &   0.117  &  0.388  & 0.753  \\
		& 32, 32  & CQ-c   &  0.110  &  0.374 & 0.775 \\
		& 64, 0 & NOCQ  &  0.290    & 0.715 & 0.970 \\
		\cline{2-6}
		& 128, 32 & CQ   &  0.566  &  0.968 & 1.000 \\
		& 128, 0 & CQ-d   &   0.195  &  0.520  & 0.856  \\
		& 96, 32 & CQ-c & 0.463 & 0.909 & 0.999 \\
		& 128, 0 & NOCQ  &  0.540    & 0.957 & 1.000 \\
		
		\hline
		\hline
		\multirow{9}{*}{$1M$GIST} &  32, 32 & CQ   &  0.063 &  0.181 & 0.454 \\
		& 32, 0 & CQ-d   &   0  & 0.002 & 0.006   \\
		& 32, 0 & NOCQ  &  0.064    & 0.173 & 0.416   \\
		\cline{2-6}
		& 64, 32 & CQ   &  0.131  & 0.380 & 0.742   \\
		& 64, 0 & CQ-d   &   0  &  0.003  & 0.007   \\
		&  32, 32 & CQ-c   &  0.063 &  0.181 & 0.454 \\
		& 64, 0 & NOCQ  &  0.140    & 0.378 & 0.730   \\
		\cline{2-6}
		&  128, 32 & CQ   &   0.232  &  0.628 & 0.948   \\
		& 128, 0 & CQ-d   &  0.027  &  0.080  & 0.226    \\
		& 96, 32 & CQ-c & 0.205 &  0.524 &  0.876  \\
		& 128, 0 & NOCQ  &  0.215    & 0.631 & 0.930    \\
		\hline
	\end{tabular}
}
\end{table}

Table~\ref{tab:IgnoreTerm}
shows the results of the two alternative schemes.
We use our optimization algorithm
to learn the CQ model,
and the search performance of CQ
is obtained
by using extra $4$ bytes to store the value of the third term
or equivalently taking higher search cost
to compute the third term
in the search stage.
We have two observations:
(1) Discarding the third term, denoted by CQ-d,
leads to dramatic search quality reduction;
(2) As expected,
our approach NOCQ gets lower search performance
than CQ, but higher than CQ-d (the third term discarded).
It should be noted that CQ uses extra $4$ bytes or much higher search cost.

For fair comparison,
we report the performance of CQ
with the total code size
consisting of both
the bytes encoding the data vector
and the extra $4$ bytes ($32$ bits) encoding the third term.
We can see that our approach performs much better.

\vspace{.1cm}
\noindent\textbf{From the Euclidean distance to the inner product.}
The Euclidean distance between the query $\mathbf{q}$
and the vector $\mathbf{x}$
can be transformed
to an inner product form,
\begin{align}
\|\mathbf{q} - \mathbf{x}\|_2^2 =& \|\mathbf{q}\|_2^2-2\mathbf{q}^\top\mathbf{x} + \|\mathbf{x}\|_2^2
= \|\mathbf{q}\|_2^2 + [\mathbf{q}^\top~1] \left[ \begin{array}{c}
-2\mathbf{x}\\
\|\mathbf{x}\|_2^2\\
\end{array}
\right]. \nonumber
\end{align}
This suggests that
in the search stage
computing the second term
is enough
and thus we can quantize the augmented data vector,
$\tilde{\mathbf{x}}=\left[ \begin{array}{c}
-2\mathbf{x}\\
\|\mathbf{x}\|_2^2\\
\end{array}\right]$.
It is shown in~\cite{DuW14}
that the inner product computation
can also benefit from table lookup
and thus is efficient.
We report the performances of quantizing
the augmented vector with the CQ approach,
which performs better than PQ and CKM.
The results are given in Table~\ref{tab:augmentedInnerProduct}.
We can see that such a scheme performs poorly.
It is because the scale of the last element $\|\mathbf{x}\|_2^2$
is very different from the scales
of other elements, and
thus the optimization is hard.
The possible solution of quantizing $\mathbf{x}$
and $\|\mathbf{x}\|_2^2$ separately
also does not lead to much improvement,
implying this is not a good direction.

The Euclidean distance between a query and the approximated vector
can also be transformed to an inner product form,
$\|\mathbf{q} - \bar{\mathbf{x}}\|_2^2 = \|\mathbf{q}\|_2^2-2\mathbf{q}^\top\bar{\mathbf{x}} + \|\bar{\mathbf{x}}\|_2^2$.
Thus, similar to quantizing the third term
in Equation~(\ref{eqn:distancetransform}),
we can quantize the square of the $L_2$ norm, $\|\bar{\mathbf{x}}\|_2^2$.
The two quantization ways
empirically show almost the same performance.
The reason might be that the two terms are very close:
$\|\bar{\mathbf{x}}\|_2^2 = \sum_{i=1}^M \sum_{j=1, j\neq i}^M \mathbf{c}_{ik_i}^T \mathbf{c}_{jk_j}
+ \sum_{i=1}^M \mathbf{c}_{ik_i}^T \mathbf{c}_{ik_i}$.

\begin{table}[t]
	\caption{Comparing the approach
		with quantizing the augmented vectors (QAV)
		and our approach NOCQ. }
	\label{tab:augmentedInnerProduct}
	\centering
\vspace{-.3cm}
\resizebox{1\linewidth}{!}{
	\begin{tabular}{|c|c|l|ccc|}
		\hline
		Dataset & \#Bits & Methods & Recall@1 & Recall@10 & Recall@100  \\
		\hline
		\hline
		\multirow{9}{*}{$1M$SIFT} &  \multirow{2}{*}{32} & QAV &  0.018    &  0.099 & 0.339 \\
		& & NOCQ & 0.073 & 0.295 & 0.704\\
		\cline{2-6}
		&  \multirow{2}{*}{64} &  QAV &   0.103    & 0.355 & 0.731 \\
		& & NOCQ & 0.290 & 0.715 & 0.970 \\
		\cline{2-6}
		&  \multirow{2}{*}{128} & QAV &   0.275    & 0.704 & 0.966 \\
		& & NOCQ & 0.540 & 0.957 & 1.000\\
		
		\hline
		\hline
		\multirow{9}{*}{$1M$GIST} &  \multirow{2}{*}{32}  & QAV &  0.016    & 0.102 & 0.304 \\
		& & NOCQ &  0.064    & 0.173 & 0.416 \\
		\cline{2-6}
		&  \multirow{2}{*}{64} & QAV &  0.059    & 0.248 & 0.581  \\
		& & NOCQ & 0.140 & 0.378 & 0.730\\
		\cline{2-6}
		&  \multirow{2}{*}{128} & QAV &  0.164    & 0.457 & 0.842   \\
		& & NOCQ & 0.215 & 0.631 & 0.930\\
		\hline
	\end{tabular}
}
\end{table}

\section{Optimization}
The formulation of near-orthogonal composite quantization
is given as follows,
\begin{align}\small
\min_{\{\mathbf{C}_m\},\{\mathbf{y}_n\}, \epsilon } ~&~\sum\nolimits_{n=1}^N\|\mathbf{x}_n -  [\mathbf{C}_1, \mathbf{C}_2, \cdots, \mathbf{C}_M]\mathbf{y}_{n}\|_2^2 \label{eqn:nearorthogonalitycompositequantization} \\
\operatorname{s.t.}
~&~ \mathbf{y}_n  = [\mathbf{y}_{n1}^\top,\mathbf{y}_{n2}^\top,\cdots,\mathbf{y}_{nM}^\top]^\top \nonumber \\
~&~ \mathbf{y}_{nm} \in \{0, 1\}^K,~\|\mathbf{y}_{nm}\|_1 = 1 \nonumber \\
~&~\sum\nolimits_{i=1}^M \sum\nolimits_{j=1, j\neq i}^M  \mathbf{y}_{ni}^\top \mathbf{C}_{i}^\top\mathbf{C}_{j}\mathbf{y}_{nj} = \epsilon \label{eqn:nearorthogonalityconstraint}\\
~&~ n=1,2,\cdots,N, m=1,2, \cdots, M.\nonumber
\end{align}
Here, ${\mathbf{C}_m}$ is a matrix of size $D\times K$,
and each column corresponds to an element
of the $m$th dictionary $\mathcal{C}_m$.
$\mathbf{y}_n$ is the composition vector,
and its subvector
$\mathbf{y}_{nm}$
is an indicator vector
with only one entry being $1$
and all others being $0$,
showing which element is selected
from the $m$th dictionary
to compose vector $\mathbf{x}_n$.

The problem formulated
above is a mixed-binary-integer program,
which consists of three groups of unknown variables:
dictionaries $\{\mathbf{C}_m\}$,
composition vectors $\{\mathbf{y}_n\}$,
and $\epsilon$.
In addition to the binary-integer constraint over $\{\mathbf{y}_n\}$,
there are near-orthogonality constraints over $\{\mathbf{C}_m\}$
given in~(\ref{eqn:nearorthogonalityconstraint}).

\subsection{Algorithm}
To handle the near-orthogonality constraint,
we propose to adopt the quadratic penalty method,
and add a penalty function that measures
the violation of the quadratic equality constraints
into the objective function,
resulting in the following objective function,
\begin{align}\small
\label{eqn:penaltyfunction}
& \phi({\{\mathbf{C}_m\}}, \{\mathbf{y}_n\}, \epsilon)
= \sum\nolimits_{n=1}^N\|\mathbf{x}_n -  \mathbf{C}\mathbf{y}_{n}\|_2^2
\\ &+ \mu \sum\nolimits_{n=1}^N(\sum\nolimits_{i \neq j}^M \mathbf{y}_{ni}^\top \mathbf{C}_{i}^\top\mathbf{C}_{j}\mathbf{y}_{nj} - \epsilon)^2,
\end{align}
where $\mathbf{C}=[\mathbf{C}_1,\mathbf{C}_2,\cdots,\mathbf{C}_M]$
and $\mu$ is the penalty parameter.
The main technique we used
is the alternative optimization method,
where each step updates one variable while fixing the others.

\vspace{.1cm}
\noindent\textbf{Update $\{\mathbf{y}_n\}$.}
It can be easily seen that
$\mathbf{y}_n$,
the composition indicator of a vector $\mathbf{x}_n$,
given $\{\mathbf{C}_m\}$ fixed,
is independent to all the other vectors
$\{\mathbf{y}_t\}_{t \neq n}$.
Then the optimization problem~(\ref{eqn:penaltyfunction}) is decomposed
into $N$ subproblems,
\begin{align}\small
\min_{\mathbf{y}_n}\|\mathbf{x}_n - \mathbf{C}\mathbf{y}_n\|_2^2
+ \mu(\sum\nolimits_{i \neq j}^M \mathbf{y}_{ni}^\top \mathbf{C}_{i}^\top\mathbf{C}_{j}\mathbf{y}_{nj} - \epsilon)^2,
\label{eqn:updateb}
\end{align}
where there are three constraints:
$\mathbf{y}_n$ is a binary vector,
$\|\mathbf{y}_{nm}\|_1=1$,
and
$\mathbf{y}_n  = [\mathbf{y}_{n1}^\top,\mathbf{y}_{n2}^\top,\cdots,\mathbf{y}_{nM}^\top]^\top$.
Generally, this optimization problem is NP-hard.
We notice that
the problem is essentially
a high-order MRF (Markov random field) problem.
We again use the alternative optimization technique
like the iterated conditional modes algorithm that is widely used to solve MRFs,
and solve the $M$ subvectors $\{\mathbf{y}_{nm}\}$ alternatively.
Given $\{\mathbf{y}_{nl}\}_{l \neq m}$ fixed,
we update $\mathbf{y}_{nm}$
by exhaustively checking
all the elements in the dictionary $\mathcal{C}_m$,
finding the element such that
the objective value is minimized,
and accordingly setting the corresponding entry of $\mathbf{y}_{nm}$ to be $1$
and all the others to be $0$.
This process is iterated several ($1$ in our implementation) times.
The optimization is similar
for solving composite quantization
and orthogonal composite quantization.

\vspace{.1cm}
\noindent\textbf{Update $\epsilon$.}
With $\{\mathbf{y}_n\}$ and $\mathbf{C}$ fixed,
it can be shown that the optimal solution is
\begin{align}
\epsilon = \frac 1 N \sum\nolimits_{n=1}^N(\sum\nolimits_{m' \neq m}^M \mathbf{y}_{nm}^{\top}\mathbf{C}_m^{\top} \mathbf{C}_{m'}\mathbf{y}_{nm'}).
\label{eqn:updateepsilon}
\end{align}

\vspace{.1cm}
\noindent\textbf{Update $\mathbf{C}$.} \label{sec:updatec}
Fixing $\{\mathbf{y}_n\}$
and $\epsilon$,
the problem is an unconstrained nonlinear optimization problem
with respect to $\mathbf{C}$.
There are many algorithms for such a problem.
We choose the quasi-Newton algorithm
and specifically the L-BFGS algorithm,
the limited-memory version of the Broyden-Fletcher-Goldfarb-Shanno (BFGS) algorithm.
It only needs a few vectors
to represent the approximation of the Hessian matrix
instead of storing the full Hessian matrix as done in the BFGS algorithm.
We adopt the publicly available implementation of L-BFGS\footnote{http://www.ece.northwestern.edu/\~{}nocedal/lbfgs.html}.
The partial-derivative
with respect to $\mathbf{C}_m$,
the input to the L-BFGS solver,
is computed as follows,
\begin{align}\small
&~\frac {\partial} {\partial \mathbf{C}_m} \phi(\{\mathbf{C}_m\}, \{\mathbf{y}_n\}, \epsilon)  \nonumber \\
=&~\sum\nolimits_{n=1}^N [2(\sum\nolimits_{l=1}^M \mathbf{C}_l \mathbf{y}_{nl} -\mathbf{x}_n) \mathbf{y}_{nm}^\top + \\
&
4\mu (\sum\nolimits_{i \neq j}^M\mathbf{y}_{ni}^\top\mathbf{C}_i^\top\mathbf{C}_j\mathbf{y}_{nj} - \epsilon  )
(\sum\nolimits_{l=1, l \neq m}^M \mathbf{C}_l\mathbf{y}_{nl})\mathbf{y}_{nm}^\top].
\label{eqn:updatec}
\end{align}

In the case of composite quantization,
there is a closed-form solution:
$\mathbf{C} = \mathbf{X}\mathbf{Y}^\top(\mathbf{Y}\mathbf{Y}^\top)^{-1}$,
where $\mathbf{X}$ is a matrix
with each column corresponding to a database vector,
and $\mathbf{Y}$ is a matrix composed of the composition vectors
of all the database vectors,
$\mathbf{Y} = [\mathbf{y}_1,\mathbf{y}_2,\cdots,\mathbf{y}_n]$.
In our implementation,
we also adopt the iterative solution for composite quantization
because we found that
the iterative solution performs better.
The optimization for orthogonal composite quantization
is similar to that for near-orthogonal composite quantization
with slight difference
on the penalty part which is $\mu \sum_{i=1}^M\sum_{j=1, j\neq 1}^M \|\mathbf{C}_i^\top \mathbf{C}_j\|_F^2$.

\subsection{Implementation Details}
The proposed algorithm is warm-started by using the solution
of product quantization. There is a penalty parameter, $\mu$, for the near-orthogonality constraint
and the orthogonality constraint.
Usually the penalty method needs to solve a series of unconstrained problems
by increasing the penalty parameter $\mu$ into infinity
to make the constraint completely satisfied.
In our case,
we find that
the near-orthogonality term is not necessarily to be exactly constant
and the search performance is still satisfactory
if the deviation of the near-orthogonality term from a constant
is relatively small compared with
the quantization error.
Therefore,
our algorithm instead relaxes this constraint
and selects the parameter $\mu$
via validation.
The validation dataset is a subset of the database
(selecting a subset is only for validation efficiency,
and it is fine that the validation set is a subset of the learning set
as the validation criterion is not the objective function value
but the search performance).
The best parameter $\mu$ is chosen
so that
the average search performance
by regarding
the validation vectors as queries
and finding $\{5, 10, 15, \cdots, 100\}$ nearest neighbors
from all the database vectors
is the best.

\subsection{Analysis}

\noindent\textbf{Complexity.}
We present the time complexity of each iteration.
At the beginning of each iteration,
we first compute inner product tables,
$\{\mathbf{c}_{ir}^\top\mathbf{c}_{js}|{i \neq j}, r, s = 1,2, \cdots, K\}$,
between the dictionary elements,
taking $O(M^2K^2D)$,
so that computing $\mathbf{y}^\top_{ni}\mathbf{C}^\top_i\mathbf{C}_j\mathbf{y}_{nj}$
can be completed by one table lookup.
The time complexities of the three updates are
given as follows.
\begin{itemize}
	\item It takes $O(MKDT_y)$ with $T_y$ being the number of iterations
	($=1$ in our implementation achieving satisfactory results)
	to update $\mathbf{y}_n$,
	i.e.,
	optimize the objective function in~(\ref{eqn:updateb}),
	and thus the time complexity of updating $\{\mathbf{y}_n\}$ is $O(NMKDT_y)$.
	
	\item It takes $O(NM^2)$
	to update $\epsilon$,
	which can be easily seen from Equation~(\ref{eqn:updateepsilon}).
	
	\item
	The main cost of updating $\{\mathbf{C}_m\}$
	in near-orthogonal composite quantization
	lies in computing the partial derivatives
	and the objective function value
	that are necessary in L-BFGS.
	For clarity, we drop the complexity terms
	that are independent of $N$
	and can be neglected for a large $N$.
	Then,
	the time complexity for updating $\{\mathbf{C}_m\}$ is
	$O(MNDT_{l}T_{c})$
	with $T_{c}$
	being the number of iterations
	($10$ in our implementation in average)
	and $T_{l}$ (set to $5$ in our implementation)
	being the number of line searches in L-BFGS.

\end{itemize}

\begin{figure}[t]
	\centering
	\includegraphics[width=0.9\linewidth]{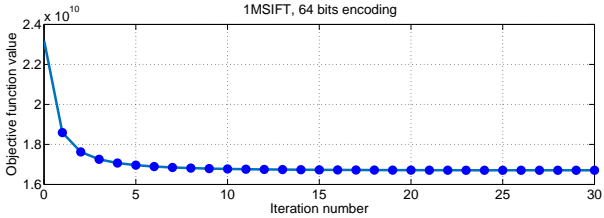}
\vspace{-.3cm}
	\caption{Convergence curve
		of our algorithm.
		The vertical axis represents the objective function value of
		Equation~(\ref{eqn:penaltyfunction})
		and the horizontal axis corresponds to the number of iterations.
		The curve is obtained
		from the result over a representative dataset
		$1M$SIFT with $64$ bits.}
	\label{fig:convergence}
\end{figure}

\noindent\textbf{Convergence.}
The objective function value
at each iteration in the algorithm
always weakly decreases.
It can be validated
that the objective function value is lower-bounded
(not smaller than $0$).
The two points indicate
the convergence of our algorithm.
The theoretic analysis
of the rate of convergence is not easy,
while the empirical results show that
the algorithm takes a few iterations
to converge.
Figure~\ref{fig:convergence} shows an example convergence curve
for near-orthogonal composite quantization.

\begin{table}[t]
	\caption{The description of the datasets.}
	\label{tab:sevendatasets}
\vspace{-.3cm}
\centering
			\begin{tabular}{|lrrc|}
				\hline
				& Base set & Query set & Dim\\
				\hline
				MNIST~\cite{LeCunBBH01}         & $60,000$ & $10,000$   & $784$ \\
				LabelMe$22K$~\cite{RussellTMF08}    & $20,019$ & $2,000$    & $512$ \\
				$1M$SIFT~\cite{JegouDS11}      & $1,000,000$ &   $10,000$   & $128$ \\
				$1M$GIST~\cite{JegouDS11}      & $1,000,000$ &   $1,000$   & $960$ \\
				$1B$SIFT~\cite{JegouTDA11}      & $1,000,000,000$ &   $10,000$   & $128$ \\
				$1M$CNN~\cite{ILSVRC15}      & $1,281,167$ &   $100,000$   & $4096$ \\
				\hline
			\end{tabular}
\centering
\end{table}

\section{Experiments}
\subsection{Setup}
\noindent\textbf{Datasets.}
We demonstrate the performance over six datasets
(with scale ranged from small to large):
MNIST\footnote{http://yann.lecun.com/exdb/mnist/}~\cite{LeCunBBH01},
$784D$ grayscale images of handwritten digits;
LabelMe$22K$~\cite{RussellTMF08},
a corpus of images expressed as $512D$ GIST descriptors;
$1M$SIFT~\cite{JegouDS11},
consisting of $1M$ $128D$ SIFT features as base vectors, $100K$ learning vectors
and $10K$ queries;
$1M$GIST~\cite{JegouDS11},
containing $1M$ $960D$ global GIST descriptors as base vectors, $500K$ learning vectors and $1K$ queries;
$1M$CNN~\cite{ILSVRC15},
with $1,281,167$ $4096D$ convolution neural network (CNN) features
as base vectors ,
$100,000$ CNN features as queries,
extracted over the ImageNet training and test images
through AlexNet~\cite{KrizhevskySH12};
and $1B$SIFT~\cite{JegouTDA11},
composed of $1B$ SIFT features as base vectors, $100M$ learning vectors and $10K$ queries.
The details of the datasets are presented in Table~\ref{tab:sevendatasets}.

\vspace{.1cm}
\noindent\textbf{Compared methods.}
We compare our approach,
near-orthogonality composite quantization (NOCQ)
with several state-of-the-art methods:
product quantization (PQ)~\cite{JegouDS11},
and Cartesian $k$-means (CKM)~\cite{NorouziF13}.
It is already shown that
PQ and CKM achieve better search accuracy than hashing algorithms
with the same code length and comparable search efficiency.
Thus, we report one result from
a representative hashing algorithm,
iterative quantization (ITQ)~\cite{GongL11}.
All the results were obtained
with the implementations
generously provided by their respective authors.
Following~\cite{JegouDS11},
we use the structured ordering for $1M$GIST
and the natural ordering for $1M$SIFT and $1B$SIFT
to get the best performance for PQ.
We choose $K=256$ as the dictionary size
which is an attractive choice
because the resulting distance lookup tables
are small
and each subindex fits into one byte~\cite{JegouDS11, NorouziF13}.

\vspace{.1cm}
\noindent\textbf{Evaluation.}
To find ANNs,
all the algorithms use asymmetric distance
(i.e., query is not encoded)
unless otherwise stated.
To compare a query with a database vector,
PQ, CKM and NOCQ conduct a few distance table lookups
and additions,
and ITQ uses asymmetric hamming distance for better search accuracy
proposed in~\cite{GordoP11}.
PQ, CKM, and NOCQ takes the same time for linear scan.
Their costs of computing the distance lookup table
are slightly different
and are negligible
when handling a large scale dataset,
except the scale is small which is handled
in our work~\cite{ZhangQTW15}.
For instance,
the cost of computing the distance lookup table
in our approach takes around $4\%$
of the cost of linear scan on $1M$SIFT.
Figure~\ref{fig:timecost}
shows the average query times
on $1M$SIFT and $1M$GIST,
which shows the time costs are similar.


The search quality is evaluated
using two measures:
recall$@R$
and mean average precision (MAP).
Recall$@R$ is defined as follows:
for each query, we retrieve its $R$ nearest items and compute the ratio of $R$ to $T$,
i.e., the fraction of $T$ ground-truth nearest neighbors are found in the retrieved $R$ items.
The average recall score over all the queries
is used as the measure.
The ground-truth nearest neighbors
are computed over the original features
using linear scan.
In the experiments,
we report the performance
with $T$ being $1$, $10$, and $50$.
The MAP score is reported
by regarding the $100$ nearest ground-truth neighbors
as relevant answers to the query.
The average precision for a query is computed as
$\sum_{t=1}^N P(t)\Delta(t)$,
where $P(t)$ is the precision at cut-off $t$ in the ranked list
and $\Delta(t)$ is the change
in recall from items $t-1$ to $t$.
We report the mean of average precisions over all the queries under different code lengths.

\begin{figure*}[t]
	\centering
	(a)~\includegraphics[width=.45\linewidth, clip]{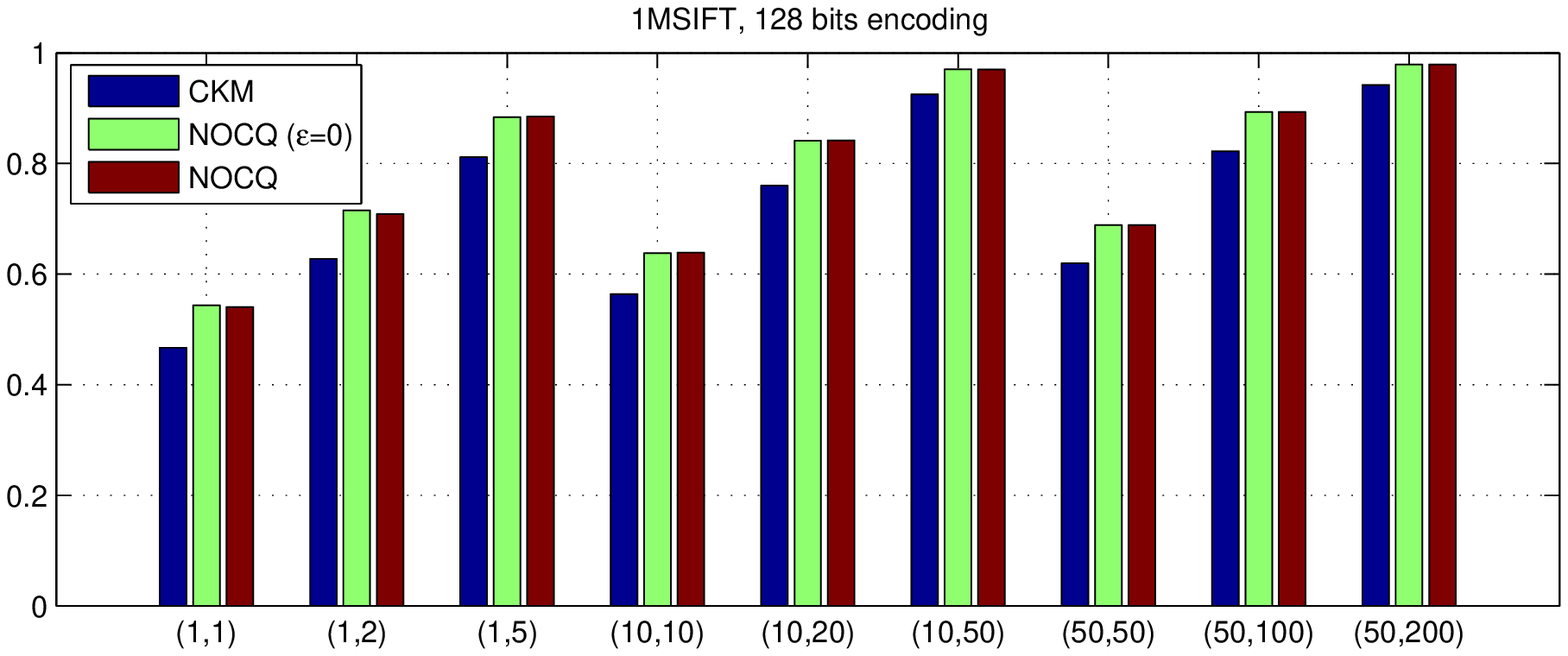}~~~
	(b)~\includegraphics[width=.45\linewidth, clip]{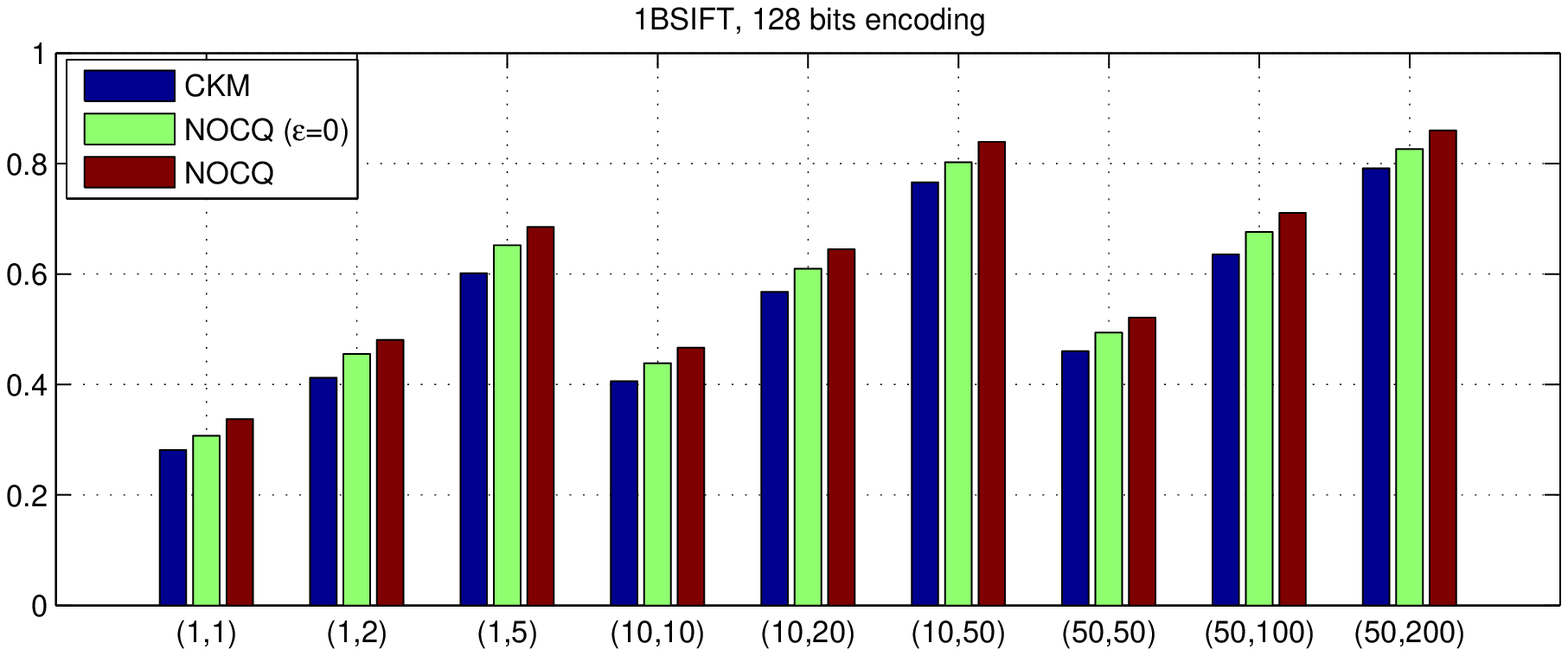}
\vspace{-.3cm}
	\caption{Illustrating the effect of $\epsilon$ on (a) $1M$SIFT
		and (b) $1B$SIFT.
		$(T,R)$ means recall@$R$
		when searching for $T$ nearest neighbors. }
	\label{fig:epsilon0}
\end{figure*}

\begin{figure}[t]
	\centering
	\includegraphics[width=.9\linewidth, clip]{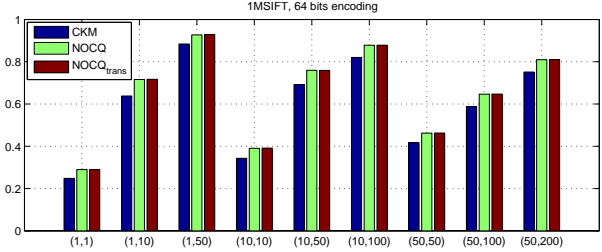}
\vspace{-.3cm}
	\caption{Illustrating the effect of translation on $1M$SIFT.
		$(T,R)$ means recall@$R$
		when searching for $T$ nearest neighbors. }
	\label{fig:translation}
\end{figure}

\subsection{Empirical analysis}

\noindent\textbf{The effect of $\epsilon$.}
The near-orthogonality variable $\epsilon$
in our approach
is learnt from the reference base.
Alternatively, one can simply set it to be zero,
$\epsilon = 0$,
indicating
that the dictionaries are mutually orthogonal
like splitting the spaces into subspaces as done in product quantization and Cartesian $k$-means.
The average quantization error in the case of learning $\epsilon$
potentially can be smaller than
that in the case of letting $\epsilon = 0$
as learning $\epsilon$ is more flexible,
and thus the search performance with learnt $\epsilon$
can be better.
The experimental results over
the $1M$SIFT and $1B$SIFT dataset
under the two schemes,
shown in Figure~\ref{fig:epsilon0}.
It can be seen that
the performance when $\epsilon$ is not limited to be zero
over $1M$SIFT is similar
but much better over $1B$SIFT.

\begin{figure*}[t]
\footnotesize
	\centering
	\includegraphics[width = .45\linewidth, clip]{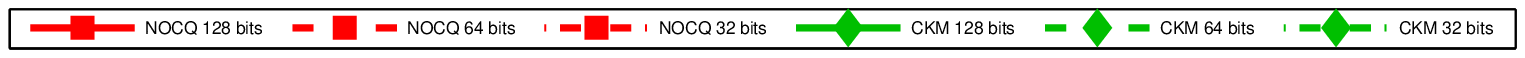}
	~~~\includegraphics[width = .45\linewidth, clip]{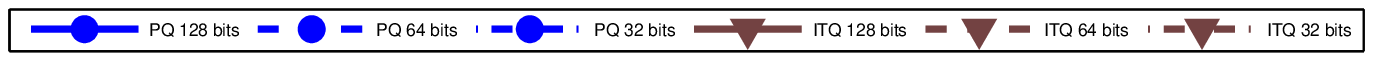}\\
	(a)~\includegraphics[width=.22\linewidth, clip]{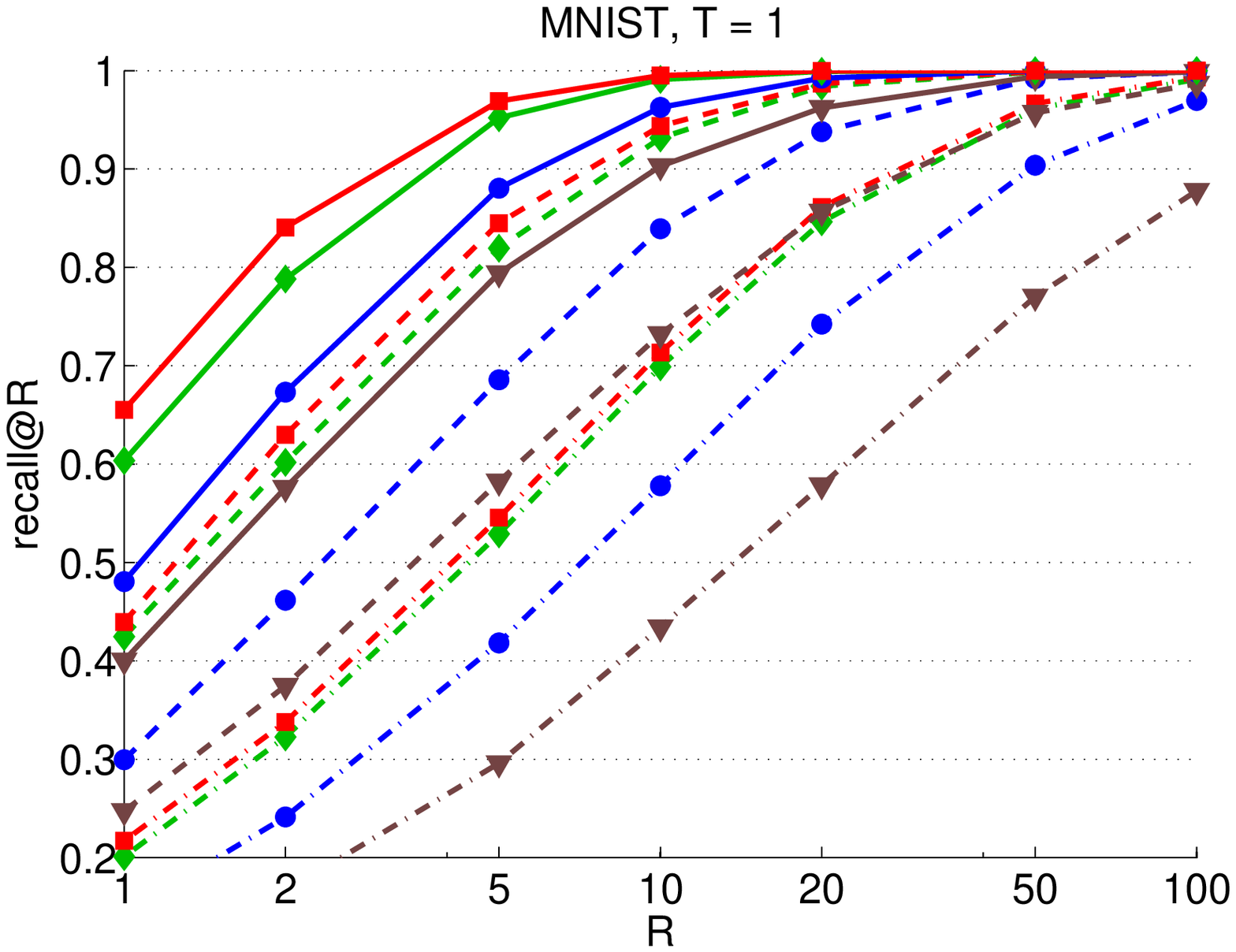}~~
	\includegraphics[width=.22\linewidth, clip]{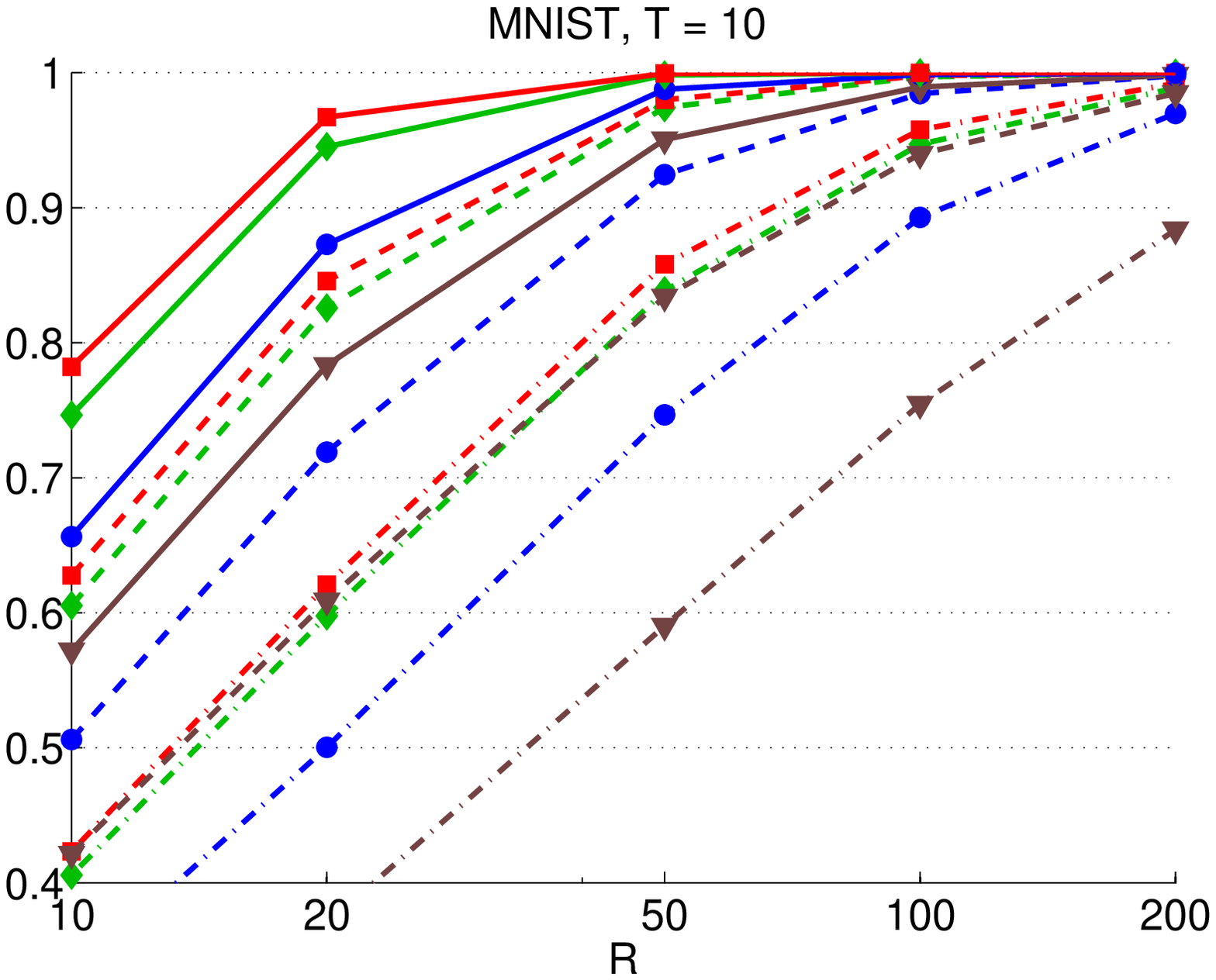}~~~
	(b)~\includegraphics[width=.22\linewidth, clip]{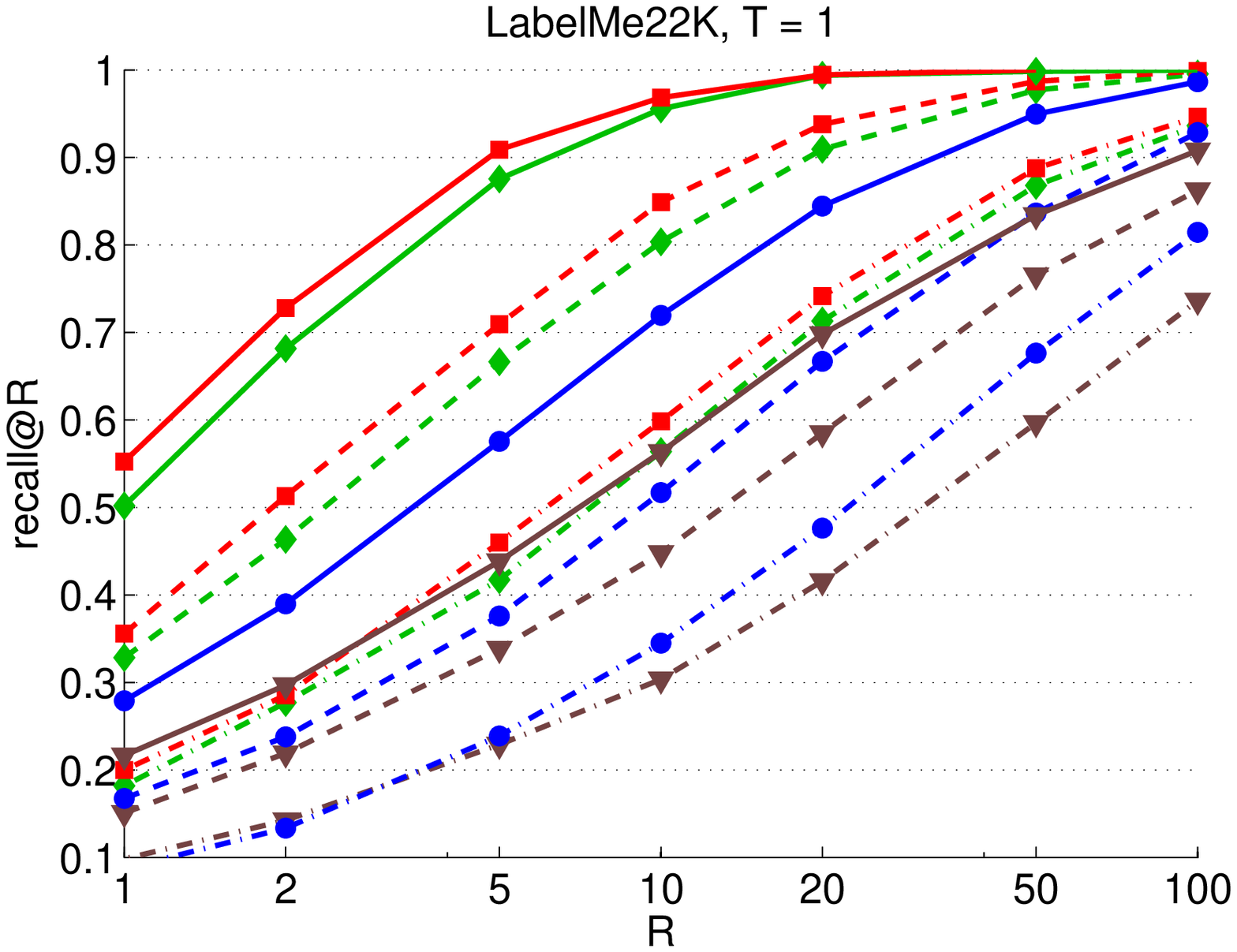}~~
	\includegraphics[width=.22\linewidth, clip]{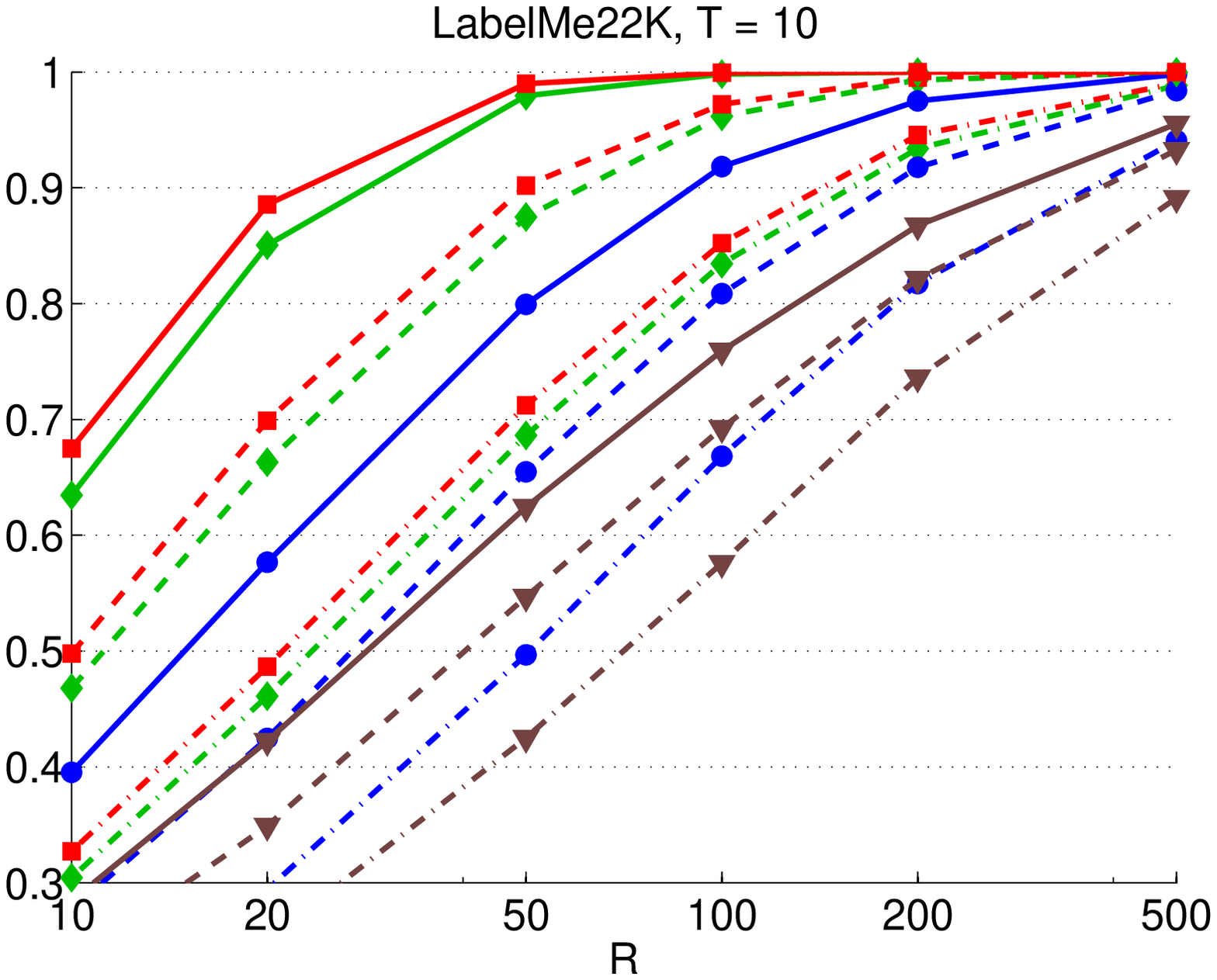}~~~
\vspace{-.3cm}
	\caption{The performance for different algorithms
		on (a) MNIST
		and (b) LabelMe$22K$
		for searching various numbers of ground truth nearest neighbors
		($T=1, 10$).}
	\label{fig:resultsofothers}
\end{figure*}

\begin{figure}[t]
	\centering
\footnotesize
	\includegraphics[width = .9\linewidth, clip]{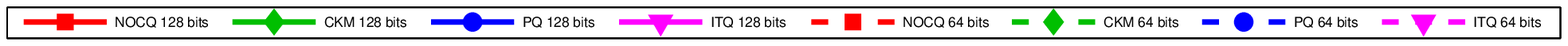}\\
	\includegraphics[width=.44\linewidth, clip]{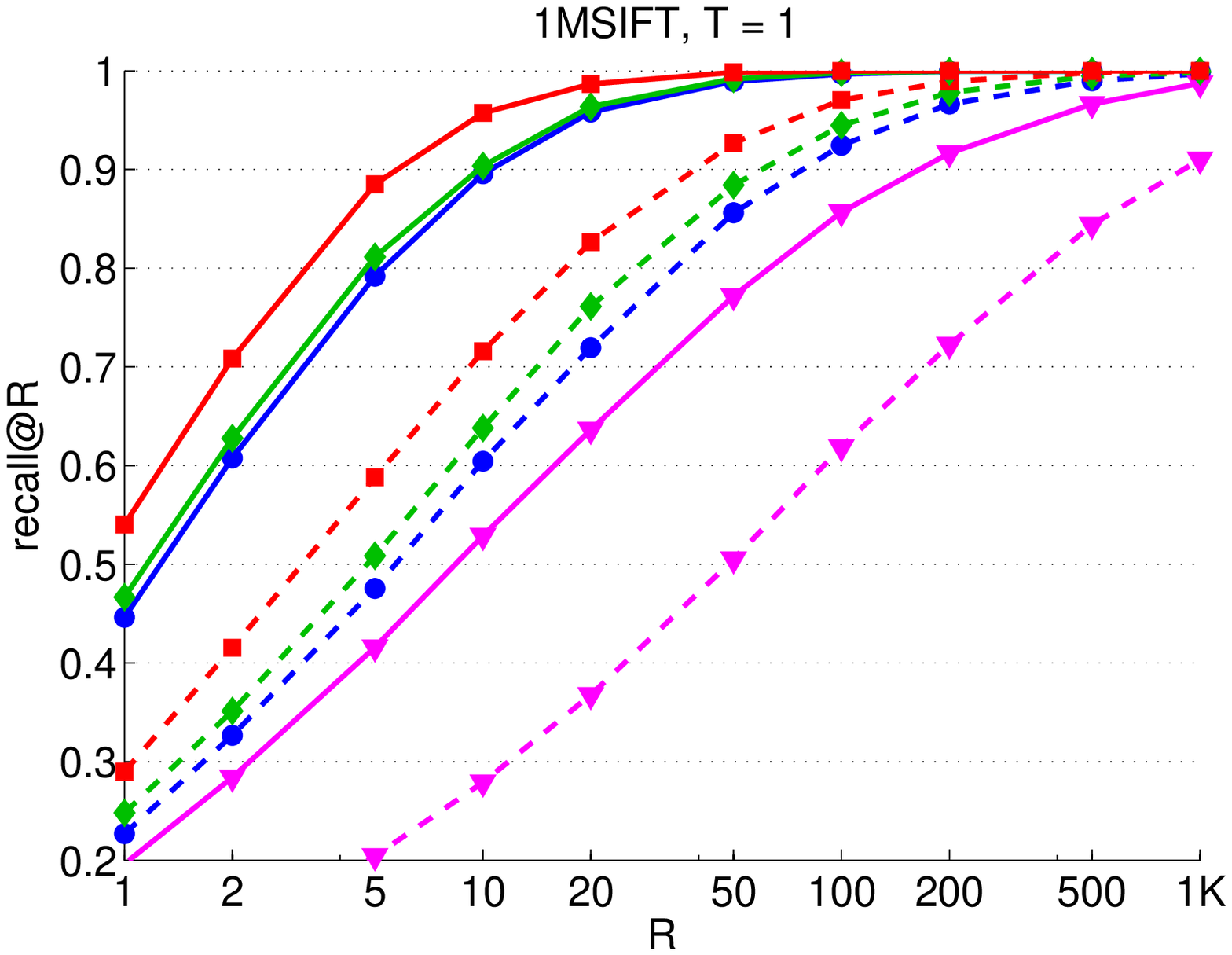}~~~
    \includegraphics[width=.44\linewidth, clip]{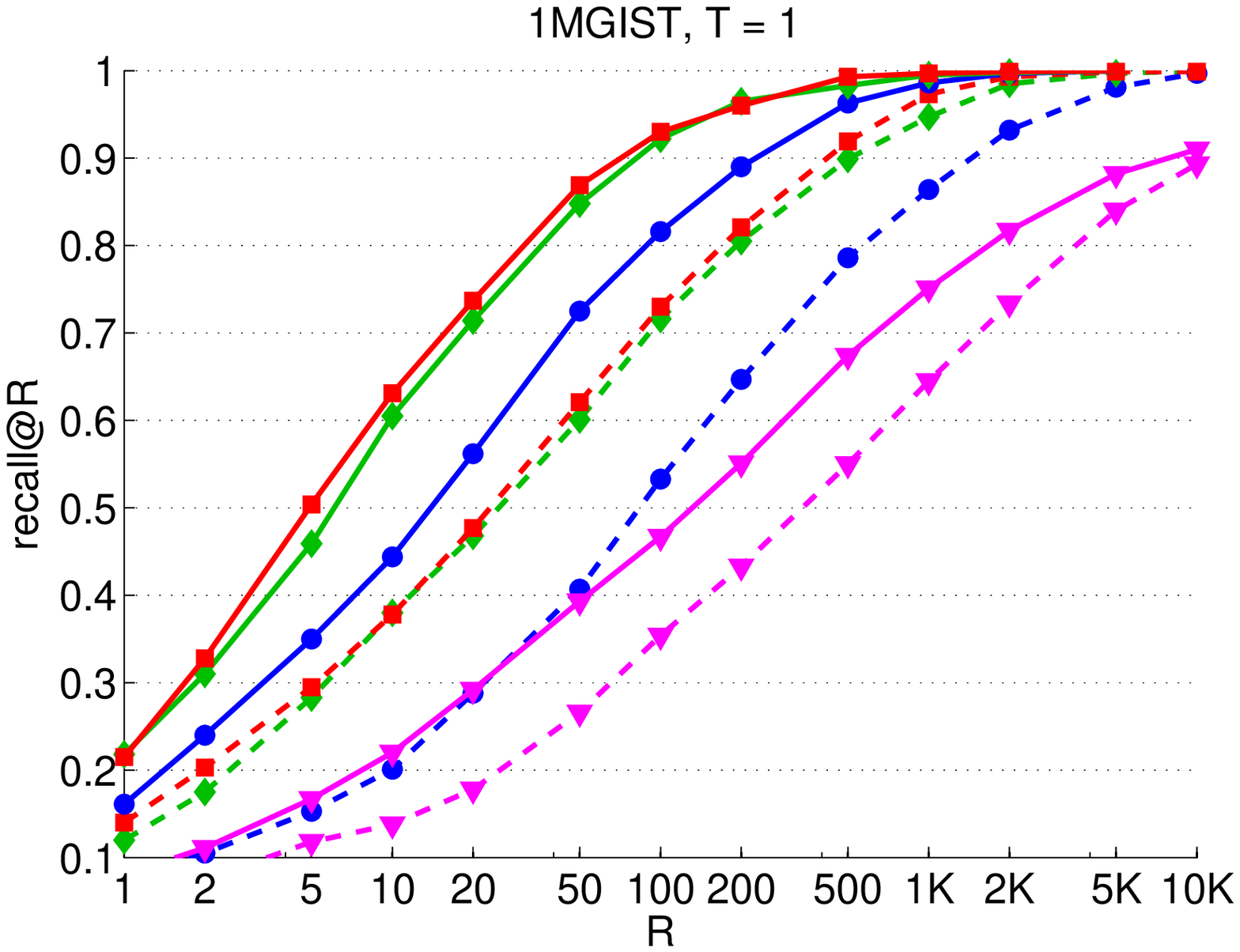}\\
	\includegraphics[width=.44\linewidth, clip]{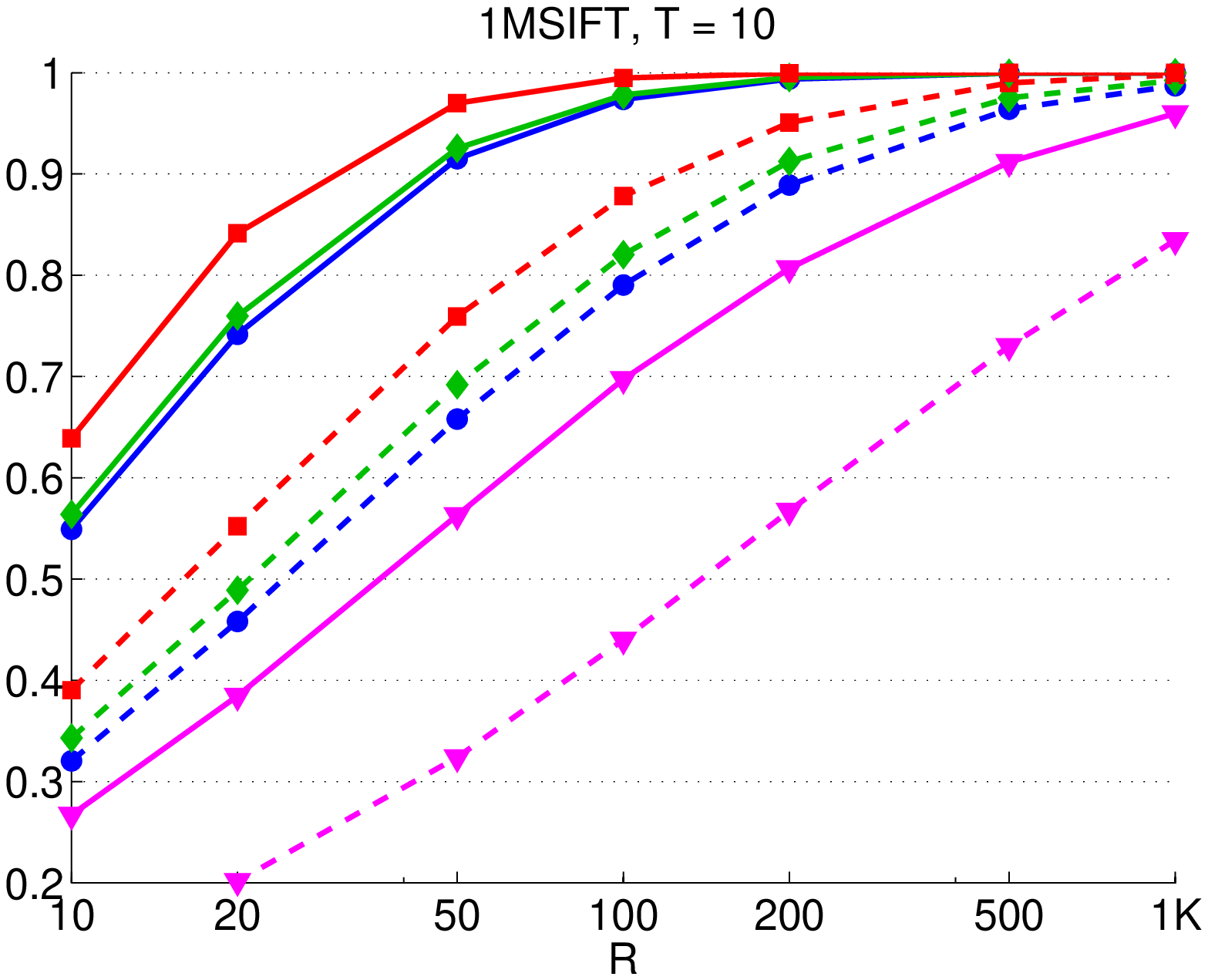}~~~
    \includegraphics[width=.44\linewidth, clip]{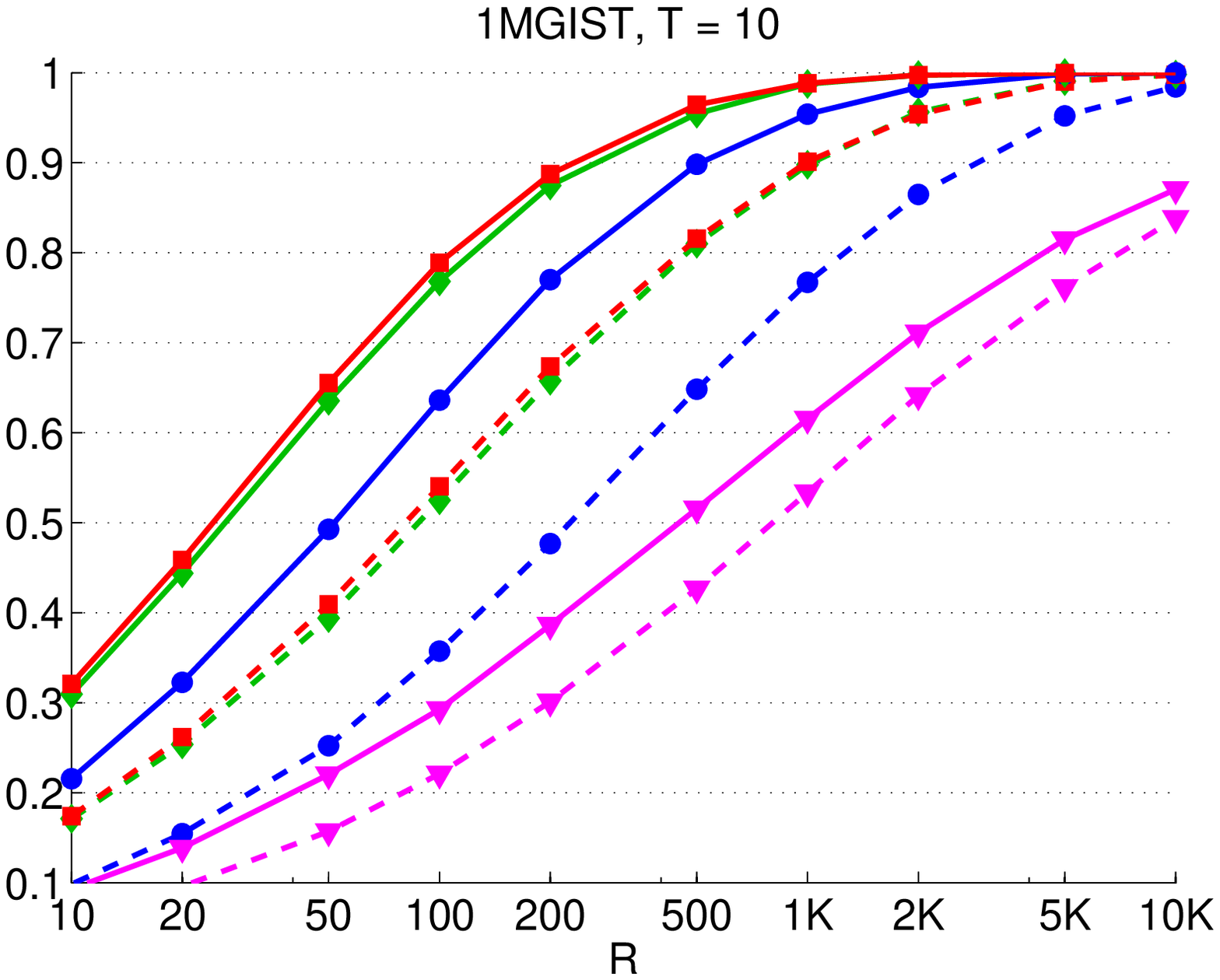}\\
	\subfigure[]{\includegraphics[width=.44\linewidth, clip]{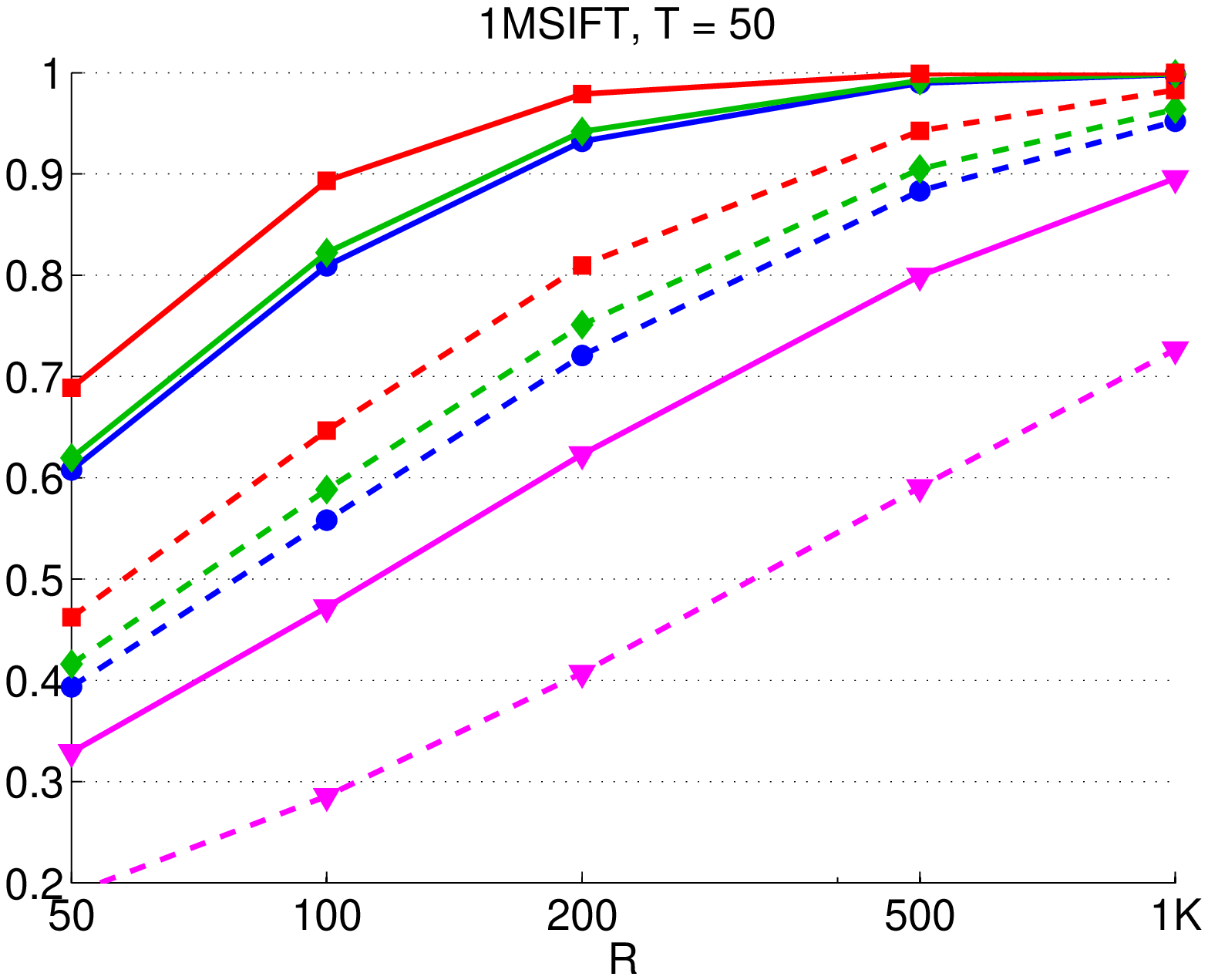}}~~~
	\subfigure[]{\includegraphics[width=.44\linewidth, clip]{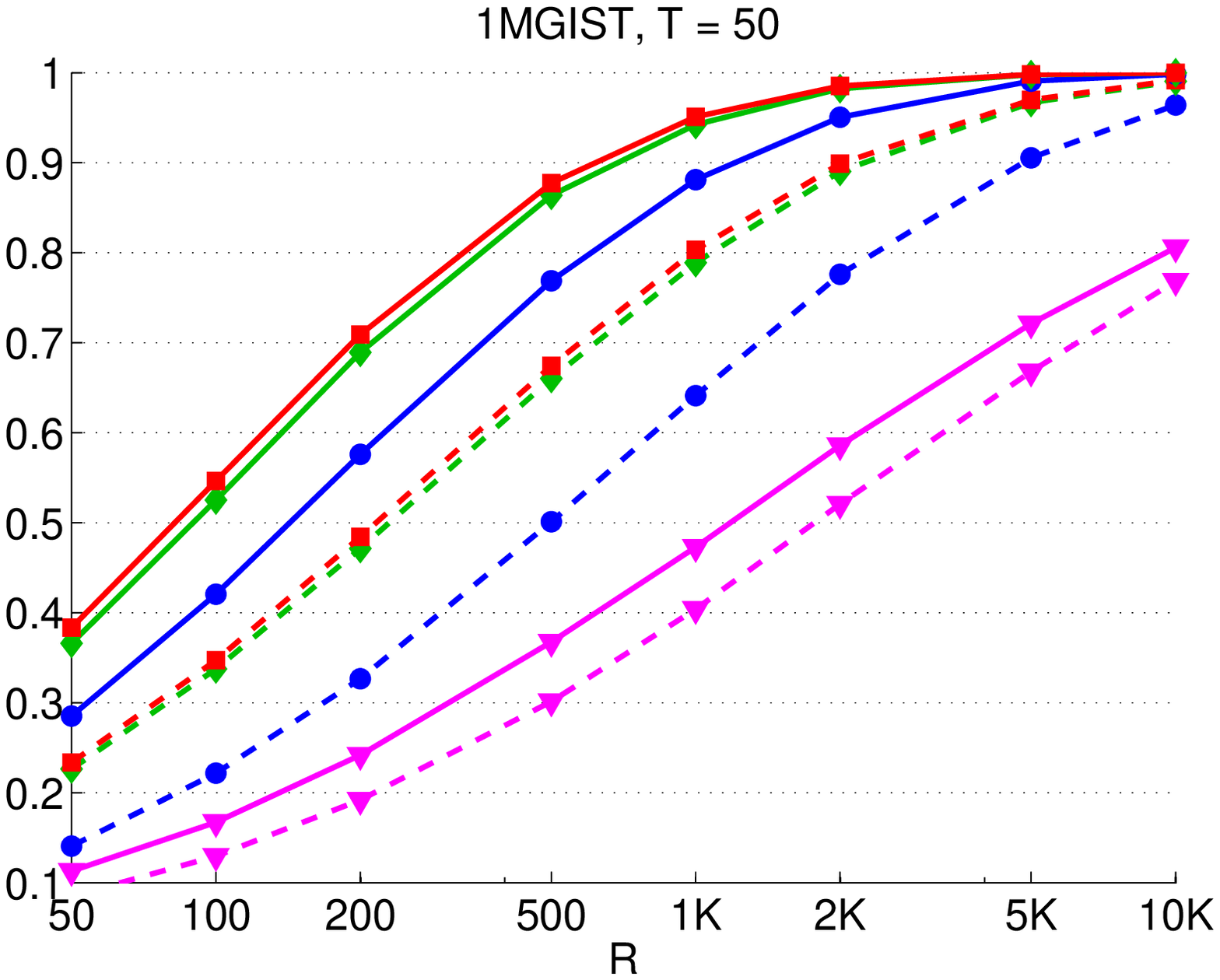}}
	\vspace{-.3cm}
	\caption{The performance for different algorithms
		on (a) $1M$SIFT and (b) $1M$GIST
		for searching various numbers of ground truth nearest neighbors
		($T=1, 10, 50$).}
	\label{fig:resultsof1MSIFT1MGIST}
\end{figure}

\begin{figure}[htb!]
	\centering
\footnotesize
	 \includegraphics[width = .9\linewidth, clip]{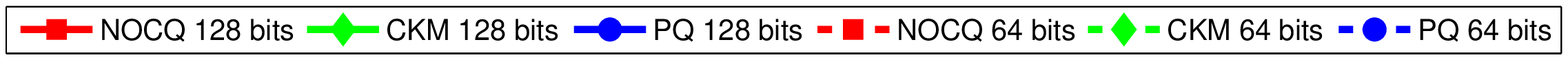}\\
	\includegraphics[width=.33\linewidth, clip]{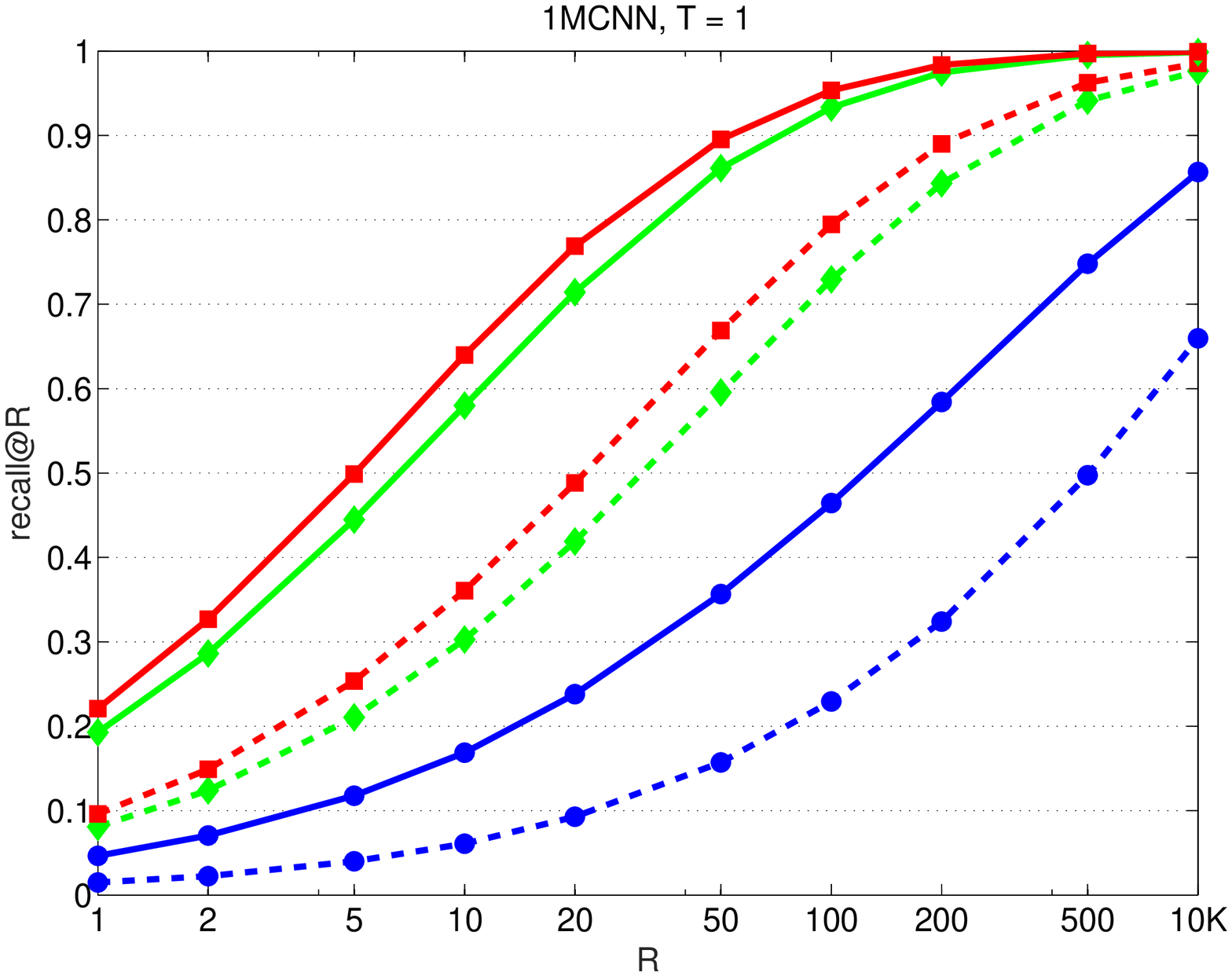}~
	\includegraphics[width=.33\linewidth, clip]{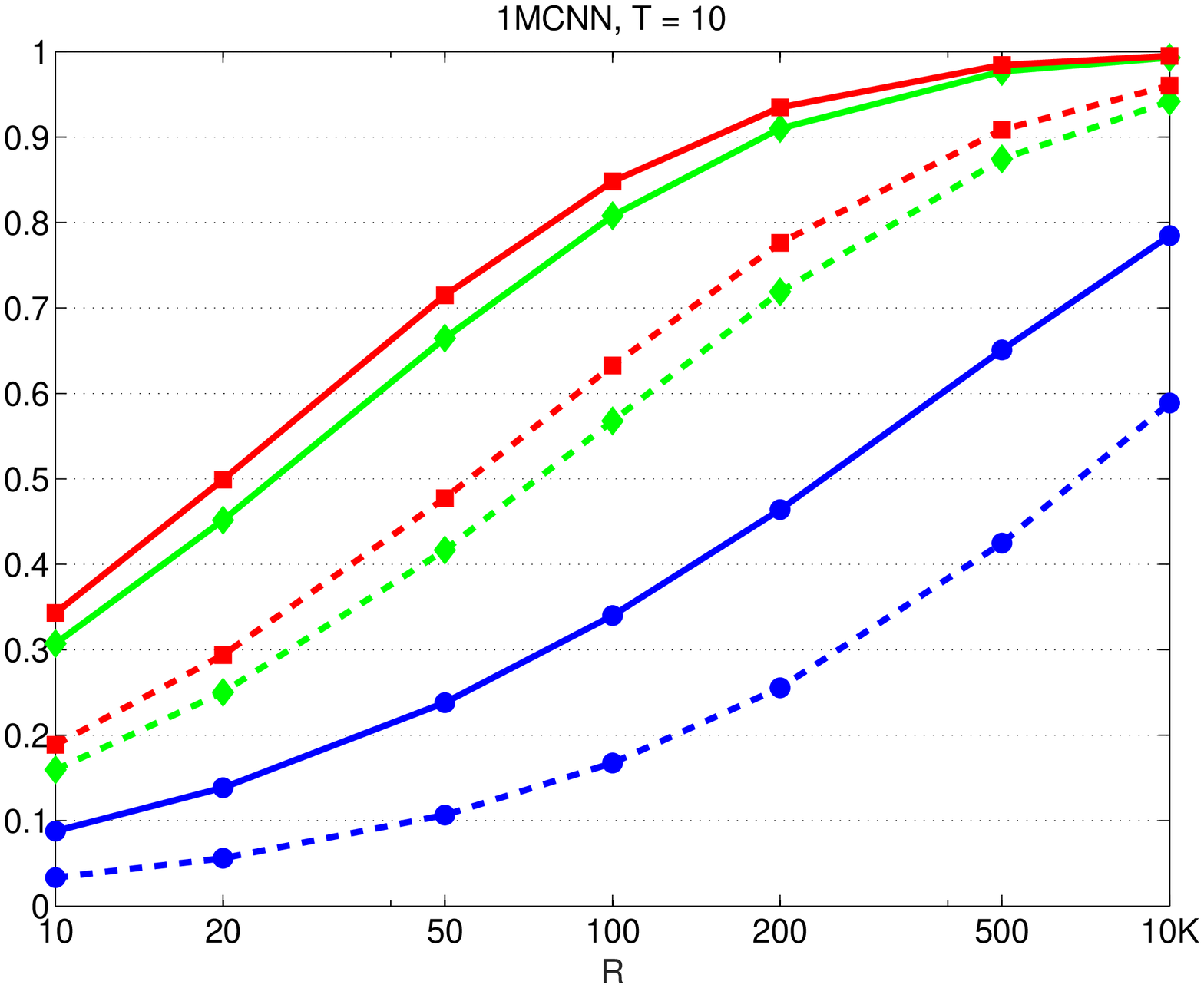}~
	\includegraphics[width=.33\linewidth, clip]{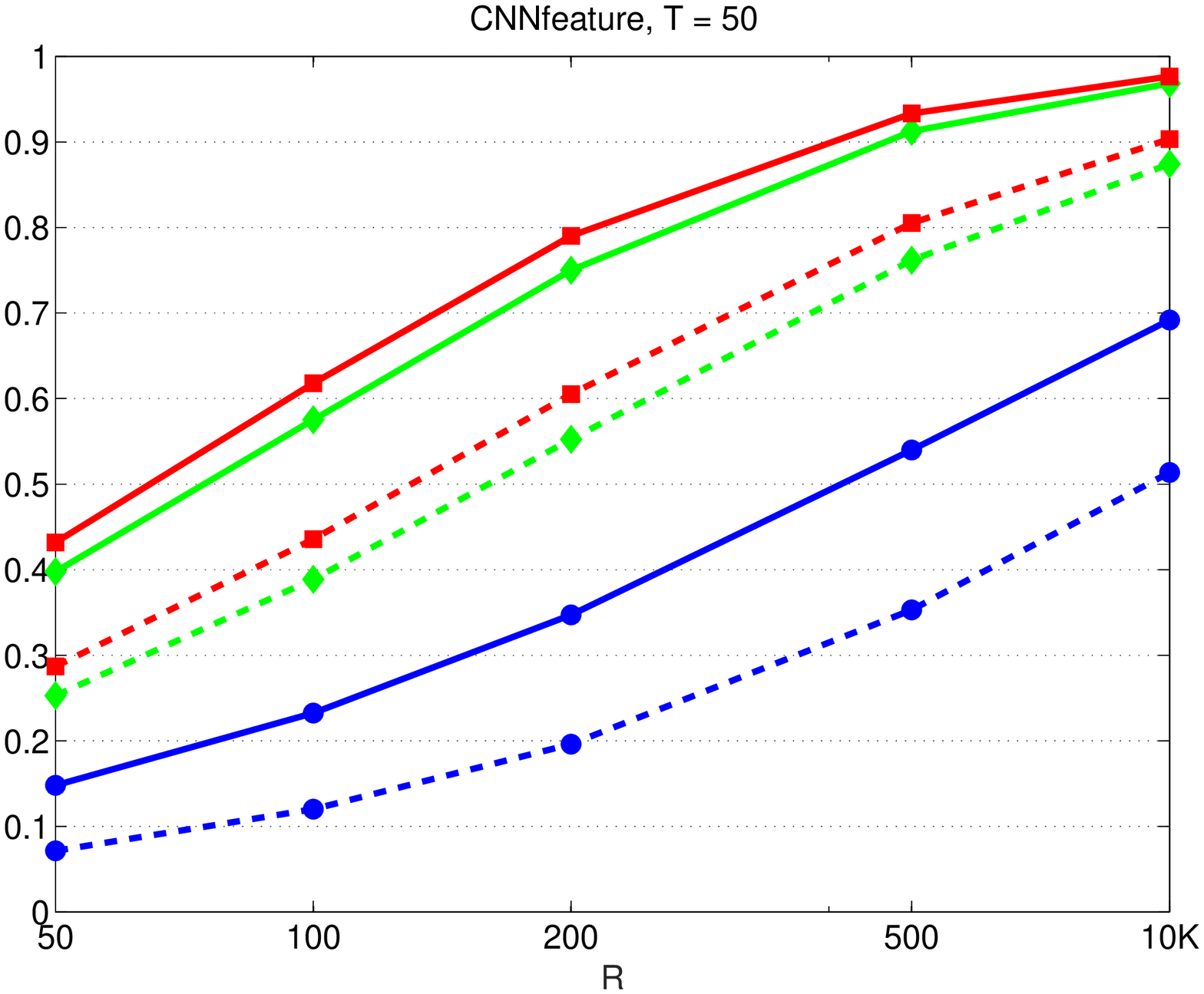}
	\vspace{-.3cm}
	\caption{The performance for different algorithms
		on CNN features
		with $64$ bits and $128$ bits
		for searching various numbers of ground truth nearest neighbors
		($T=1, 10, 50$).}
	\label{fig:resultsofCNN}
\end{figure}

\vspace{.1cm}
\noindent\textbf{The effect of a global translation.}
One potential extension of our approach
is to introduce an offset,
denoted as $\mathbf{t}$,
to translate $\mathbf{x}$.
Introducing the offset does not increase
the storage cost as it is a global parameter.
The objective function with such an offset
is as follows:
$\min\nolimits_{\{\mathbf{C},\mathbf{t},
	\mathbf{y}_1,\cdots,\mathbf{y}_N\}}\sum_{n=1}^N\|\mathbf{x}_n -  \mathbf{t} - \mathbf{C}\mathbf{y}_{n}\|_2^2$.
Our experiments
indicate that this introduction
does not influence the performance too much.
An example result on $1M$SIFT with $64$ bits is shown in Figure~\ref{fig:translation}.
The reason might be
that the contribution of the offset
to the quantization distortion reduction
is relatively small compared with
that from the composition of selected dictionary elements.

\begin{figure}[htb!]
	\centering
	\includegraphics[width = .9\linewidth, clip]{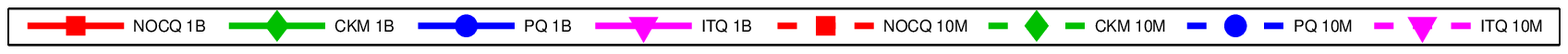}\\
	\includegraphics[width=.44\linewidth, clip]{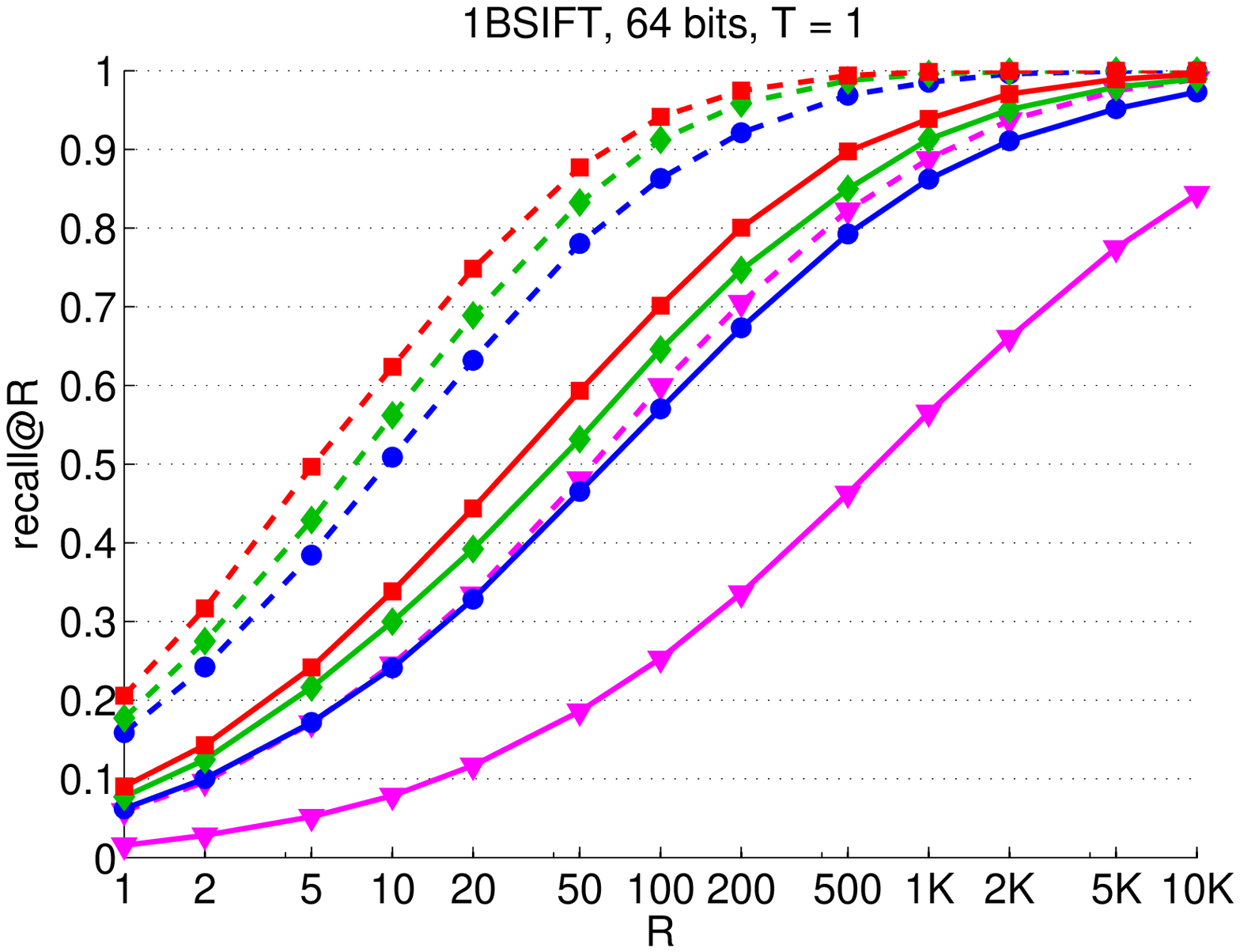}~~~
\includegraphics[width=.44\linewidth, clip]{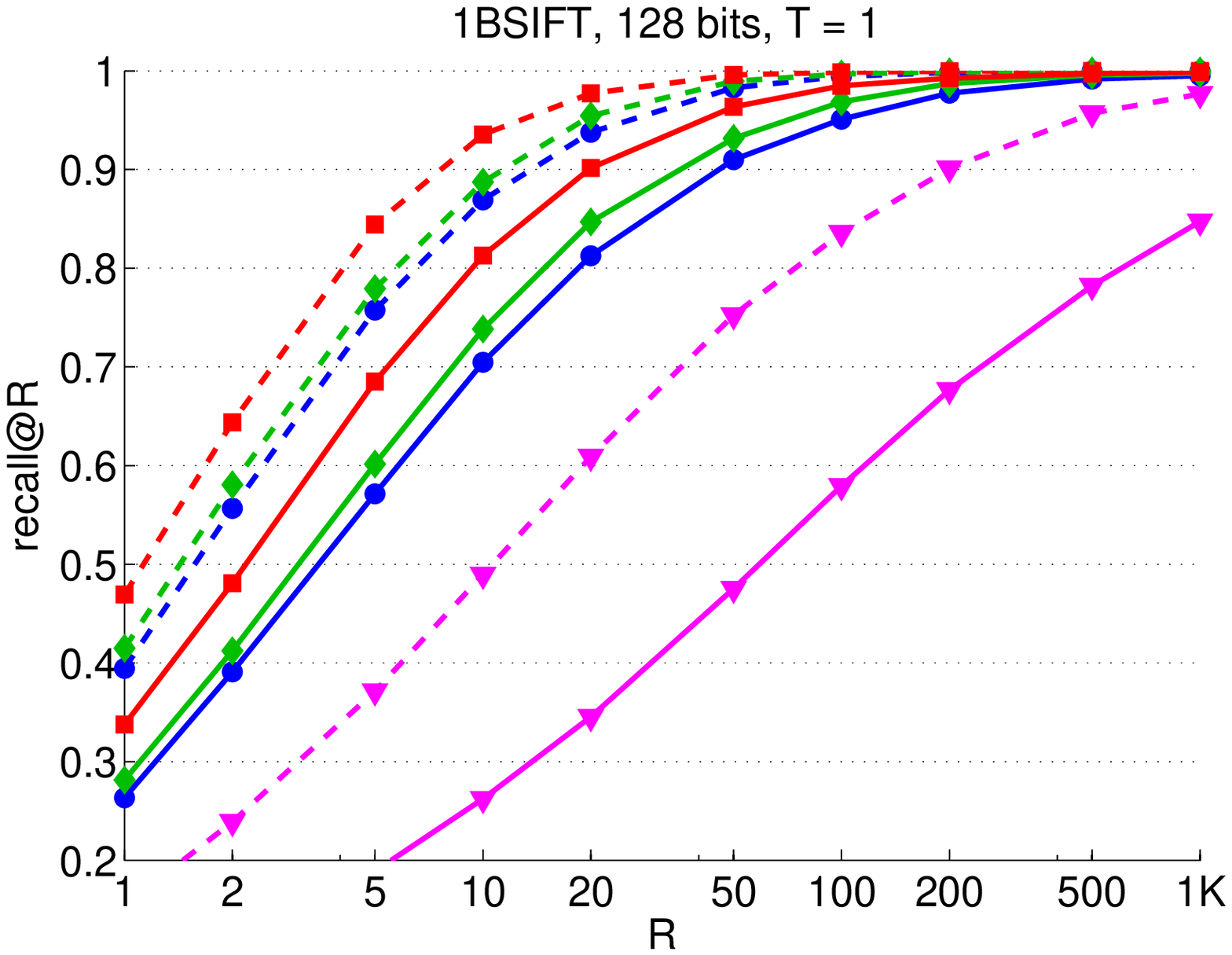} \\
	\includegraphics[width=.44\linewidth, clip]{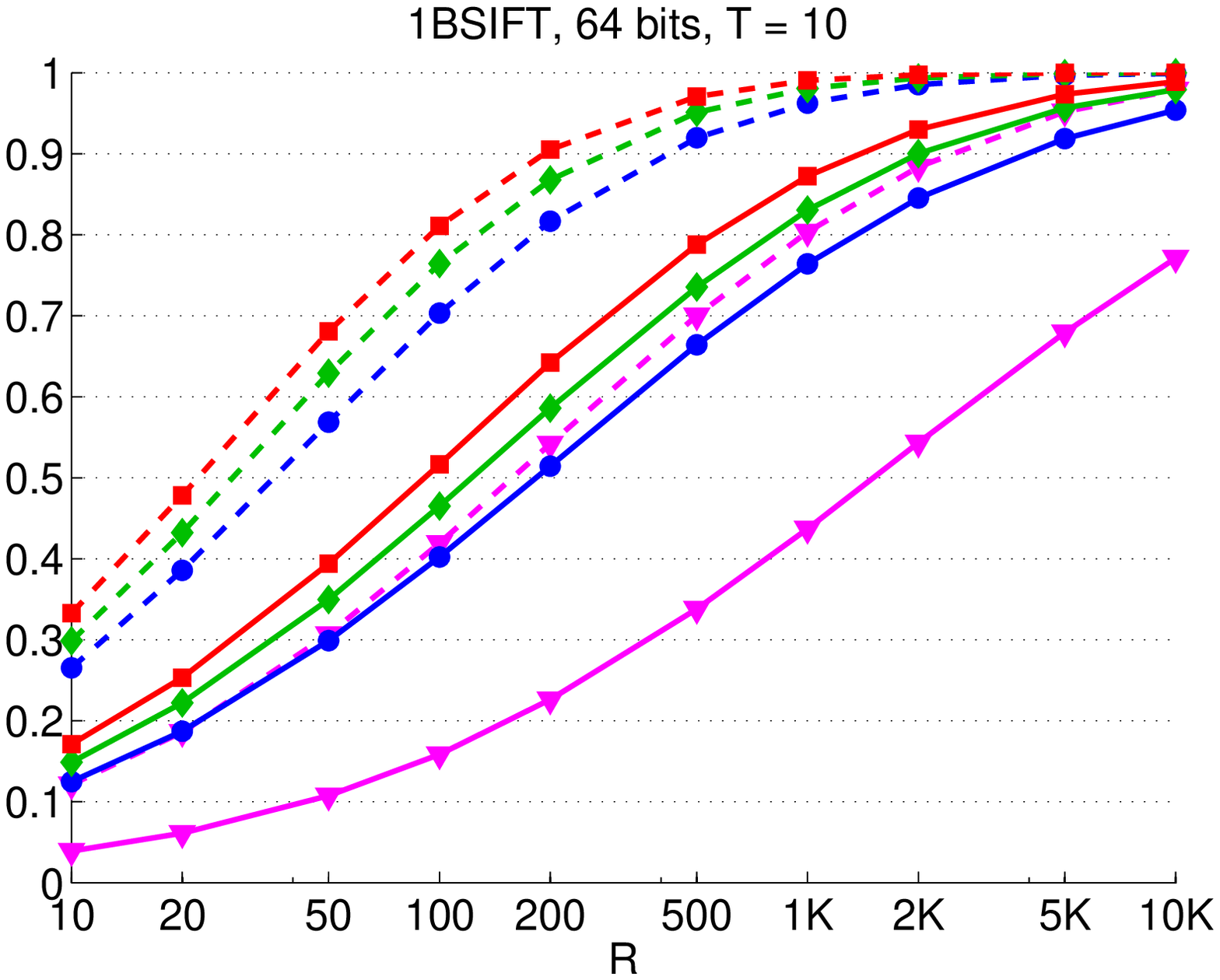}~~~
\includegraphics[width=.44\linewidth, clip]{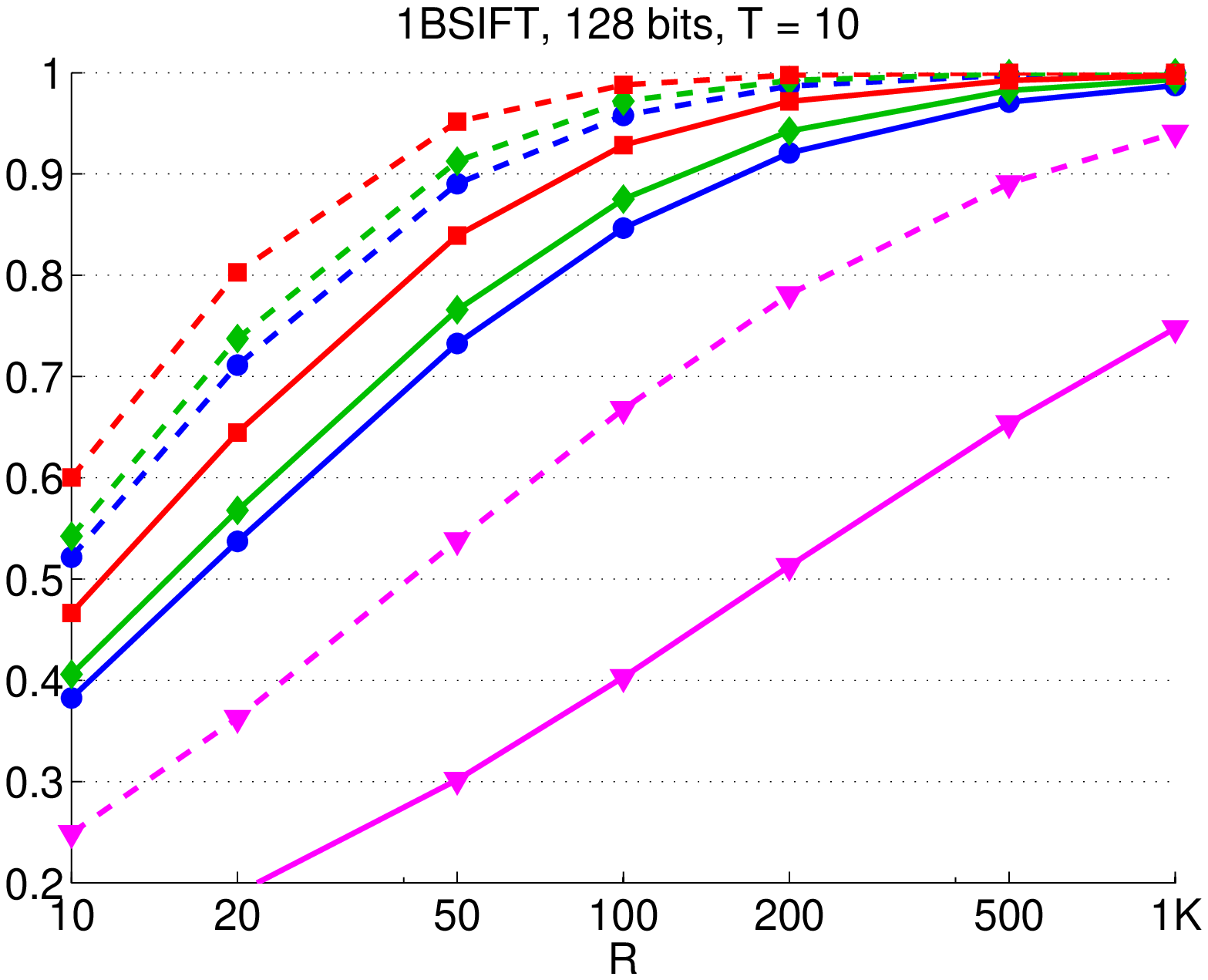} \\
	\subfigure[]{\includegraphics[width=.44\linewidth, clip]{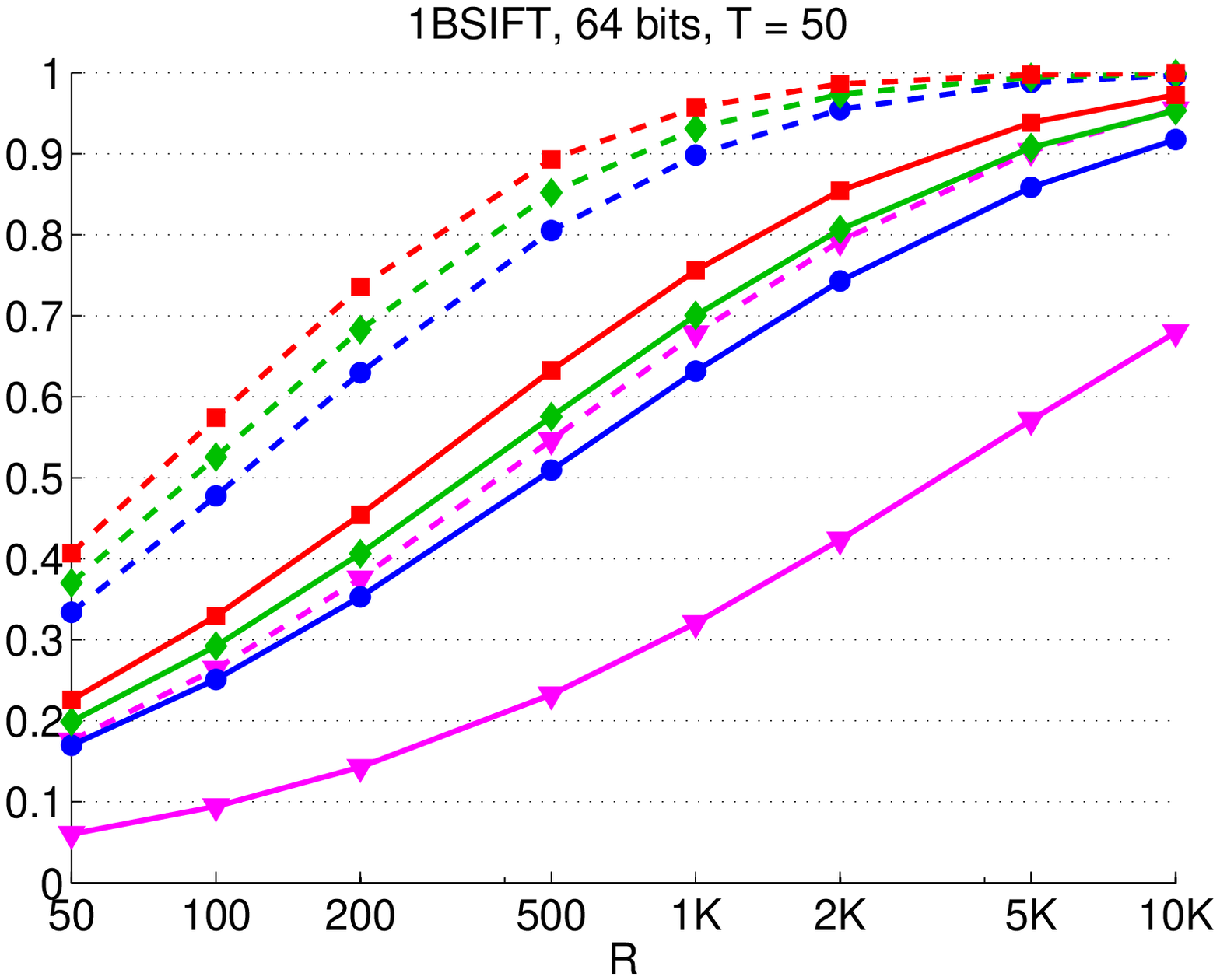}}~~~
	\subfigure[]{\includegraphics[width=.44\linewidth, clip]{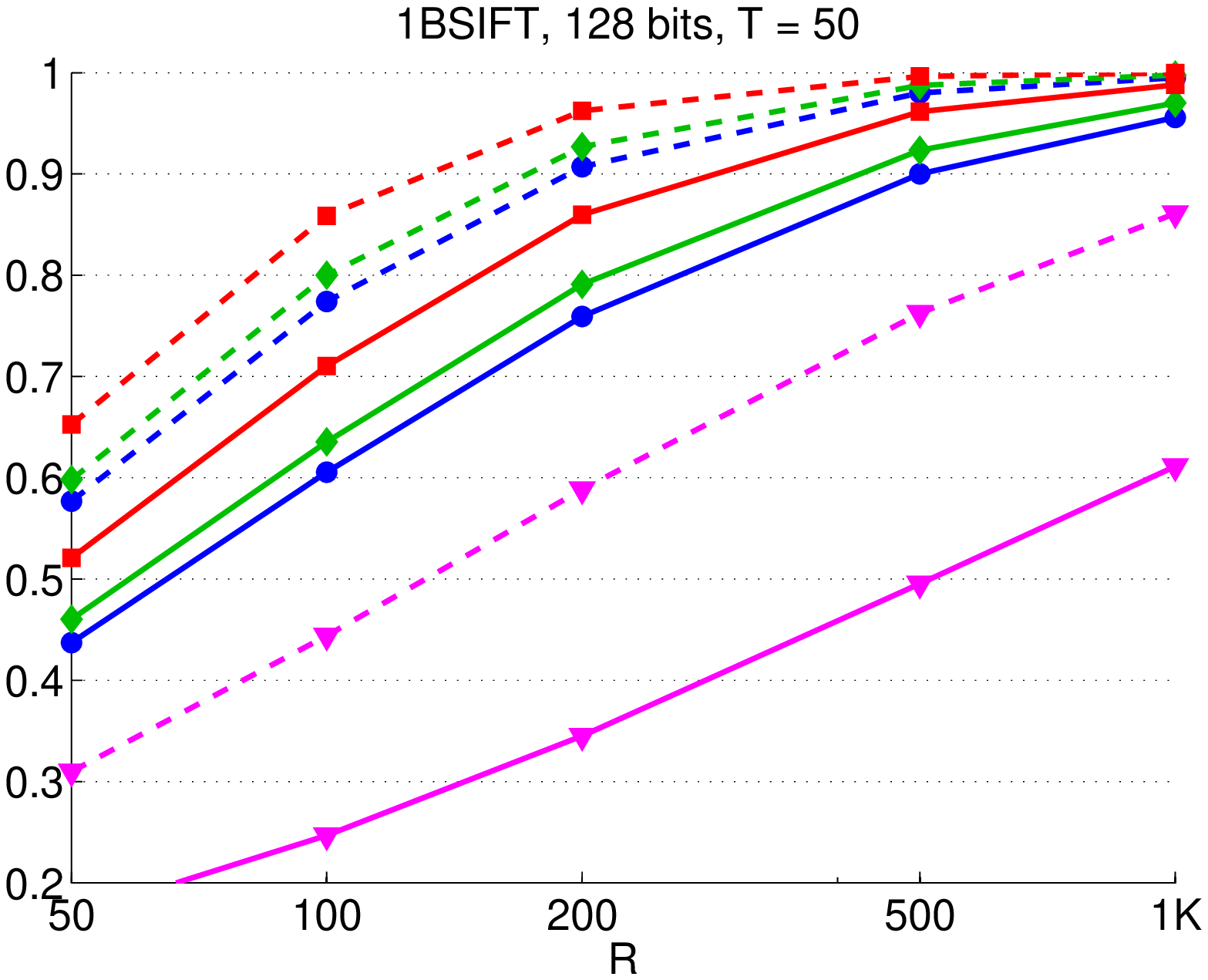}}
	\vspace{-.3cm}
	\caption{The performance for different algorithms
		on $1B$ SIFT
		with (a) $64$ bits and (b) $128$ bits
		for searching various numbers of ground truth nearest neighbors
		($T=1, 10, 50$).}
	\label{fig:resultsof1BSIFT}
\end{figure}

\subsection{Comparison}

\noindent\textbf{Recall@R.}
Figure~\ref{fig:resultsofothers} shows the comparison
on MNIST
and LabelMe$22K$.
One can see that
the vector approximation algorithms,
our approach (NOCQ), CKM,
and PQ,
outperform ITQ.
It is as expected
because the information loss in Hamming embedding used in ITQ
is much larger.
PQ also performs not so good
because it does not well exploit the data information
for subspace partitioning.
Our approach (NOCQ) is superior over CKM,
and performs the best.
The improvement seems a little small,
but it is actually significant
as the datasets are relatively small
and the search is relatively easy.


Figure~\ref{fig:resultsof1MSIFT1MGIST} shows the results of
large scale datasets: $1M$SIFT and $1M$GIST,
using codes of $64$ bits and $128$ bits
for searching $1$, $10$, and $50$ nearest neighbors.
It can be seen that
the gain obtained by our approach is significant for $1M$SIFT.
For example,
the recall$@10$ with $T = 1$ and $64$ bits
for our approach
is $71.59\%$,
about $11\%$ better than the recall $60.45\%$ of PQ,
and $7\%$ larger than the recall
$63.83\%$ of CKM.
The reason of the relatively small improvement on $1M$GIST
for NOCQ over CKM
might be
that CKM already achieves very large improvement over product quantization
and the improvement space is relatively small.

Figure~\ref{fig:resultsofCNN} shows the performance
on another large dataset, $1M$CNN,
of a higher dimension. We use $10,000$ dataset vectors randomly sampled from the base set for efficient training. We do not show the ITQ results because ITQ performs much lower than PQ. It can be seen that our approach (NOCQ) outperforms
PQ and CKM with both $64$ bits and $128$ bits encoding.
For example, the recall$@10$ for our approach with $T=1$
and $64$ bits is $36.07\%$, about $6\%$ better than the recall of CKM, which is $30.28\%$, and about $30\%$ better than the recall of PQ, which is $6.08\%$.
With $128$ bits encoding and $T=1$, the recall of our approach is $63.98\%$, about $6\%$ better than the recall $58.00\%$ of CKM and $47\%$ better than the recall $16.85\%$ of PQ. Note that CKM outperforms PQ significantly,
which indicates that space partition is very important for $1M$CNN dataset. Our approach, on the other hand, still gets large improvement over CKM due to the more accurate data approximation.


Figure~\ref{fig:resultsof1BSIFT} shows the performance
for a very large dataset,
$1B$SIFT.
Similar to~\cite{NorouziF13},
we use the first $1M$ learning vectors for efficient training.
It can be seen that our approach, NOCQ, gets the best performance
and the improvement is consistently significant.
For example,
the recall$@100$ from our approach on the $1B$ base set with $64$ bits
for $T = 1$ is $70.12\%$
while from CKM it is $64.57\%$.
Besides the performance over
all the $1B$ database vectors,
we also show the performance
on a subset of $1B$ base vectors, the first $10M$ database vectors.
As we can see,
the performance on $1B$ vectors
is worse than that on $10M$ vectors,
which is reasonable
as searching over a larger dataset
is more difficult.
The notable observation
is that
the improvement of our approach over Cartesian $k$-means
on the larger dataset, $1B$ database vectors,
is much more significant
than
that on $10M$ database vectors.

\begin{figure*}[t]
	\centering
\footnotesize
	(a){\includegraphics[width=.145\linewidth, clip]{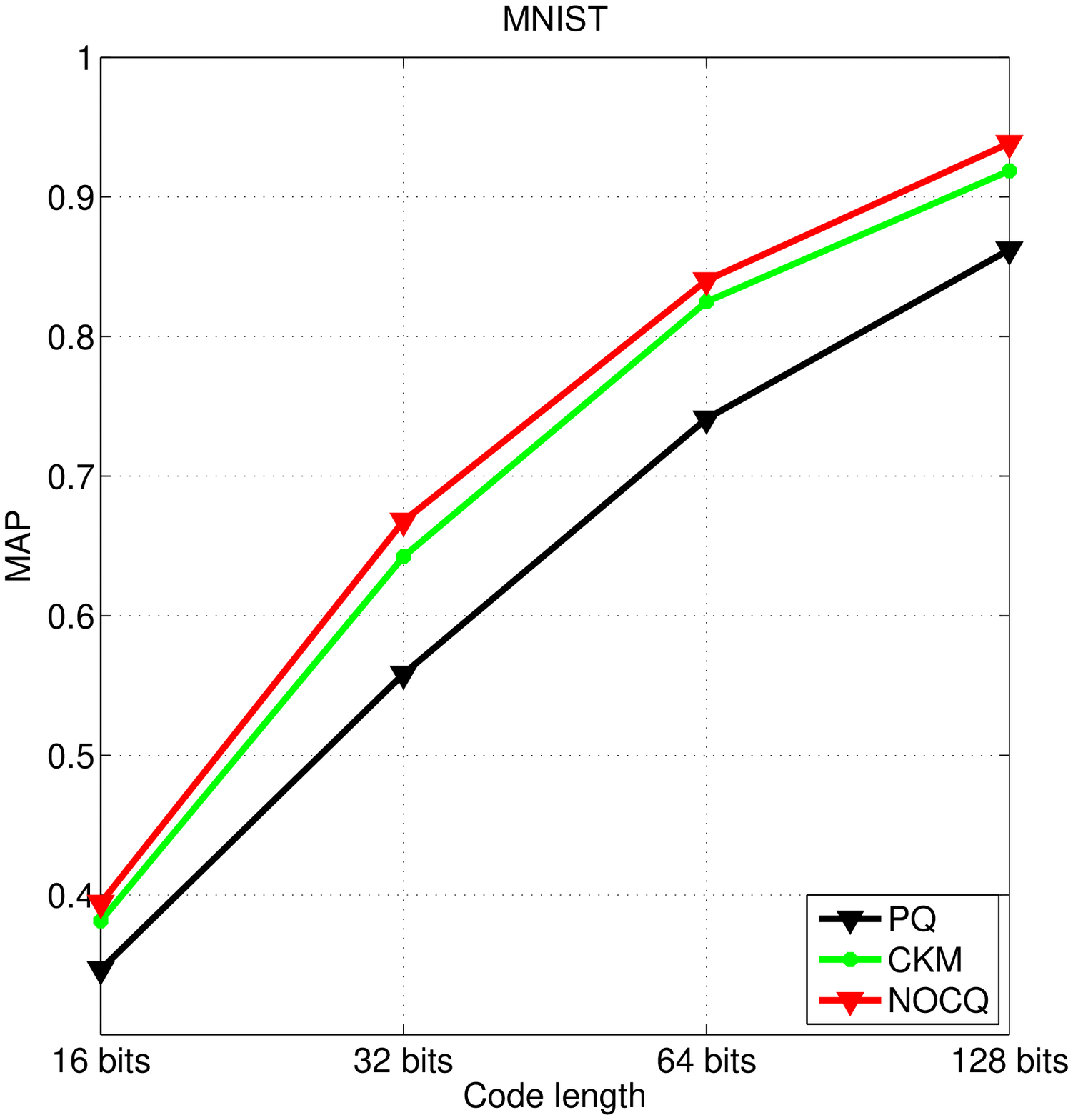}}
	(b){\includegraphics[width=.145\linewidth, clip]{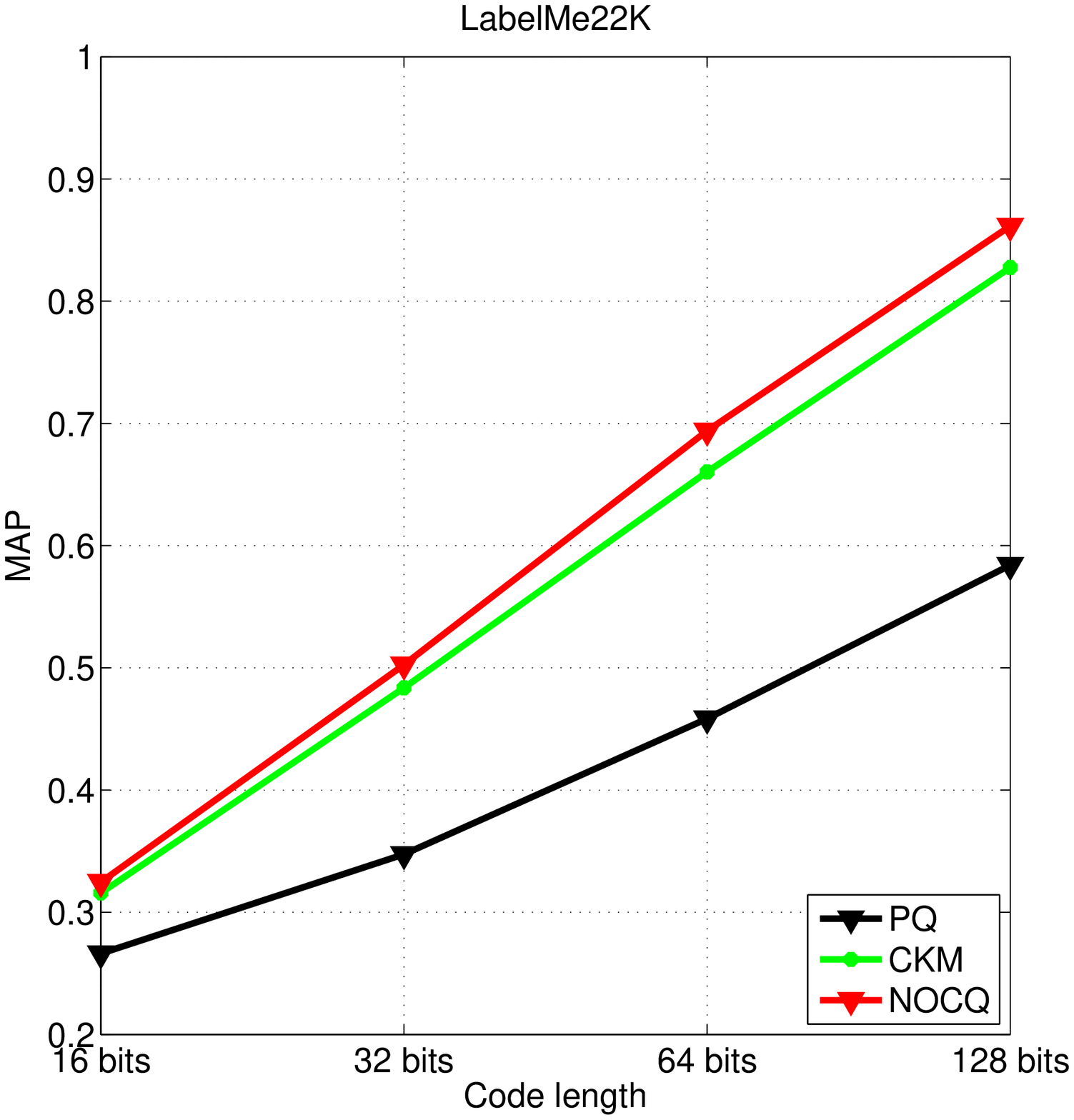}}
	(c){\includegraphics[width=.145\linewidth, clip]{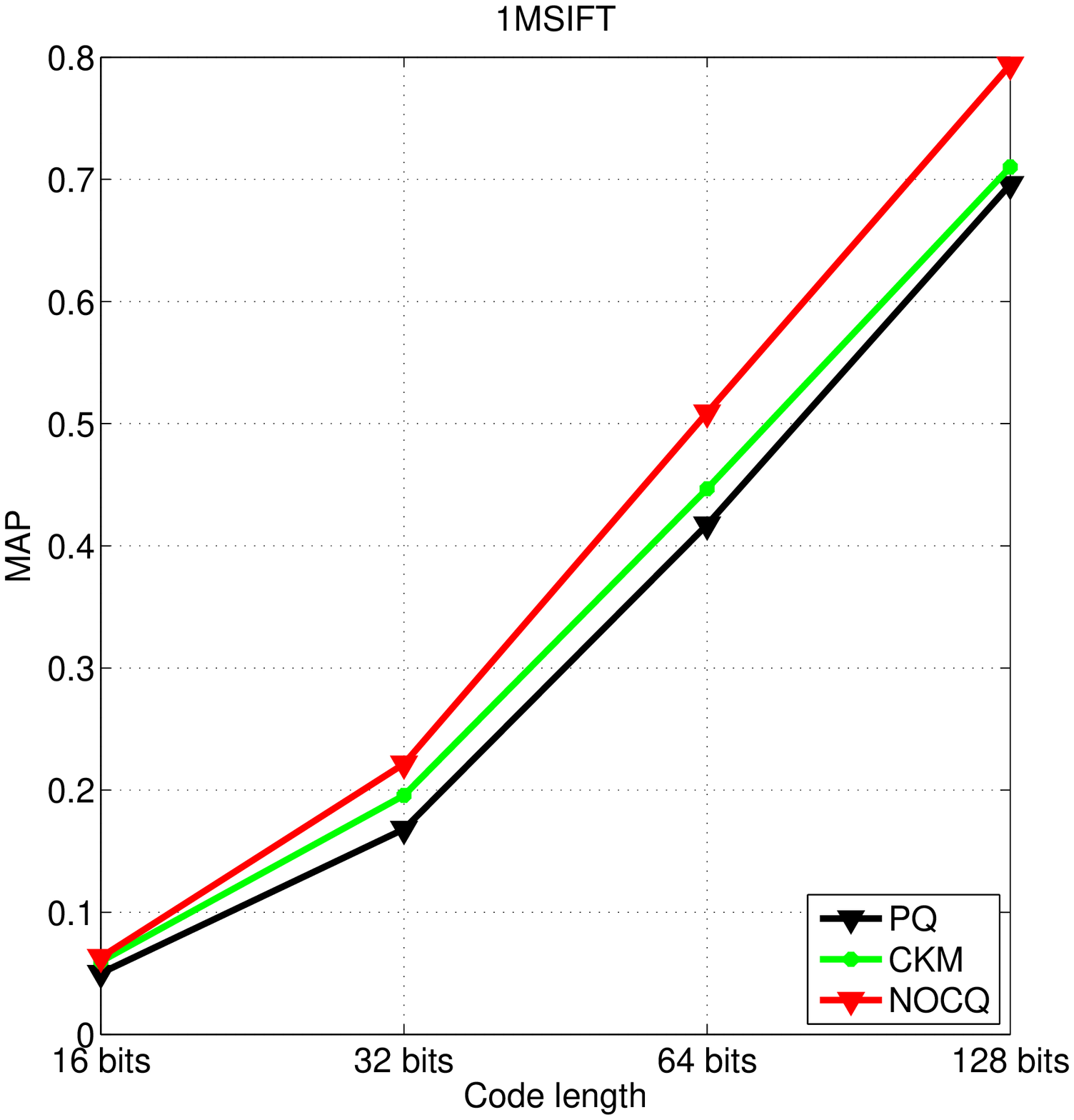}}
	(d){\includegraphics[width=.145\linewidth, clip]{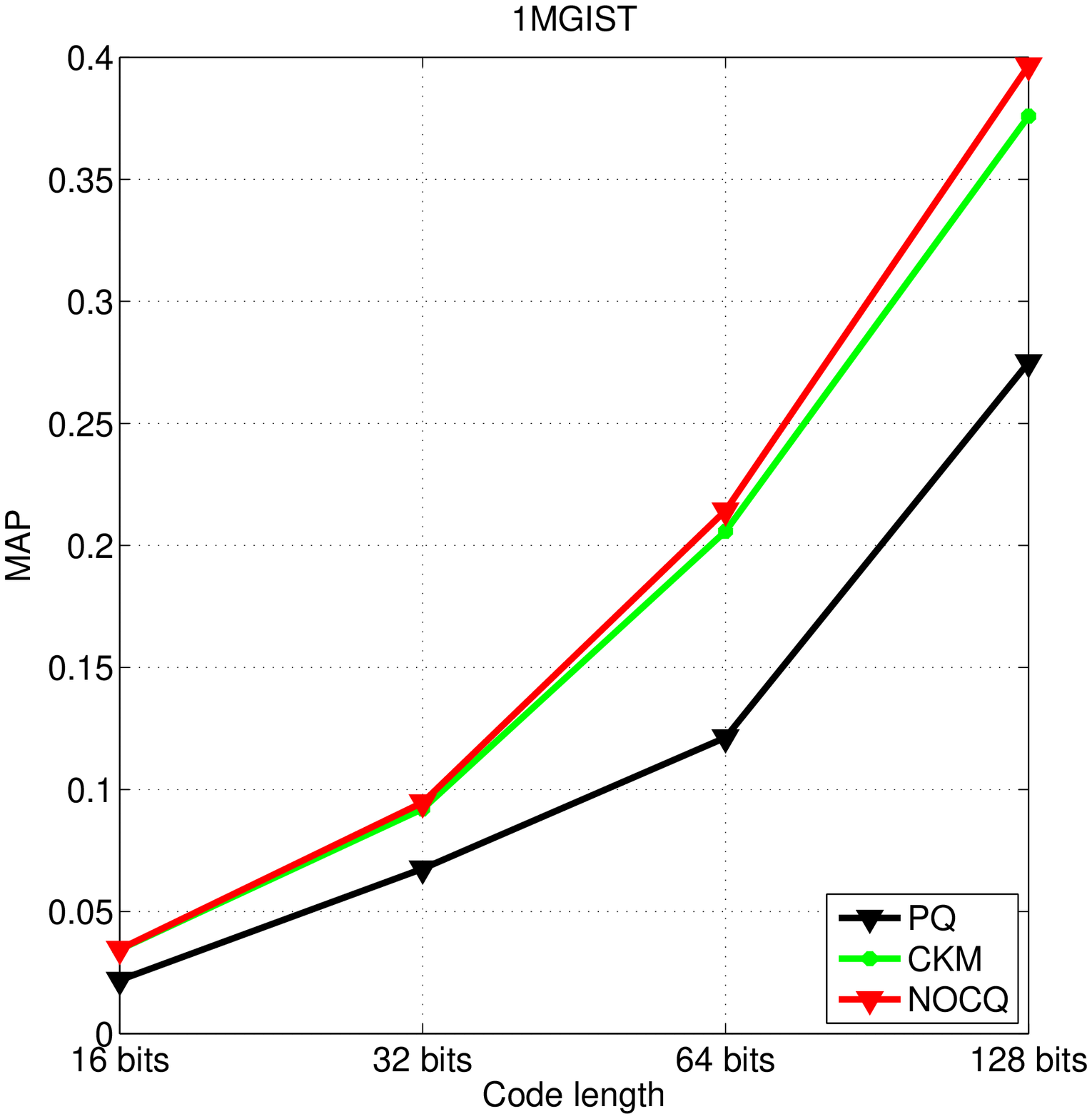}}
	(e){\includegraphics[width=.145\linewidth, clip]{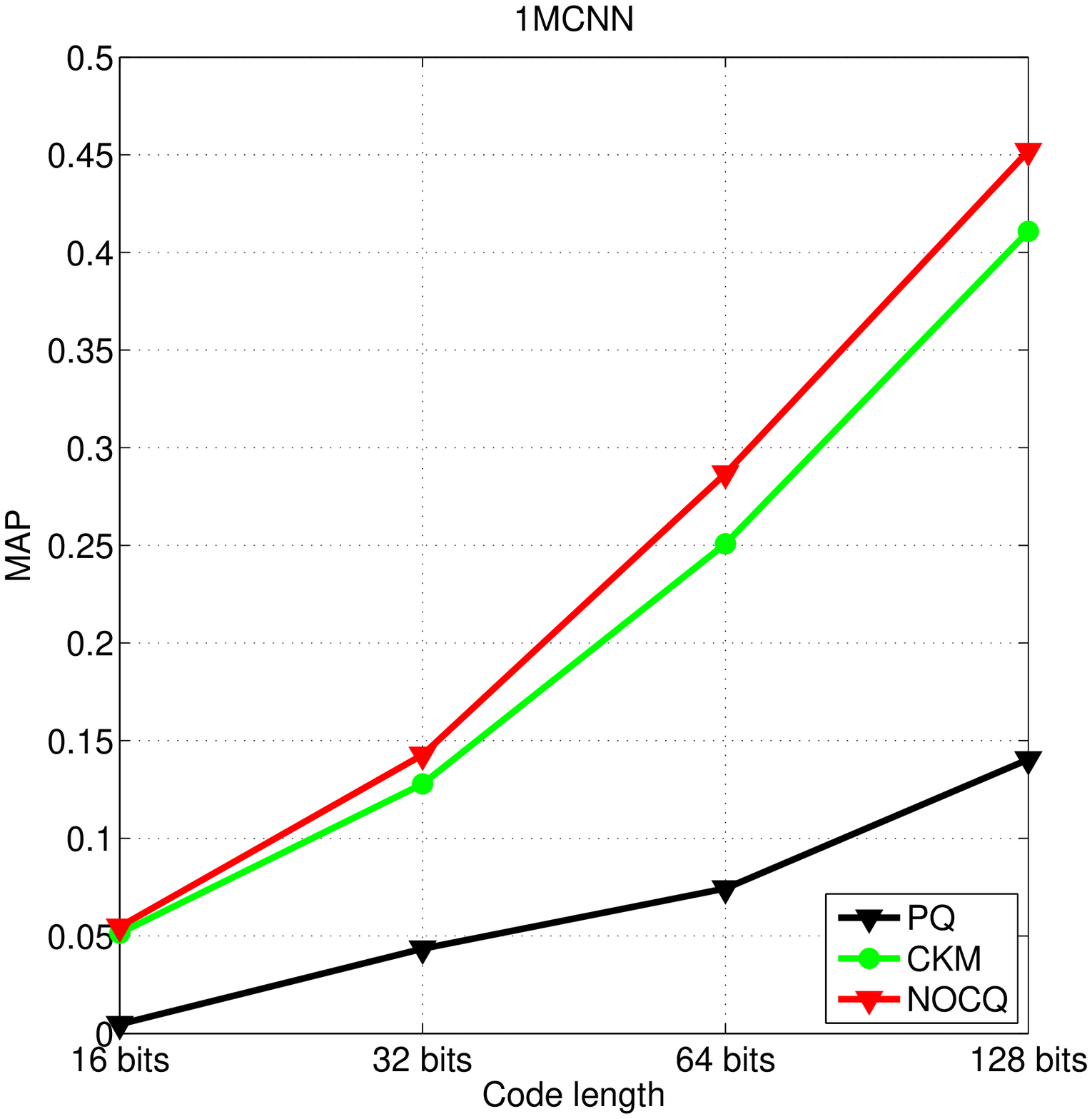}}
	(f){\includegraphics[width=.145\linewidth, clip]{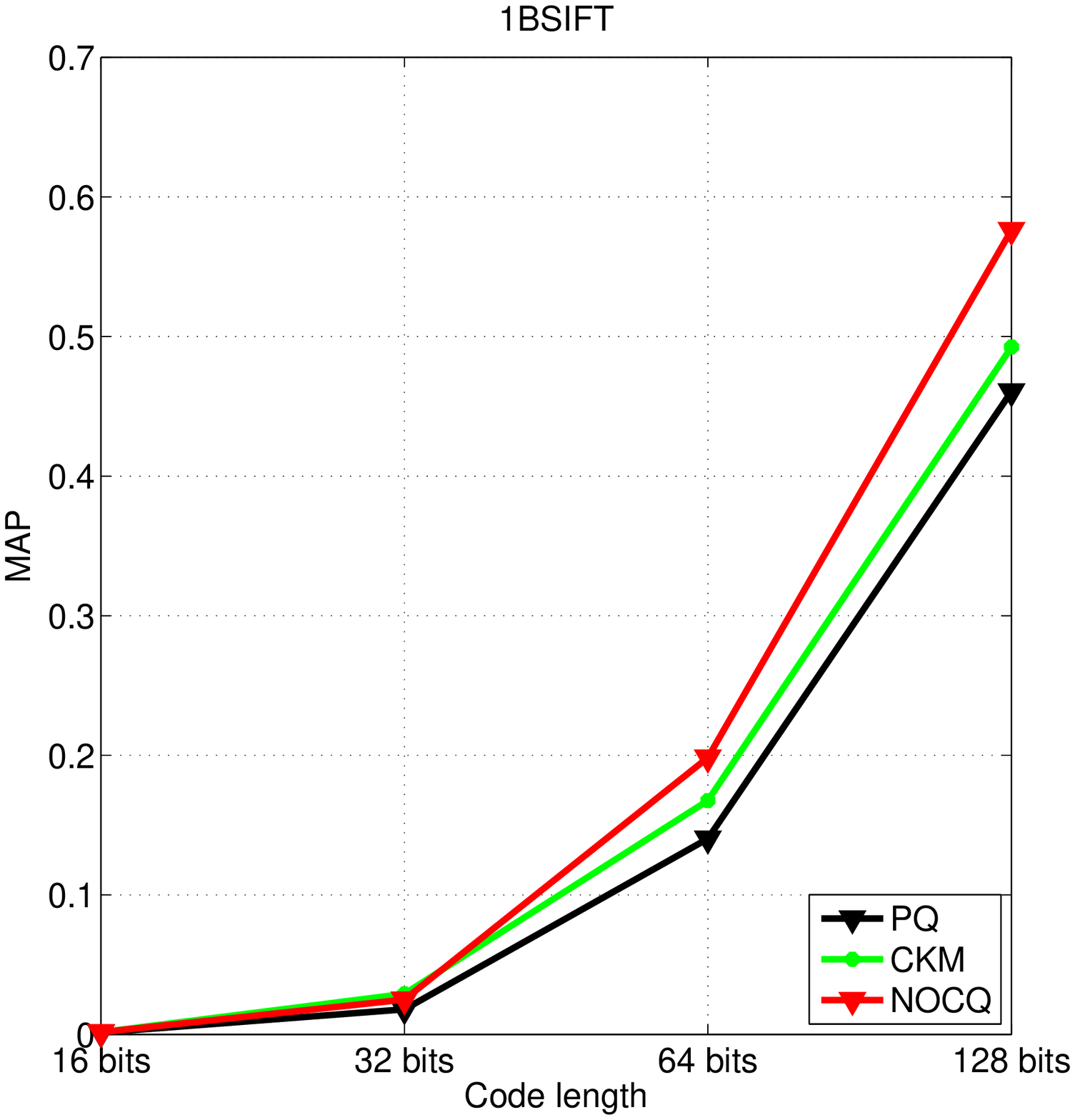}}
\vspace{-.4cm}
	\caption{The performance
		in terms of MAP vs. code length
		for different algorithms
		on (a) MNIST, (b) LabelMe$22K$.
		(c) $1M$SIFT, (d) $1M$GIST,
		(e) $1M$CNN and (f) $1B$SIFT.}
	\label{fig:resultsofmap}\vspace{-.3cm}
\end{figure*}

\vspace{.1cm}
\noindent\textbf{MAP vs. \#bits.}
Figure~\ref{fig:resultsofmap}
shows the MAP performance on (a) MNIST, (b) LabelMe$22K$.
(c) $1M$SIFT, (d) $1M$GIST,
(e) $1M$CNN and (f) $1B$SIFT
with various code lengths.
We can see that our approach (NOCQ) performs the best
on all the datasets.
It is worth noting that the improvement obtained on $1B$SIFT is significant
since searching in 1 billion vectors is not easy.
For instance,
the MAP scores of $1M$SIFT on 64 bits for NOCQ
and PQ are $0.51$ and $0.417$,
and the relative improvement is about $22.3\%$.
The MAP scores for NOCQ and CKM are $0.51$ and
$0.447$, and the improvement reaches $14.1\%$.

\section{Applications}
\subsection{Inverted Multi-Index}
Inverted multi-index~\cite{BabenkoL12} performs product quantization on the database vectors to
generate the cell centroids which store
the list of vectors that lie in a cluster.
The multi-sequence algorithm~\cite{BabenkoL12} is introduced
to efficiently produce a sequence of multi-index cells
ordered by the increasing distances between the query
and the cell centroids,
with the aim of retrieving the NN candidates.
After that, the retrieved candidates are often reranked based on their short codes,
e.g., through product quantization~\cite{JegouTDA11}.
We follow the Multi-D-ADC scheme~\cite{BabenkoL12},
which, in the reranking stage,
apply the compact coding algorithm
to the residual displacement between each vector
and its closest cell centroids obtained through the indexing stage.

All the compared methods can be applied to build inverted multi-index (coarse quantization used in the candidate retrieval stage)
and compact code representation (fine quantization used in the reranking stage).
On the candidate retrieval stage,
the distance table
between the query and the coarse codebooks is computed
before performing the multi-sequence algorithm for candidate retrieval.
On the reranking stage,
there are two ways for distance computation:
with reconstructing the database vector,
and without reconstructing the database vector
but through looking up distance tables like~\cite{BabenkoL14}.
Here we show how to accelerate the distance computation
without reconstructing the database vector
in our approach.

\noindent\textbf{Efficient distance computation.}
We adopt two dictionaries $\mathbf{C}_1$ and $\mathbf{C}_2$, suggested by~\cite{BabenkoL14}
for coarse quantization,
and represent the dictionaries for fine quantization
by $\{\mathbf{R}_1, \mathbf{R}_2, \cdots, \mathbf{R}_M\}$.
Let the approximation of a vector $\mathbf{x}$
be $\bar{\mathbf{x}} = \sum_{i=1}^2 \mathbf{c}_{ik_i}
+ \sum_{j=1}^M \mathbf{r}_{jk_j}$.
The acceleration idea is inspired by~\cite{BabenkoL14},
and illustrated below.
Expanding the approximated distance computation,
\begin{align}
&~ \|\mathbf{q} - (\sum\nolimits_{i=1}^2 \mathbf{c}_{ik_i} + \sum\nolimits_{j=1}^M \mathbf{r}_{jk_j})\|_2^2 \\
= &~\|\mathbf{q}\|_2^2 + \sum\nolimits_{i=1}^2 \|\mathbf{c}_{ik_i}\|_2^2 + \sum\nolimits_{j=1}^M \|\mathbf{r}_{jk_j}\|_2^2  - 2 \sum\nolimits_{i=1}^2 \mathbf{q}^{\top}\mathbf{c}_{ik_i} \nonumber \\
&
- 2 \sum\nolimits_{j=1}^M \mathbf{q}^{\top}\mathbf{r}_{jk_j}
+ 2 \sum\nolimits_{i=1}^2 \sum\nolimits_{j=1}^M \mathbf{c}_{ik_i}^{\top}\mathbf{r}_{jk_j} \nonumber \\
& + 2 \mathbf{c}_{1k_1}^{\top}\mathbf{c}_{2k_2} + 2 \sum\nolimits_{j=1}^M\sum\nolimits_{m=1, m\neq j}^M \mathbf{r}_{jk_j}^{\top}\mathbf{r}_{mk_m},
\end{align}
we can see that the right-hand-side contains $8$ terms.
The first term only depends on the query,
and is not necessary to compute.
The second and the third terms are the summation of the $L_2$ norms of the selected quantization centers,
where the norms are offline computed,
and hence the complexity is $O(M)$.
In the fourth term the inner products have been computed
in the candidate retrieval stage,
and this complexity is $O(1)$.
The fifth term takes $O(M)$ time
if the inner products between the query and the dictionary elements
for the fine quantizer
are precomputed.
The sixth term takes $O(M)$
when the inner products between the elements of the coarse and fine dictionaries are precomputed offline.
The last two terms are omitted because they are approximately equal to a constant
(the constant equals 0 for PQ and CKM).
In summary, the online time complexity is $O(M)$
with precomputing the inner product table storing the inner products between the query and the fine dictionary elements.

\begin{table}[t]
	\caption{Comparison of Multi-D-ADC system with different quantization algorithms in terms of recall$@R$ with $R$ being $1,10,100$, time cost (in millisecond) with database vector reconstruction ($T1$), time cost (in millisecond) without database vector reconstruction but through distance lookup tables ($T2$). $L$ is the length of the candidate list reranked by the system.}
	\label{tab:InvertedMultiIndex}
\centering
\vspace{-.3cm}
\resizebox{1\linewidth}{!}{
		\begin{tabular}{|l|c|ccc|c|c|}
			\hline
			Alg. & $L$ & $R@1$ & $R@10$ & $R@100$ & $T1$ & $T2$\\
			\hline
			\hline
			\multicolumn{7}{|c|}{BIGANN, 1 billion SIFTs, $64$ bits per vector}\\
			\hline
			PQ  &  \multirow{5}*{$ 10000 $} & $ 0.158 $ & $ 0.479 $ &$  0.713 $ &$  6.2 $ & $ 4.1 $ \\
			CKM  &  & $ 0.181 $ & $ 0.525 $ & $ 0.751 $ & $ 11.9 $ & $ 4.6 $ \\
			NOCQ  &  & $ 0.195 $ & $ 0.558 $ & $ 0.765 $ & $ 15.7 $ & $ 7.1 $\\
			SNOCQ1  &  & $ 0.184 $ & $ 0.530 $ & $ 0.736 $ & $ 7.3 $ & $ 4.3 $\\
			SNOCQ2  &  & $ 0.191 $ & $ 0.546 $ & $ 0.754 $ & $ 8.6 $  & $ 4.5 $ \\
			\hline
			PQ  &  \multirow{5}*{$ 30000 $} & $ 0.172 $ & $ 0.507 $ & $ 0.814 $ & $ 13.2 $ & $ 9.8 $ \\
			CKM  &  & $ 0.193 $ & $ 0.556 $ & $ 0.851 $ & $ 30.3 $ & $ 10.1 $\\
			NOCQ  &  & $ 0.200 $ & $ 0.597 $ & $ 0.869 $ & $ 42.6 $ &$  12.9 $\\
			SNOCQ1  &  & $ 0.192 $ & $ 0.571 $ & $ 0.849 $ & $ 15.8 $ & $ 9.9 $\\
			SNOCQ2 &  & $ 0.198 $ & $ 0.586 $ & $ 0.860 $ & $ 19.9 $ &$  10.0 $ \\
			\hline
			PQ  &  \multirow{5}*{$ 100000 $} & $ 0.173 $ & $ 0.517 $ & $ 0.862 $ & $ 37.4 $ & $ 30.5 $\\
			CKM  &  & $ 0.195 $ & $ 0.568 $ & $ 0.892 $ &  $ 95.8 $ & $ 31.6 $\\
			NOCQ  &  &$  0.204 $ & $ 0.612 $ & $ 0.920 $ & $ 125.9 $  & $ 33.4 $\\
			SNOCQ1 &  & $ 0.194 $ & $ 0.584 $ & $ 0.903 $ & $ 43.7 $ & $ 30.9 $\\
			SNOCQ2  &  & $ 0.199 $ & $ 0.597 $ & $ 0.907 $ & $ 58.6 $ &$  31.2 $\\
			\hline
			\hline
			\multicolumn{7}{|c|}{BIGANN, 1 billion SIFTs, $128$ bits per vector}\\
			\hline
			PQ  & \multirow{5}*{$ 10000 $} & $ 0.312 $ & $ 0.673 $ & $ 0.739 $ & $ 7.0 $ & $ 5.5 $\\
			CKM  &  & $ 0.357 $ & $ 0.718 $ & $ 0.772 $ & $ 12.4 $ & $ 5.8 $\\
			NOCQ  &  & $ 0.379 $ & $ 0.738 $ & $ 0.781 $ & $ 29.0 $ & $ 7.9 $\\
			SNOCQ1  &  & $ 0.347 $ & $ 0.702 $ & $ 0.755 $ & $ 8.2 $ & $ 5.6 $\\
			SNOCQ2  &  & $ 0.368 $ & $ 0.725 $ & $ 0.773 $ &$  9.5 $ & $ 5.7 $\\
			\hline
			PQ  &  \multirow{5}*{$ 30000 $} & $ 0.337 $ & $ 0.765 $ & $ 0.883 $ & $ 15.8 $ & $ 14.1 $ \\
			CKM  &  & $ 0.380 $ & $ 0.811 $ & $ 0.903  $& $ 32.7 $ & $ 14.4 $\\
			NOCQ  &  & $ 0.404 $ & $ 0.833 $ & $ 0.906 $ & $ 76.4 $ & $ 16.8 $\\
			SNOCQ1  &  & $ 0. 372 $ & $ 0.802 $ & $ 0.890 $ & $ 18.9 $ & $ 14.3 $\\
			SNOCQ2  &  & $ 0.392 $ & $ 0.821 $ & $ 0.904 $ & $ 25.8 $ & $ 14.4 $\\
			\hline
			PQ  &  \multirow{5}*{$ 100000 $} & $ 0.345 $ & $ 0.809 $ & $ 0.964 $ & $ 48.7 $ & $ 43.3 $\\
			CKM  &  & $ 0.389 $ & $ 0.848 $ & $ 0.970 $ & $ 107.6 $ & $ 44.9 $\\
			NOCQ  &  & $ 0.413 $ & $ 0.877 $ & $ 0.975 $ & $ 242.3 $ & $ 47.3 $\\
			SNOCQ1  &  & $ 0.381 $ & $ 0.852 $ & $ 0.969 $ & $ 59.3 $ & $ 43.6 $\\
			SNOCQ2  &  & $ 0.401 $ & $ 0.858 $ & $ 0.971 $ &  $ 77.4 $ & $ 43.9 $\\
			\hline
		\end{tabular}
}
\end{table}

\noindent\textbf{Results.}
We compare the proposed approach, NOCQ,
with PQ and CKM,
and use them to train the coarse and fine quantizers.
In addition,
we also report the results of the sparse version
of our approach~\cite{ZhangQTW15}:
the dictionary is sparse,
thus the distance table computation is accelerated.
We consider two sparsity degrees:
one,
termed as SNOCQ1, has the same sparse degree (i.e., the number of nonzeros $= KD$)
with PQ,
and the other, SNOCQ2,
in which the number of nonzeros is equal to $\min (KD + D^2, MKD)$,
has similar distance table computation cost with CKM.
Following~\cite{BabenkoL12}, for all the approaches,
we use $K=2^{14}$ to build coarse dictionaries
and $M=8,16$ for compact code representation.
The performance comparison
in terms of recall$@R$ ($R = 1,10,100$),
$T1$ (query time cost for the scheme with database vector reconstruction),
$T2$ (query time cost for the scheme without database vector reconstruction but through distance lookup tables)
with respect to the length of the retrieved candidate list $L$  ($L = 10000,30000,100000$)
is summarized in Table~\ref{tab:InvertedMultiIndex}.

It can be seen in terms of the search accuracy that
our approach NOCQ performs the best
and that our approach SNOCQ1 (SNOCQ2) performs better than PQ (CKM).
Compared with the $T1$ and $T2$ columns, our approaches,
SNOCQ1 and SNOCQ2, like PQ and CKM,
are both accelerated,
and according to the $T2$ column, the query costs of our approaches (SNOCQ1 and SNOCQ2) are almost the same to that of the most efficient approach PQ.
Considering the overall performance
in terms of the query cost for the accelerated scheme ($T2$)
and the recall,
one can see that SNOCQ2 performs the best.
For example,
the query cost of SNOCQ2 is $36\%$ smaller than that of NOCQ
when retrieving $10000$ candidates with $64$ bits
and the recall decreases less than $2.2\%$.
In other cases,
the recall decrease is always less than $3\%$
while the query cost is saved at least $7\%$.
In summary,
our approach NOCQ is the best choice
if the code length, thus storage and memory cost,
is top priority,
or if a larger number of candidate images
are needed to rerank,
and its sparse version performs the best
when concerning about both query time and storage cost.

\subsection{Inner Product Similarity Search}
In this section,
we study the application of composite quantization
to maximum inner product similarity search,
which aims to find a database vector $\mathbf{x}^*$
so that $\mathbf{x}^* = \arg\max_{\mathbf{x} \in \mathcal{X}} \texttt{<} \mathbf{q}, \mathbf{x}\texttt{>}$.
Given the approximation
$\mathbf{x} \approx \bar{\mathbf{x}} = \sum_{m=1}^M \mathbf{c}_{mk_m}$,
the approximated inner product between a query $\mathbf{q}$
and a database vector $\mathbf{x}$
is computed by
$\texttt{<}  \mathbf{q}, \mathbf{x} \texttt{>}
\approx \texttt{<}  \mathbf{q}, \bar{\mathbf{x}} \texttt{>}
= \sum_{m=1}^M \texttt{<}  \mathbf{q}, \mathbf{c}_{mk_m} \texttt{>}$.
The computation cost can be reduced to $O(M)$
by looking up an inner product table
storing $MK$ inner products,
$\{\texttt{<} \mathbf{q}, \mathbf{c}_{mk}\texttt{>}~|~ m=1, \cdots, M,
k=1, \cdots, K\}$.

Similar to the triangle inequality for upper-bounding the Euclidean distance approximation error,
we have a property
for the inner product distance approximation error,
and the property is given as,
\begin{Property}
	\label{property:innerproductbound}
	Given a query vector $\mathbf{q}$,
	a data vector $\mathbf{x}$
	and its approximation $\bar{\mathbf{x}}$,
	the absolute difference between
	the true inner product
	and the approximated inner product
	is upper-bounded:
	\begin{align}
	|\texttt{<} \mathbf{q}, \mathbf{x} \texttt{>} - \texttt{<} \mathbf{q}, \bar{\mathbf{x}} \texttt{>}|
	\leqslant \|\mathbf{x} - \bar{\mathbf{x}}\|_2 \|\mathbf{q}\|_2.
	\end{align}
\end{Property}

The upper bound
is related to the $L_2$ norms of $\mathbf{q}$,
meaning that the bound depends on the query $\mathbf{q}$
(in contrast,
the upper bound for Euclidean distance does not depend on the query).
However,
the solution in inner product similarity search
does not depend on the $L_2$ norm of the query
as queries with different $L_2$ norm have the same solution,
i.e.,
$\mathbf{x}^* = \arg\max_{\mathbf{x} \in \mathcal{X}} \texttt{<} \mathbf{q}, \mathbf{x}\texttt{>}
= \arg\max_{\mathbf{x} \in \mathcal{X}} \texttt{<} s\mathbf{q}, \mathbf{x}\texttt{>}$,
where $s$ is an arbitrary positive number.
In this sense, it also holds
that
more accurate vector approximation can potentially
lead to better inner product similarity search.

\begin{figure}[t]
	\centering
	\includegraphics[width = .9\linewidth, clip]{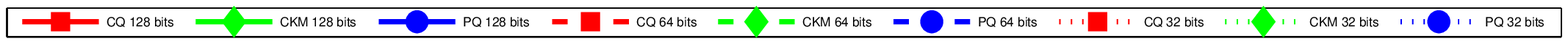}\\
	\subfigure[]{\includegraphics[width=.3\linewidth, clip]{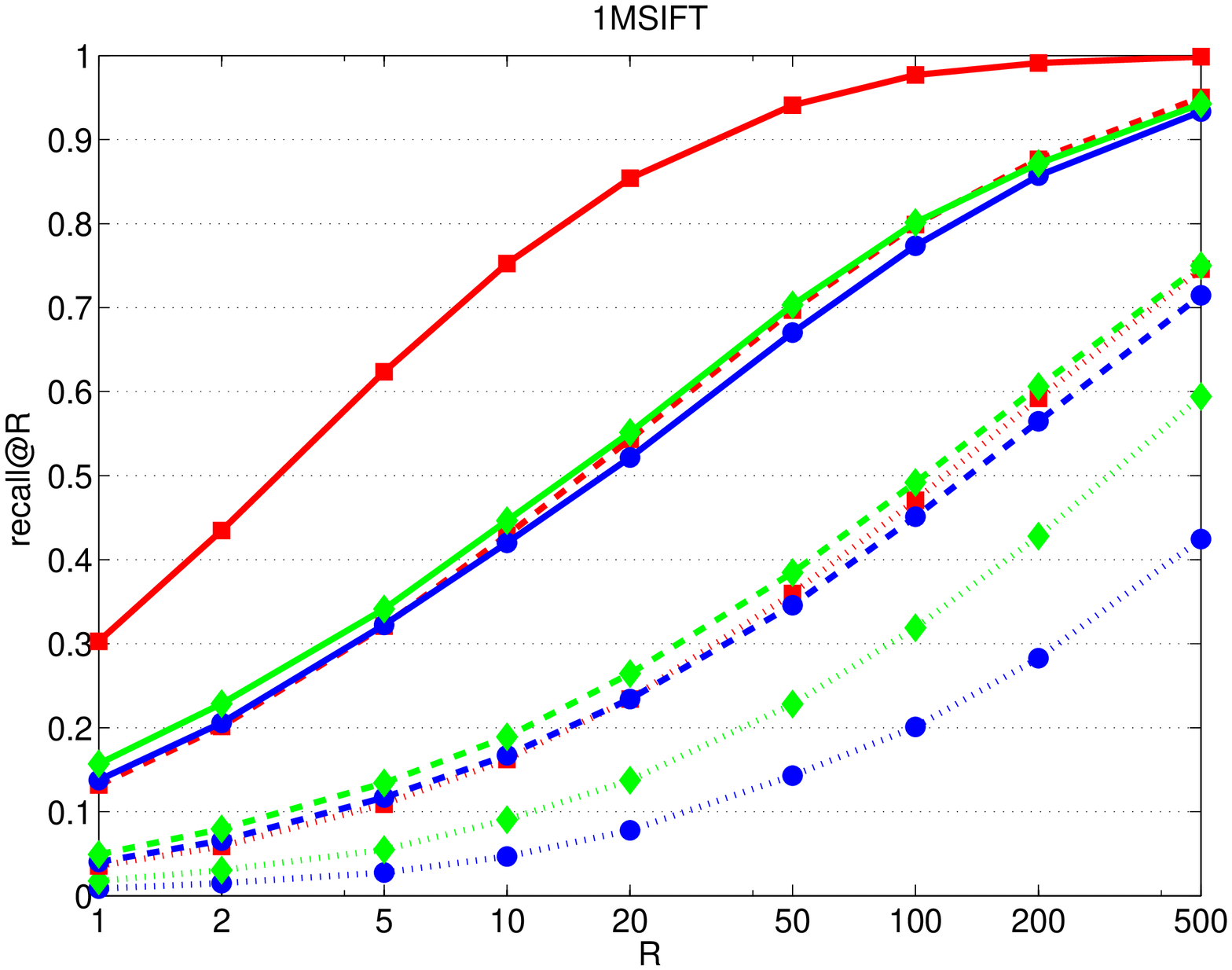}}
	\subfigure[]{\includegraphics[width=.3\linewidth, clip]{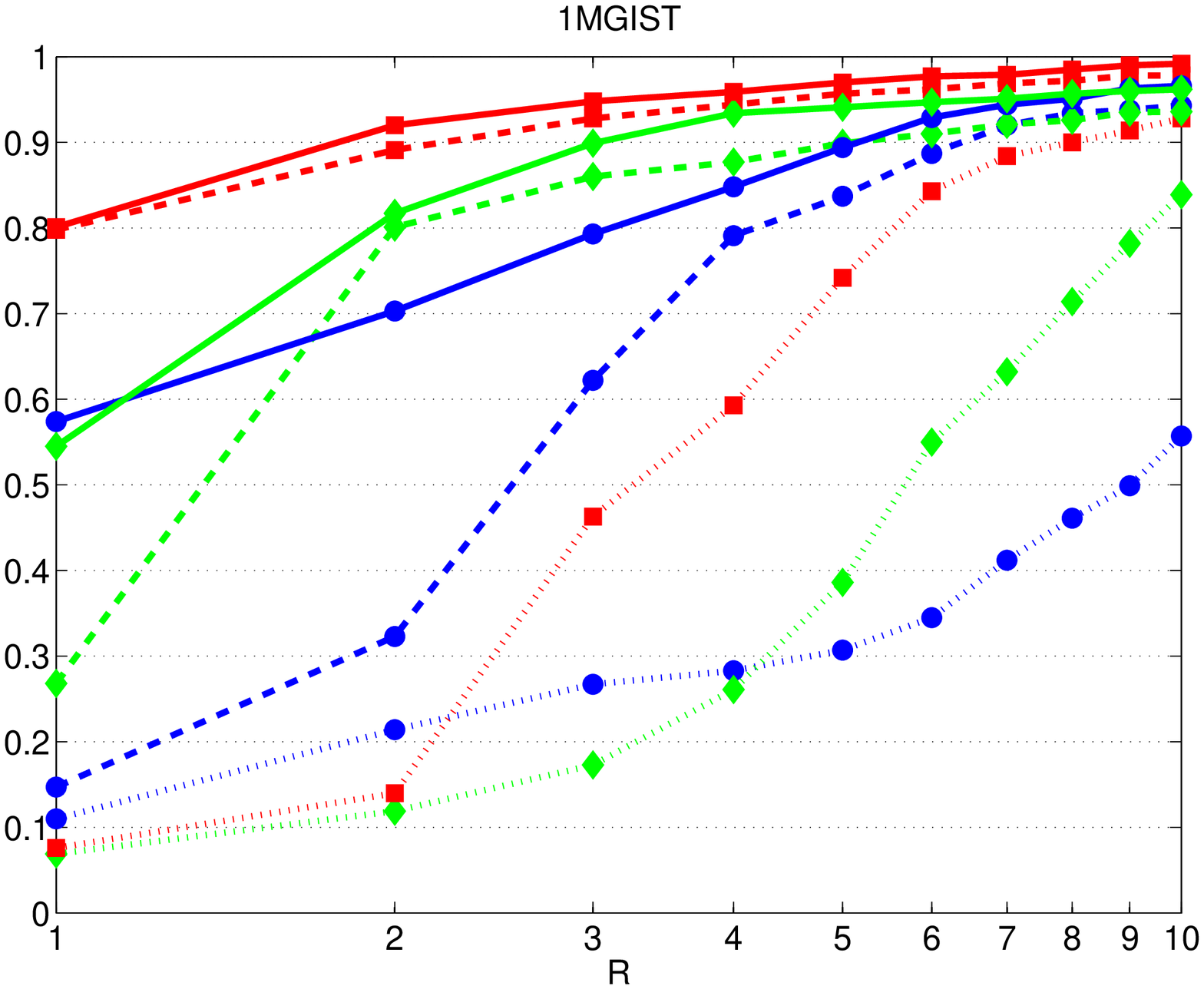}}
	\subfigure[]{\includegraphics[width=.3\linewidth, clip]{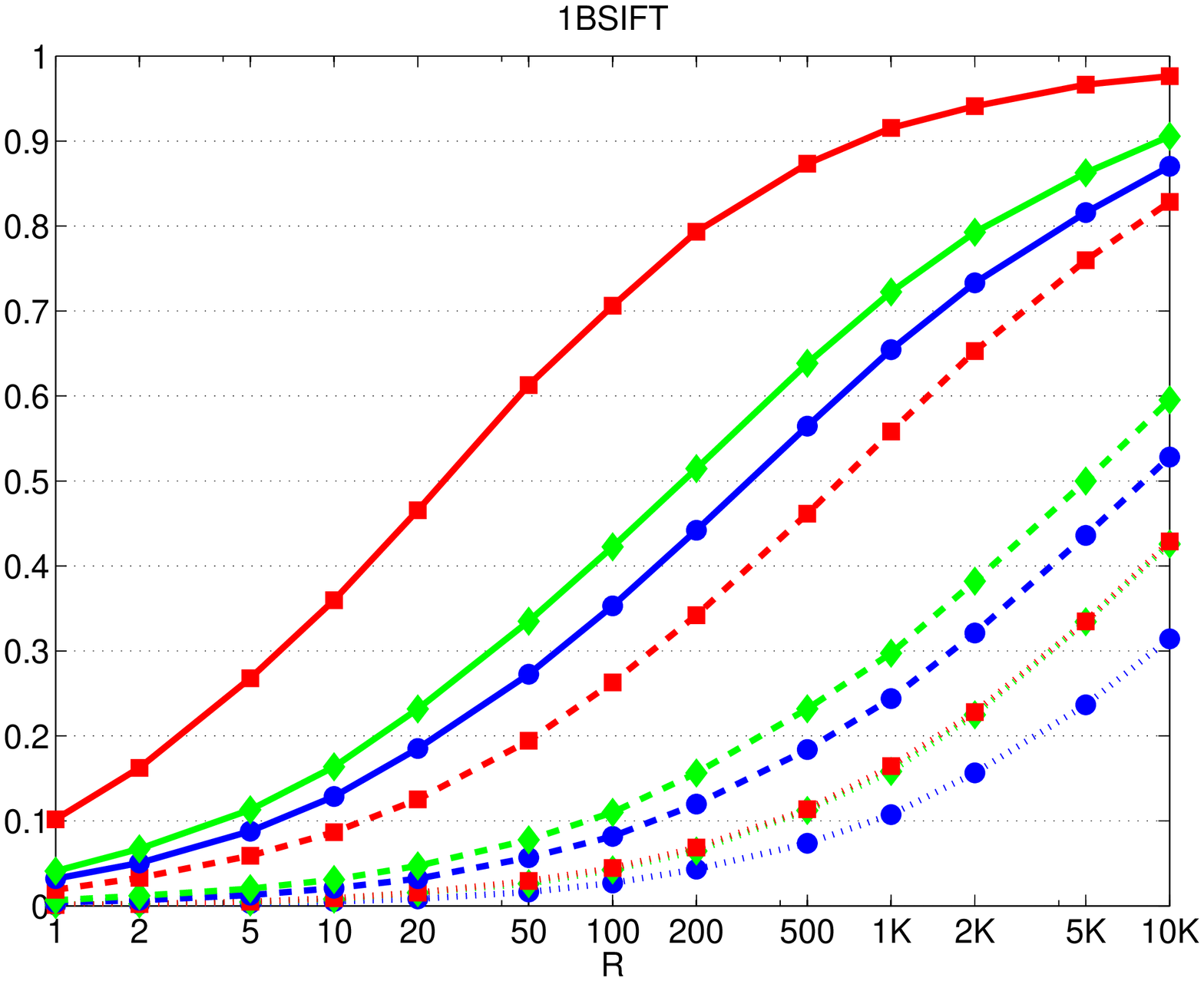}}
\vspace{-.3cm}
	\caption{The inner product search performance for different algorithms
		on (a) $1M$SIFT, (b) $1M$GIST and (c) $1B$SIFT.}
	\label{fig:resultsofinnerproductsearch}
\end{figure}

The results of different algorithms over
$1M$SIFT, $1M$GIST and $1B$SIFT are
shown in Figure~\ref{fig:resultsofinnerproductsearch}.
The groundtruth nearest neighbors of a given query
are computed by linear scanning all the database vectors and
evaluating the inner product similarity between the query and each database vector.
We can see that our approach gets the best performance
on all the datasets with various code lengths.

\subsection{Query Compression for Mobile Search}
Mobile search is an emerging direction of image and information retrieval.
It is usually expected
that the object transmitted from the client to the server
is as small as possible.
In this section,
we conduct an experiment:
the database points are processed
using the compact coding approach,
e.g., NOCQ, PQ and CKM;
the query is transformed to a compact code
(in the client side);
the query reconstructed (in the server side)
from the transformed compact code
is compared with the database points.
The query is compressed
in our approach NOCQ
using the non-orthogonal version
CQ as the reconstruction quality is better,
while the same coding scheme is used
in both the client and server sides
for other approaches.

\begin{figure}[t]
	\centering
	\includegraphics[width = .9\linewidth, clip]{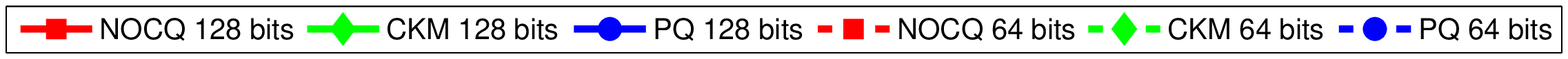}\\
	\subfigure[]{\includegraphics[width=.3\linewidth, clip]{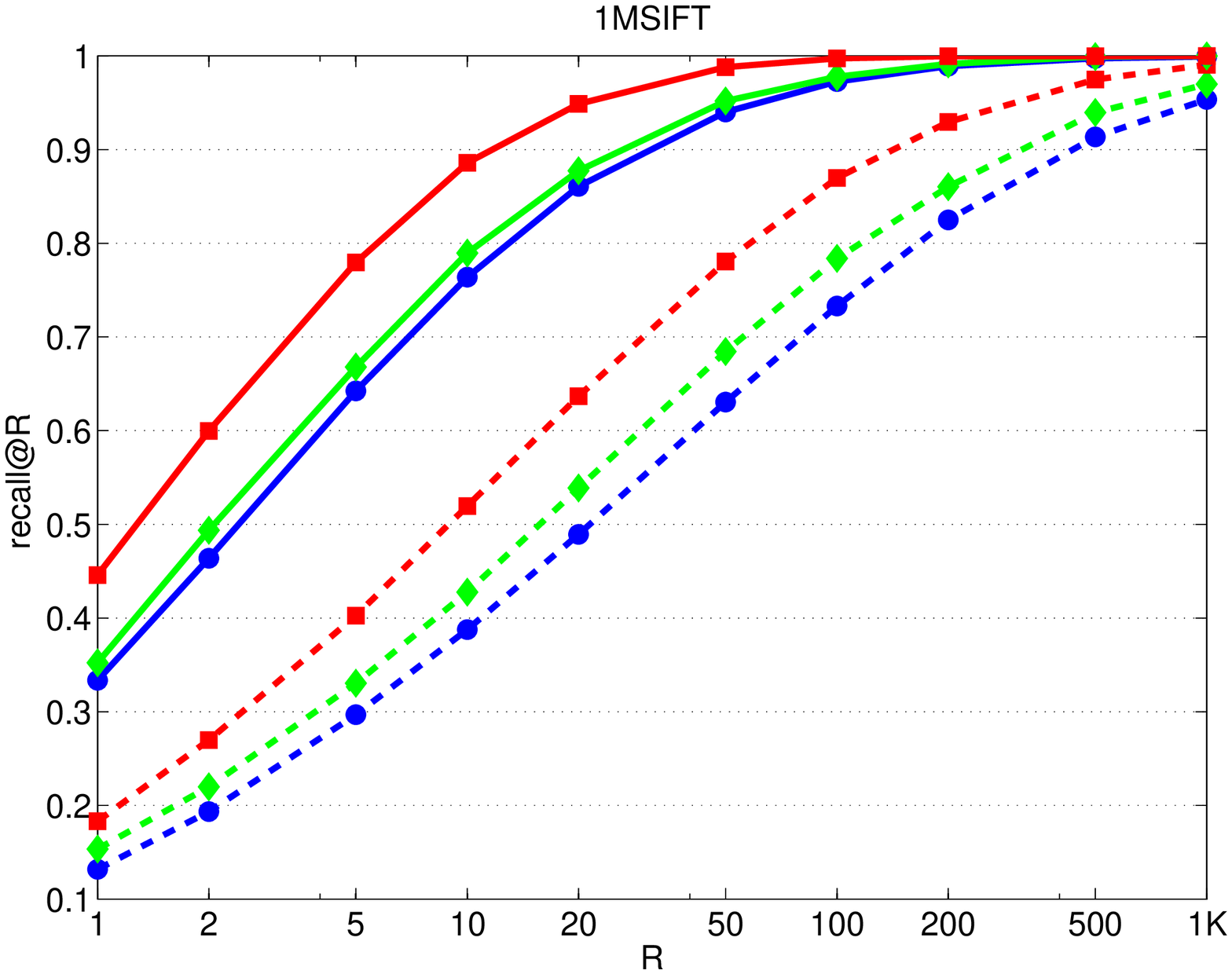}}
	\subfigure[]{\includegraphics[width=.3\linewidth, clip]{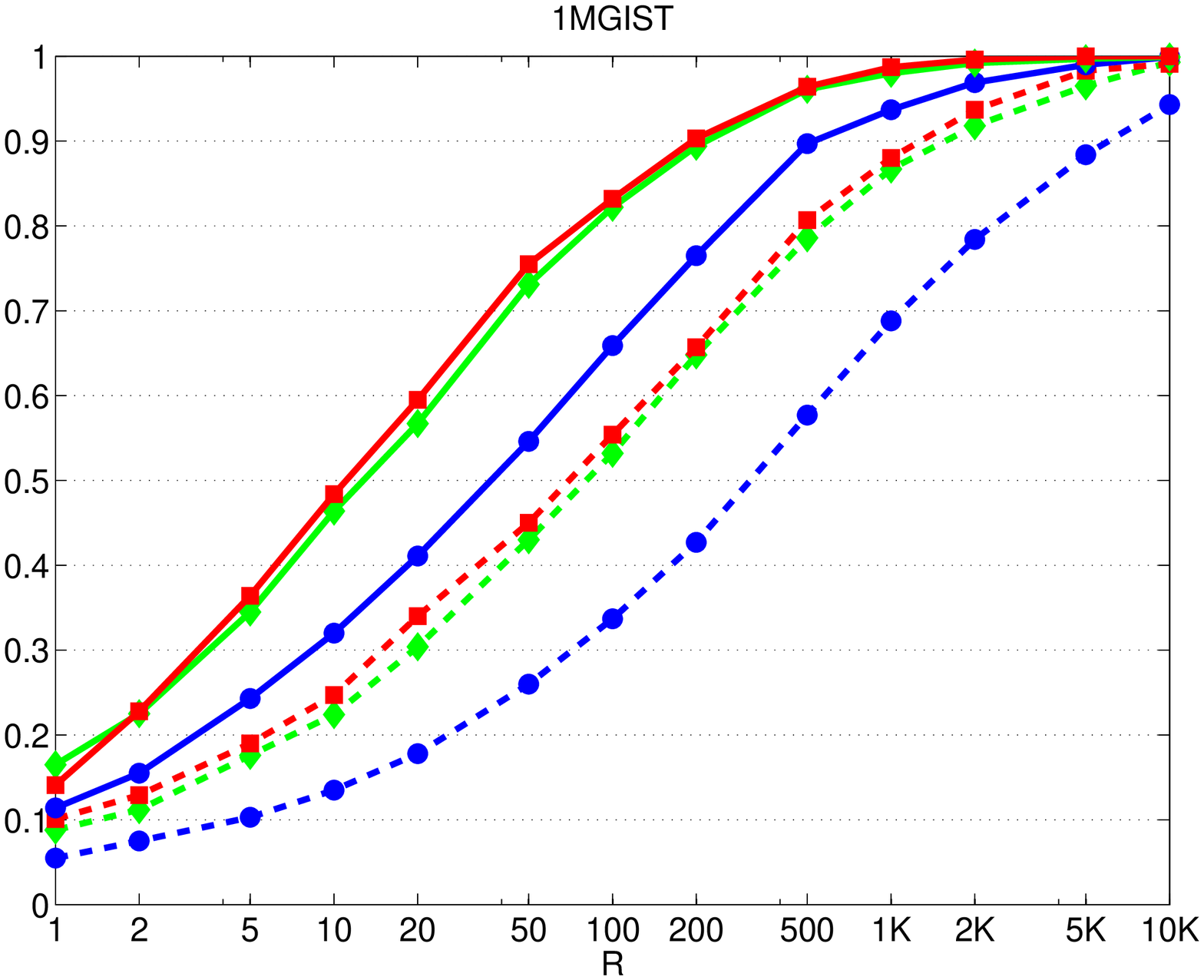}}
	\subfigure[]{\includegraphics[width=.3\linewidth, clip]{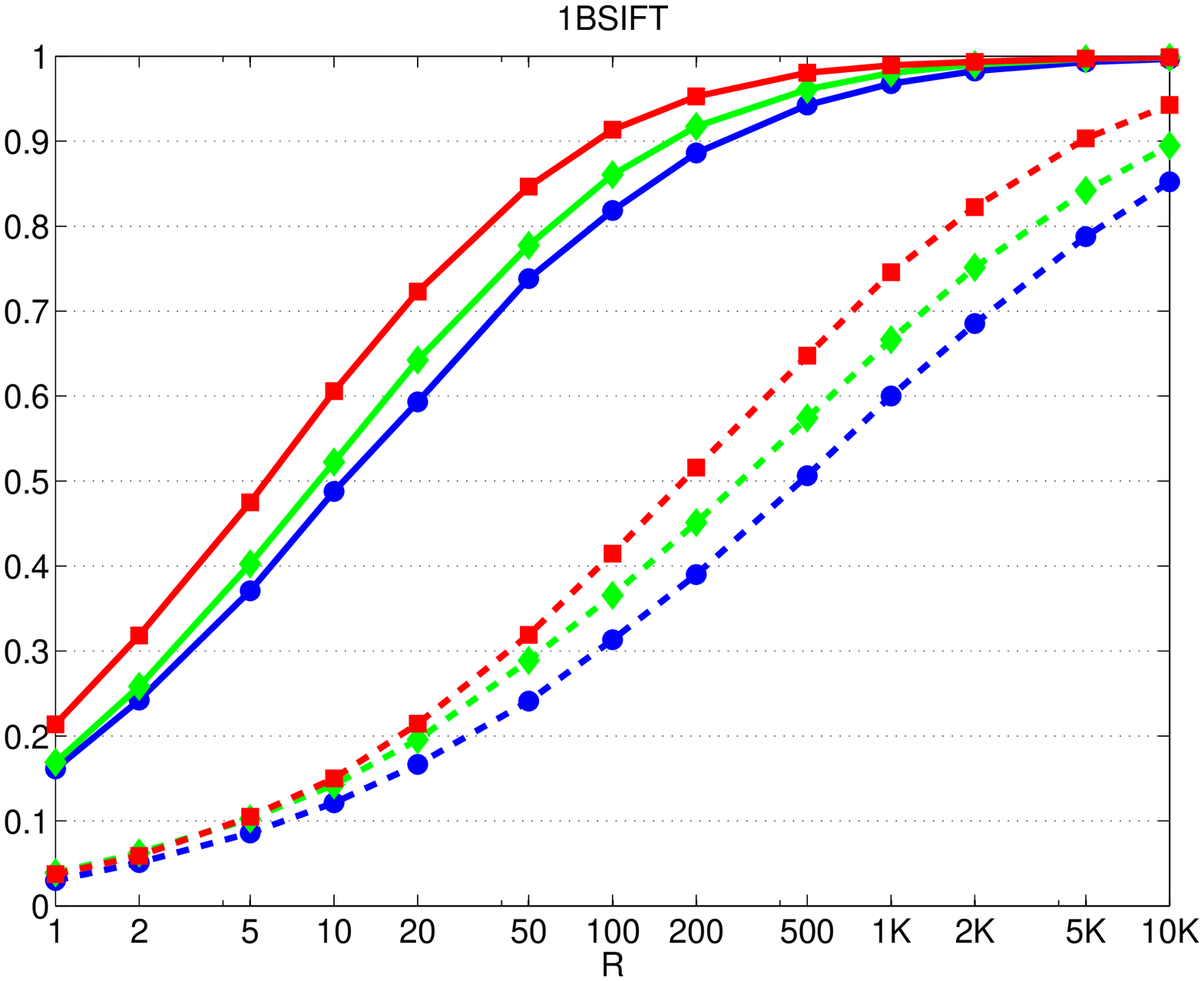}}
\vspace{-.3cm}
	\caption{The search performance
		with query reconstructed
		on (a) $1M$SIFT, (b) $1M$GIST, and (c) $1B$SIFT
		over $64$ bits and $128$ bits.}
	\label{fig:resultsofqueryreconstructive}
\end{figure}

The results are shown in Figure~\ref{fig:resultsofqueryreconstructive}
over $1M$SIFT, $1M$GIST, and $1B$SIFT (its whole data is used as the search database)
with $64$ bits and $128$ bits
for $1$-NN search.
It is as expected that our approach consistently performs the best.
In particularly,
the improvement gain of our approach
over the second best method, CKM,
is even more signification
compared with
the results shown in Figures~\ref{fig:resultsof1MSIFT1MGIST}
and~\ref{fig:resultsof1BSIFT}
obtained without compressing the query.
For example,
the recall gain for $1M$SIFT with $128$ bits
at position $1$
is increased from $0.3522$ to $0.4459$,
and
the recall gain for $1B$SIFT with $128$ bits
at position $10$
is increased from $0.5224$ to $0.6057$.

\begin{table}[t]
	\caption{The performance over the Holidays dataset in terms of MAP using different length of codes encoding.}
\vspace{-.3cm}
	\label{tab:HolidayPerformance}
\centering
		\begin{tabular}{|c|c|cccc|}
			\hline
			& $\sharp$Bits & ITQ & PQ & CKM & NOCQ\\
			\hline
			\multirow{3}*{Fisher}  & $ 32 $ & $ 0.413 $ & $ 0.504 $ & $ 0.537 $  & $ 0.550 $\\
			& $ 64 $ & $ 0.533 $ & $ 0.548 $ & $ 0.578 $   & $ 0.622 $\\
			& $ 128 $ & $ 0.588 $ & $ 0.579 $ & $ 0.598 $ & $ 0.634 $\\
			\hline
			\multirow{3}*{VLAD}	& $ 32 $ & $ 0.438 $ & $ 0.513 $ & $ 0.545 $   & $ 0.578 $\\
			& $ 64  $& $ 0.537 $ & $ 0.574 $ & $ 0.598 $  & $ 0.632 $\\
			& $ 128 $ & $ 0.607 $ & $ 0.586 $ &$  0.609  $ & $ 0.644 $\\
			\hline
		\end{tabular}
\end{table}

\begin{table}[t]
	\caption{The performance over the UKBench dataset in terms of scores using different length of codes encoding.}
\vspace{-.3cm}
	\label{tab:UkbenchPerformance}
\centering
		\begin{tabular}{|c|c|cccc|}
			\hline
			& $\sharp$Bits & ITQ & PQ & CKM  & NOCQ\\
			\hline
			\multirow{3}*{Fisher}  & $ 32 $ & $ 2.116 $ & $ 2.203 $ & $ 2.606 $  &  $ 2.740 $\\
			& $ 64 $ & $ 2.632 $ & $ 2.618 $ & $ 2.894 $  &  $ 3.009 $\\
			& $ 128 $ & $ 2.881 $ & $ 2.851 $ & $ 3.039 $  & $ 3.154 $ \\
			\hline
			\multirow{3}*{VLAD}	& $ 32 $ & $ 2.093 $ & $ 2.214 $ & $ 2.631 $   & $ 2.746 $\\
			& $ 64 $ &$  2.617 $ & $ 2.629 $ & $ 2.925 $  & $ 3.046 $  \\
			& $ 128 $ & $ 2.895 $ & $ 2.878 $ & $ 3.069 $  & $ 3.185 $\\
			\hline
		\end{tabular}
\end{table}

\subsection{Application to object retrieval}
We report the results of the applications to object retrieval.
In object retrieval,
images are represented as an aggregation of local descriptors,
often thousands of dimension.
We evaluate the performances
over the $4096$-dimensional Fisher vectors~\cite{PerronninD07}
and the $4096$-dimensional VLAD vectors~\cite{JegouDSP10}
extracted
from the INRIA Holidays dataset~\cite{JegouDS08}
that contains $500$ query and
$991$ relevant images,
and the UKBench dataset~\cite{NisterS06}
that contains $2550$ groups of $4$ images each
(totally $10200$ images).

The search performances
in terms of mean average precision (MAP)~\cite{JegouDS08}
for the Holidays dataset
and
the score~\cite{NisterS06} for the UKBench dataset
are shown in Table~\ref{tab:HolidayPerformance}
and Table~\ref{tab:UkbenchPerformance}.
It can be seen
that NOCQ performs the best,
which is because our approach (NOCQ)
produces better vector approximation.

\section{Conclusion}
In this paper,
we present a compact coding approach,
near-orthogonal composite quantization,
to approximate nearest neighbor search.
The superior search accuracy
stems from
that it exploits the composition of dictionary elements
to approximate a vector,
yielding smaller quantization errors.
The search efficiency is guaranteed
by imposing the near-orthogonality constraint
and discarding its computation.
The empirical results
suggest that near-orthogonal composite quantization
outperforms existing methods for search under the Euclidean-distance
and composite quantization achieves superior performance
for search with the inner-product similarity.
Composite quantization has been extended
to semantic quantization~\cite{WangZQTW16}
for semantic retrieval
and multi-modality quantization~\cite{ZhangW16} for multi-modality search.



\section*{Appendix: Proof of Theorem~\ref{theorem:upperbound}}
\begin{proof}
	We have the following inequality,
	\begin{align}
	&|\tilde{d}(\mathbf{q},\bar{\mathbf{x}})-\hat{d}(\mathbf{q},\mathbf{x})| \nonumber \\
	=~& |\tilde{d}(\mathbf{q},\bar{\mathbf{x}})- \hat{d}(\mathbf{q}, \bar{\mathbf{x}}) + \hat{d}(\mathbf{q}, \bar{\mathbf{x}}) -  \hat{d}(\mathbf{q},\mathbf{x})| \nonumber \\
	\leq~& |\tilde{d}(\mathbf{q},\bar{\mathbf{x}})- \hat{d}(\mathbf{q}, \bar{\mathbf{x}})| +
	|\hat{d}(\mathbf{q}, \bar{\mathbf{x}}) - \hat{d}(\mathbf{q},\mathbf{x})|.
	\end{align}
	
	We will show (a) $|\tilde{d}(\mathbf{q},\bar{\mathbf{x}})- \hat{d}(\mathbf{q}, \bar{\mathbf{x}})| \leq |\delta|^{1/2}$
	and (b) $|\hat{d}(\mathbf{q}, \bar{\mathbf{x}}) - \hat{d}(\mathbf{q},\mathbf{x})| \leq \|\mathbf{x} - \bar{\mathbf{x}}\|_2$,
	respectively.
	
	The proof for (a) is given as follows,
	\begin{align}
	&|\tilde{d}(\mathbf{q},\bar{\mathbf{x}})- \hat{d}(\mathbf{q}, \bar{\mathbf{x}})|^2 \nonumber \\
	\leq ~& |\tilde{d}(\mathbf{q},\bar{\mathbf{x}}) - \hat{d}(\mathbf{q}, \bar{\mathbf{x}})| |\tilde{d}(\mathbf{q},\bar{\mathbf{x}}) + \hat{d}(\mathbf{q}, \bar{\mathbf{x}})| \nonumber \\
	=~& |\tilde{d}^2(\mathbf{q},\bar{\mathbf{x}}) - \hat{d}^2(\mathbf{q}, \bar{\mathbf{x}}) | \label{eqn:ptt} \\
	=~& |\delta| \label{eqn:prr}.
	\end{align}
	The last equality from (\ref{eqn:ptt}) to (\ref{eqn:prr}) holds because we have
	$\hat{d}^2(\mathbf{q},\bar{\mathbf{x}}) = \tilde{d}^2(\mathbf{q},\bar{\mathbf{x}}) + \delta$.
	
	The proof for (b) is presented in the following.
	For convenience, we denote $\eta = (M-1)\|\mathbf{q}\|_2^2 \geq 0$.
	\begin{align}
	&|\hat{d}(\mathbf{q}, \bar{\mathbf{x}}) - \hat{d}(\mathbf{q},\mathbf{x})| \nonumber \\
	=~& |\sqrt{d^2(\mathbf{q}, \bar{\mathbf{x}})+\eta } - \sqrt{ d^2(\mathbf{q}, \mathbf{x}) + \eta} | \nonumber  \\
	=~&\frac{ |(d^2(\mathbf{q}, \bar{\mathbf{x}})+\eta ) -( d^2(\mathbf{q}, \mathbf{x}) +\eta)  |}
	{\sqrt{d^2(\mathbf{q}, \bar{\mathbf{x}})+\eta }+ \sqrt{ d^2(\mathbf{q}, \mathbf{x}) + \eta} } \nonumber \\
	\leq~& \frac{ | d^2(\mathbf{q}, \bar{\mathbf{x}}) - d^2(\mathbf{q}, \mathbf{x})|}
	{d(\mathbf{q}, \bar{\mathbf{x}}) +  d(\mathbf{q}, \mathbf{x})} \nonumber  \\
	= ~& |d(\mathbf{q}, \bar{\mathbf{x}}) - d(\mathbf{q}, \mathbf{x}) |  \nonumber \\
	\leq ~& d(\mathbf{x}, \bar{\mathbf{x}}) ~(\text{by the triangle inequality}) \nonumber \\
	=~& \|\mathbf{x} - \bar{\mathbf{x}}\|_2.
	\end{align}
	We can easily validate (b)
	in the case that the denominator happens to be $0$
	in the above proof.
	
	Thus, the theorem holds.
\end{proof}

\section*{Appendix: Proof of Property~\ref{property:innerproductbound}}
\begin{proof}
	The proof is simple
	and given as follows.
	Look at the absolute value of the inner product approximation error,
	\begin{align}
	&|\texttt{<} \mathbf{q}, \mathbf{x} \texttt{>} - \texttt{<}\mathbf{q}, \bar{\mathbf{x}}\texttt{>}| \\
	=~& |\texttt{<}\mathbf{q}, \mathbf{x} -\bar{\mathbf{x}}\texttt{>} |~\text{\emph {(by the distributive property)}}\\
	\leq~&\| \mathbf{x} -\bar{\mathbf{x}} \|_2 \| \mathbf{q} \|_2\\
	\leq~& C\| \mathbf{x} -\bar{\mathbf{x}} \|_2.
	\end{align}
	Thus, the approximation error is upper-bounded by
	$C\| \mathbf{x} -\bar{\mathbf{x}} \|_2$, assuming that $\|\mathbf{q}\|_2$ is upper-bounded by a constant $C$.
\end{proof}

\bibliographystyle{ieee}
\bibliography{hash,bow}

\begin{thebibliography}{10}\itemsep=-1pt

\bibitem{BabenkoK14}
A.~Babenko and V.~Lempitsky.
\newblock Additive quantization for extreme vector compression.
\newblock In {\em CVPR}, pages 931--939, 2014.

\bibitem{BabenkoK15}
A.~Babenko and V.~Lempitsky.
\newblock Tree quantization for large-scale similarity search and
  classification.
\newblock In {\em CVPR}, 2015.

\bibitem{BabenkoL12}
A.~Babenko and V.~S. Lempitsky.
\newblock The inverted multi-index.
\newblock In {\em CVPR}, pages 3069--3076, 2012.

\bibitem{BabenkoL14}
A.~Babenko and V.~S. Lempitsky.
\newblock Improving bilayer product quantization for billion-scale approximate
  nearest neighbors in high dimensions.
\newblock {\em CoRR}, abs/1404.1831, 2014.

\bibitem{DuW14}
C.~Du and J.~Wang.
\newblock Inner product similarity search using compositional codes.
\newblock {\em CoRR}, abs/1406.4966, 2014.

\bibitem{FriedmanBF77}
J.~H. Friedman, J.~L. Bentley, and R.~A. Finkel.
\newblock An algorithm for finding best matches in logarithmic expected time.
\newblock {\em ACM Trans. Math. Softw.}, 3(3):209--226, 1977.

\bibitem{GeHK013}
T.~Ge, K.~He, Q.~Ke, and J.~Sun.
\newblock Optimized product quantization for approximate nearest neighbor
  search.
\newblock In {\em CVPR}, pages 2946--2953, 2013.

\bibitem{GongKVL12}
Y.~Gong, S.~Kumar, V.~Verma, and S.~Lazebnik.
\newblock Angular quantization-based binary codes for fast similarity search.
\newblock In {\em NIPS}, pages 1205--1213, 2012.

\bibitem{GongL11}
Y.~Gong and S.~Lazebnik.
\newblock Iterative quantization: A procrustean approach to learning binary
  codes.
\newblock In {\em CVPR}, pages 817--824, 2011.

\bibitem{GongLGP13}
Y.~Gong, S.~Lazebnik, A.~Gordo, and F.~Perronnin.
\newblock Iterative quantization: A procrustean approach to learning binary
  codes for large-scale image retrieval.
\newblock {\em IEEE Trans. Pattern Anal. Mach. Intell.}, 35(12):2916--2929,
  2013.

\bibitem{GordoP11}
A.~Gordo and F.~Perronnin.
\newblock Asymmetric distances for binary embeddings.
\newblock In {\em CVPR}, pages 729--736, 2011.

\bibitem{GordoPGL14}
A.~Gordo, F.~Perronnin, Y.~Gong, and S.~Lazebnik.
\newblock Asymmetric distances for binary embeddings.
\newblock {\em IEEE Trans. Pattern Anal. Mach. Intell.}, 36(1):33--47, 2014.

\bibitem{GrayN98}
R.~M. Gray and D.~L. Neuhoff.
\newblock Quantization.
\newblock {\em IEEE Transactions on Information Theory}, 44(6):2325--2383,
  1998.

\bibitem{HeoLY14}
J.-P. Heo, Z.~Lin, and S.-E. Yoon.
\newblock Distance encoded product quantization.
\newblock In {\em CVPR}, pages 2139--2146, 2014.

\bibitem{JegouDS08}
H.~Jegou, M.~Douze, and C.~Schmid.
\newblock Hamming embedding and weak geometric consistency for large scale
  image search.
\newblock In {\em ECCV}, pages 304--317, 2008.

\bibitem{JegouDS11}
H.~J{\'e}gou, M.~Douze, and C.~Schmid.
\newblock Product quantization for nearest neighbor search.
\newblock {\em IEEE Trans. Pattern Anal. Mach. Intell.}, 33(1):117--128, 2011.

\bibitem{JegouDSP10}
H.~J{\'e}gou, M.~Douze, C.~Schmid, and P.~P{\'e}rez.
\newblock Aggregating local descriptors into a compact image representation.
\newblock In {\em CVPR}, pages 3304--3311, 2010.

\bibitem{JegouTDA11}
H.~J{\'e}gou, R.~Tavenard, M.~Douze, and L.~Amsaleg.
\newblock Searching in one billion vectors: Re-rank with source coding.
\newblock In {\em ICASSP}, pages 861--864, 2011.

\bibitem{KalantidisA14}
Y.~Kalantidis and Y.~Avrithis.
\newblock Locally optimized product quantization for approximate nearest
  neighbor search.
\newblock In {\em CVPR}, pages 2329--2336, 2014.

\bibitem{KongL12a}
W.~Kong and W.-J. Li.
\newblock Isotropic hashing.
\newblock In {\em NIPS}, pages 1655--1663, 2012.

\bibitem{KrizhevskySH12}
A.~Krizhevsky, I.~Sutskever, and G.~E. Hinton.
\newblock Imagenet classification with deep convolutional neural networks.
\newblock In {\em Advances in Neural Information Processing Systems 25: 26th
  Annual Conference on Neural Information Processing Systems 2012. Proceedings
  of a meeting held December 3-6, 2012, Lake Tahoe, Nevada, United States.},
  pages 1106--1114, 2012.

\bibitem{LeCunBBH01}
Y.~LeCun, L.~Bottou, Y.~Bengio, and P.~Haffner.
\newblock Gradient-based learning applied to document recognition.
\newblock In {\em Intelligent Signal Processing}, pages 306--351. IEEE Press,
  2001.

\bibitem{LiuWJJC12}
W.~Liu, J.~Wang, R.~Ji, Y.-G. Jiang, and S.-F. Chang.
\newblock Supervised hashing with kernels.
\newblock In {\em CVPR}, pages 2074--2081, 2012.

\bibitem{MatsuiYZ15}
Y.~Matsui, T.~Yamasaki, and K.~Aizawa.
\newblock Pqtable: Fast exact asymmetric distance neighbor search for product
  quantization using hash tables.
\newblock 2015.

\bibitem{MujaL09}
M.~Muja and D.~G. Lowe.
\newblock Fast approximate nearest neighbors with automatic algorithm
  configuration.
\newblock In {\em VISSAPP (1)}, pages 331--340, 2009.

\bibitem{NisterS06}
D.~Nist{\'e}r and H.~Stew{\'e}nius.
\newblock Scalable recognition with a vocabulary tree.
\newblock In {\em CVPR (2)}, pages 2161--2168, 2006.

\bibitem{NorouziF13}
M.~Norouzi and D.~J. Fleet.
\newblock Cartesian k-means.
\newblock In {\em CVPR}, pages 3017--3024, 2013.

\bibitem{PerronninD07}
F.~Perronnin and C.~R. Dance.
\newblock Fisher kernels on visual vocabularies for image categorization.
\newblock In {\em CVPR}, 2007.

\bibitem{ILSVRC15}
O.~Russakovsky, J.~Deng, H.~Su, J.~Krause, S.~Satheesh, S.~Ma, Z.~Huang,
  A.~Karpathy, A.~Khosla, M.~Bernstein, A.~C. Berg, and L.~Fei-Fei.
\newblock Imagenet large scale visual recognition challenge.
\newblock {\em International Journal of Computer Vision (IJCV)}, pages 1--42,
  April 2015.

\bibitem{RussellTMF08}
B.~C. Russell, A.~Torralba, K.~P. Murphy, and W.~T. Freeman.
\newblock Labelme: {A} database and web-based tool for image annotation.
\newblock {\em International Journal of Computer Vision}, 77(1-3):157--173,
  2008.

\bibitem{ShakhnarovichDI06}
G.~Shakhnarovich, T.~Darrell, and P.~Indyk.
\newblock {\em Nearest-Neighbor Methods in Learning and Vision: Theory and
  Practice}.
\newblock The MIT press, 2006.

\bibitem{Silpa-AnanH08}
C.~Silpa-Anan and R.~Hartley.
\newblock Optimised kd-trees for fast image descriptor matching.
\newblock In {\em CVPR}, 2008.

\bibitem{WangKC12}
J.~Wang, S.~Kumar, and S.-F. Chang.
\newblock Semi-supervised hashing for large-scale search.
\newblock {\em IEEE Trans. Pattern Anal. Mach. Intell.}, 34(12):2393--2406,
  2012.

\bibitem{WangL12}
J.~Wang and S.~Li.
\newblock Query-driven iterated neighborhood graph search for large scale
  indexing.
\newblock In {\em ACM Multimedia}, pages 179--188, 2012.

\bibitem{WangSSJ14}
J.~Wang, H.~T. Shen, J.~Song, and J.~Ji.
\newblock Hashing for similarity search: {A} survey.
\newblock {\em CoRR}, abs/1408.2927, 2014.

\bibitem{WangWSXSL14}
J.~Wang, J.~Wang, J.~Song, X.-S. Xu, H.~T. Shen, and S.~Li.
\newblock Optimized cartesian $k$-means.
\newblock {\em CoRR}, abs/1405.4054, 2014.

\bibitem{WangZSSS16}
J.~Wang, T.~Zhang, J.~Song, N.~Sebe, and H.~T. Shen.
\newblock A survey on learning to hash.
\newblock {\em CoRR}, abs/1606.00185, 2016.

\bibitem{WangZQTW16}
X.~Wang, T.~Zhang, G.-J. Qi, J.~Tang, and J.~Wang.
\newblock Supervised quantization for similarity search.
\newblock In {\em CVPR}, pages 2018--2026, 2016.

\bibitem{XuBLCHC13}
B.~Xu, J.~Bu, Y.~Lin, C.~Chen, X.~He, and D.~Cai.
\newblock Harmonious hashing.
\newblock In {\em IJCAI}, 2013.

\bibitem{YuKGC14}
F.~Yu, S.~Kumar, Y.~Gong, and S.-F. Chang.
\newblock Circulant binary embedding.
\newblock In {\em ICML (2)}, pages 946--954, 2014.

\bibitem{YuanL06}
M.~Yuan and Y.~Lin.
\newblock Model selection and estimation in regression with grouped variables.
\newblock {\em Journal of the Royal Statistical Society, Series B}, 68:49--67,
  2006.

\bibitem{ZhangDW14}
T.~Zhang, C.~Du, and J.~Wang.
\newblock Composite quantization for approximate nearest neighbor search.
\newblock In {\em ICML (2)}, pages 838--846, 2014.

\bibitem{ZhangQTW15}
T.~Zhang, G.-J. Qi, J.~Tang, and J.~Wang.
\newblock Sparse composite quantization.
\newblock In {\em CVPR}, pages 4548--4556, 2015.

\bibitem{ZhangW16}
T.~Zhang and J.~Wang.
\newblock Collaborative quantization for cross-modal similarity search.
\newblock In {\em CVPR}, pages 2036--2045, 2016.

\end{thebibliography}
\end{document}